\documentclass[times, review, 10pt]{elsarticle}

\usepackage{amssymb}

\usepackage{amsmath}
\usepackage{amsthm}
\usepackage{bm}
\usepackage{multirow}
\usepackage{graphics}
\usepackage{diagbox}
\usepackage{bbm}
\usepackage{caption}
\usepackage{xcolor}
\usepackage{dsfont}
\usepackage{subfigure}
\usepackage{booktabs}
\usepackage{makecell}
\usepackage{algorithm}
\usepackage{algorithmic}
\usepackage{textcomp}
\usepackage{bbding}
\usepackage{verbatim} 
\usepackage[algo2e,linesnumbered,lined,boxed,commentsnumbered,ruled]{algorithm2e}

\usepackage{hyperref}
 
\usepackage{cleveref}
\Crefname{figure}{Fig}{Figures}
\newtheorem{assumption}{Assumption}
\newtheorem{myDef}{Definition}

\journal{Pattern Recognition}

\begin{document}
\begin{frontmatter}

\title{{Beyond Deceptive Flatness: Dual-Order Solution for Strengthening Adversarial Transferability}}

\author[scu1]{Zhixuan Zhang}

\author[scu2]{Pingyu Wang}

\author[FDI]{Xingjian Zheng}

\author[scu1]{Linbo Qing}

\author[scut]{Qi Liu}

 \affiliation[scu1]{organization={School of Cyber Science and Engineering},
            addressline={Sichuan University},
            city={Chengdu},
            postcode={610207},
            state={Sichuan},
            country={China}}
            
\affiliation[scu2]{organization={College of Electronics and Information Engineering},
            addressline={Sichuan University},
            city={Chengdu},
            postcode={610065},
            state={Sichuan},
            country={China}}
            
\affiliation[FDI]{
            addressline={Company of Frost Drill Intellectual Software Pte. Ltd},
            country={Singapore}}

\affiliation[scut]{organization={School of Future Technology},
            addressline={South China University of Technology},
            city={Guangzhou},
            postcode={511442},
            state={Guangdong},
            country={China}}

\begin{abstract}
Transferable attacks generate adversarial examples on surrogate models to fool unknown victim models, posing real-world threats and growing research interest. Despite focusing on flat losses for transferable adversarial examples, recent studies still fall into suboptimal regions, especially the flat-yet-sharp areas, termed as deceptive flatness. In this paper, we introduce a novel black-box gradient-based transferable attack from a perspective of dual-order information. Specifically, we feasibly propose Adversarial Flatness (AF) to the deceptive flatness problem and a theoretical assurance for adversarial transferability. Based on this, using an efficient approximation of our objective, we instantiate our attack as Adversarial Flatness Attack (AFA), addressing the altered gradient sign issue. Additionally, to further improve the attack ability, we devise MonteCarlo Adversarial Sampling (MCAS) by enhancing the inner-loop sampling efficiency. The comprehensive results on ImageNet-compatible dataset demonstrate superiority over six baselines, generating adversarial examples in flatter regions and boosting transferability across model architectures. When tested on input transformation attacks or the Baidu Cloud API, our method outperforms baselines. 
\end{abstract}



\end{frontmatter}

\section{Introduction}
\label{Introduction}
Despite the excellent pattern recognition capabilities of Deep Neural Networks (DNNs) in various tasks, they remain vulnerable to adversarial attacks, generating imperceptible perturbed adversarial examples that cause misclassifications. And, it represents a significant risk for DNN-based applications that are security sensitive \cite{pr1, pr3}. Essentially, adversarial attacks aim to uncover DNNs' vulnerabilities and improve model resilience through defense strategies.  

\begin{figure*}[!t]
    \centering
    \includegraphics[width=0.31\linewidth]{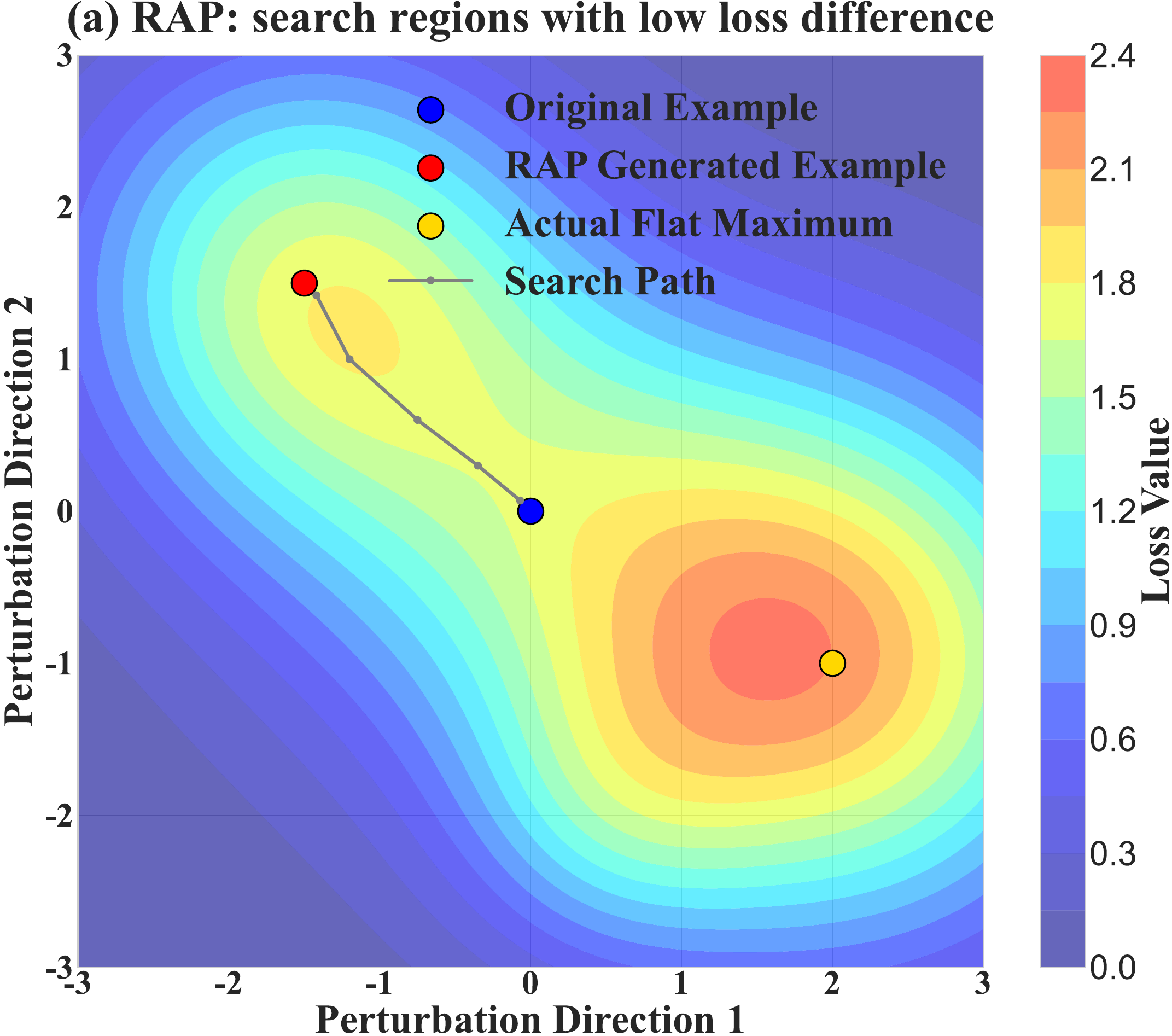}
    \hspace{-0.1mm}
    \includegraphics[width=0.31\linewidth]{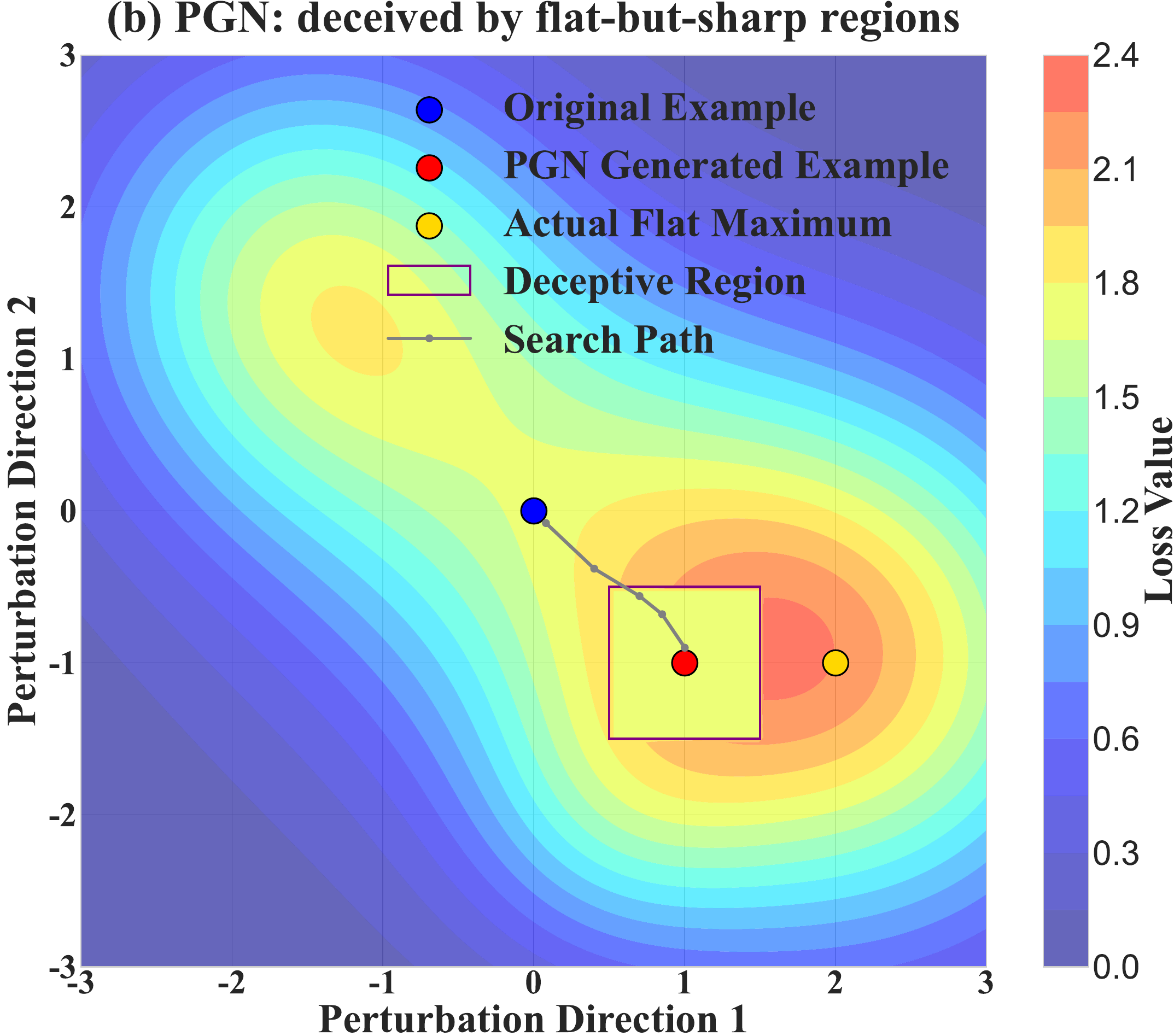}
    \vskip 1pt
    \centering
    \includegraphics[width=0.31\linewidth]{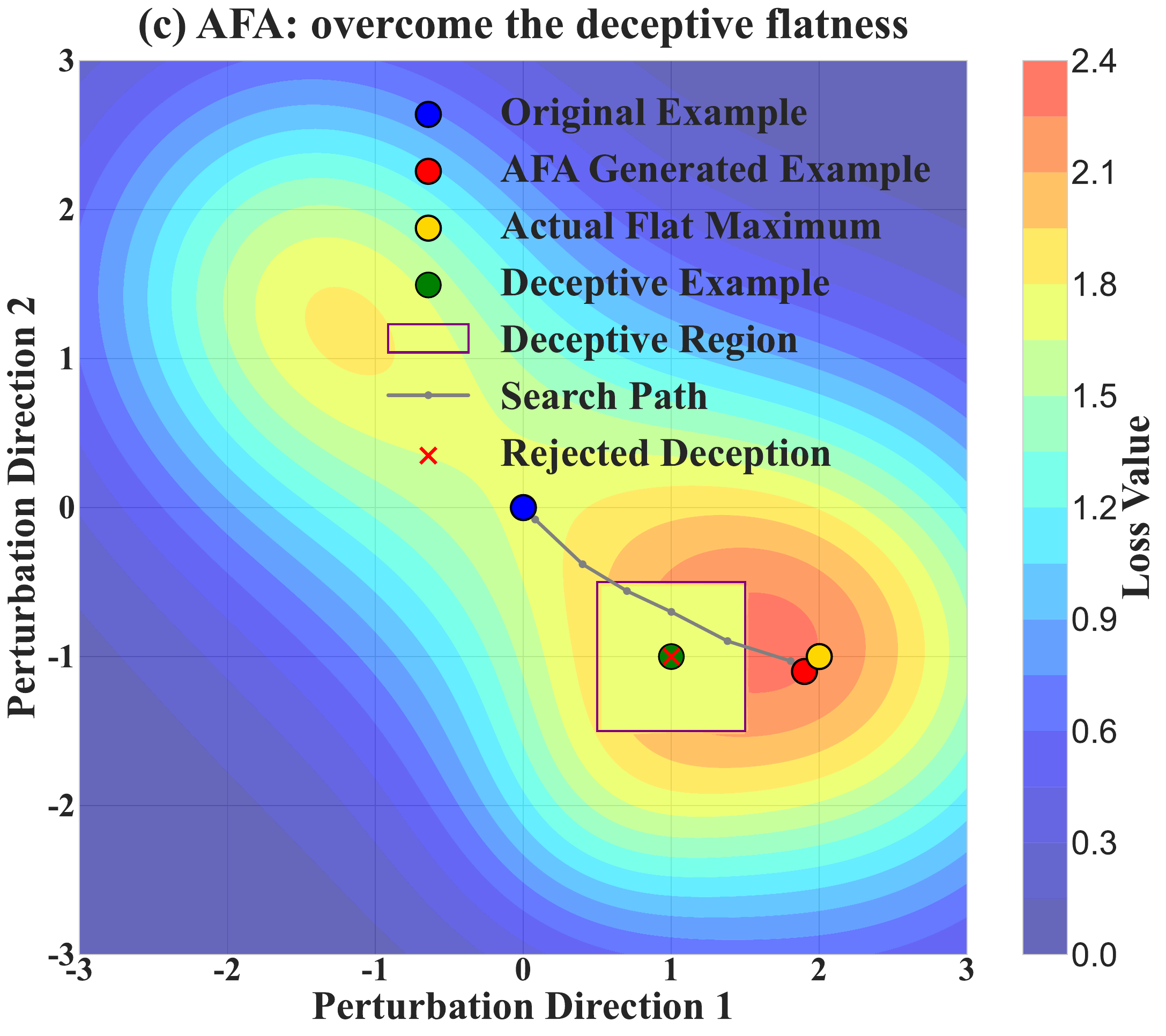}
    \hspace{-0.1mm}
    \includegraphics[width=0.31\linewidth]{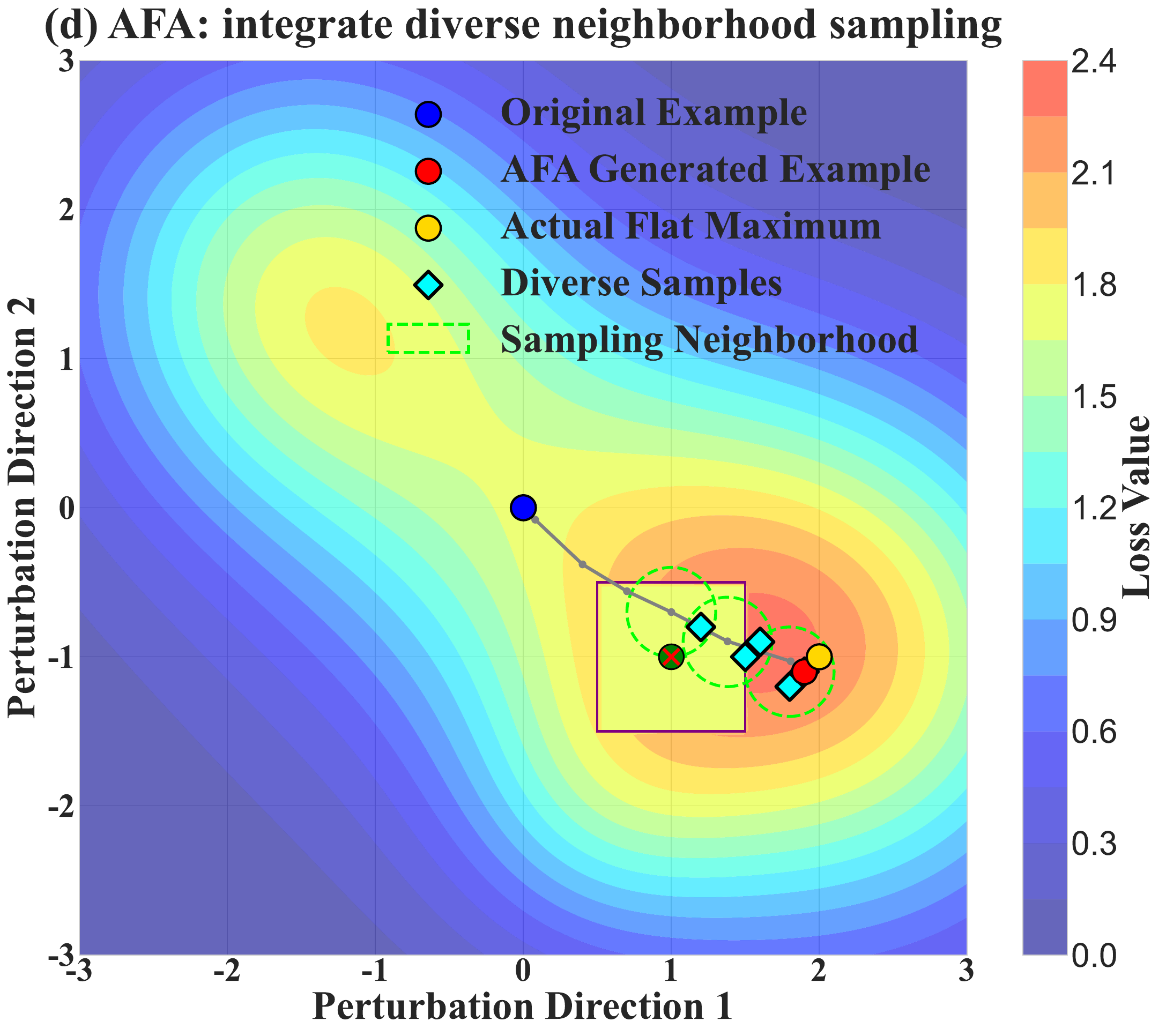}
    
    \caption{An example showing: (a)(b) the limitations of single-order flatness methods, and (c)(d) the improvements from our dual-order gradient fusion and diversified sampling strategy.}
\label{example_both_AZF_AFF}
\end{figure*}

Depending on whether the attacker has full access to the target model's internal information, adversarial attacks can be categorized as white-box attacks and black-box attacks. The former \cite{PGD, BIM}, with complete access to the target model, often achieves remarkable Attack Success Rates (ASRs). The impracticality of these methods in real-world scenarios stems from the privacy and security of target models. On the other hand, the effectiveness of the latter is restricted, since they have much lower ASRs than white-box attacks. Recently, several techniques have been suggested to improve the inferior results in black-box scenarios, using gradient estimation \cite{pr2}, input transformation \cite{DIM, SIM, TIM, Admix, SSA} and model ensembles \cite{MI}. While recent works \cite{RAP, NCS} improve attack capability by minimizing the loss difference within a gradient neighborhood (zeroth-order flatness), they may overlook truly flat local regions when the neighborhood radius is too small. To address this problem, another study \cite{PGN} investigates stable neighboring gradients (first-order flatness), but it still encounters the issue of flat-yet-sharp regions, termed deceptive flatness, as illustrated in Fig. \ref{example_both_AZF_AFF}. Notably, the flat-yet-sharp region typically arises from the unstable training process of the selected surrogate models, which may be attributed to the mathematical properties of the loss function itself (such as saturation regions). This identified misleading area could lead to increased computational costs and overfitting of the surrogate model. This indicates that these techniques may not be adequate for effectively targeting other black-box models. 

In this study, to eliminate the impact of deceptive flatness on adversarial transferability, we propose a novel gradient-based black-box transferable attack technique from a perspective of dual-order information. First, based on the deceptive flatness issue we have identified, we propose an Adversarial Flatness (AF) to effectively integrate global zero-order gradient information to compensate for the limitations of first-order flatness. And, we provide a theoretical guarantee for its adversarial transferability. Subsequently, through the utilization of an efficient estimation of our objective, we are able to flexibly implement our method as the Adversarial Flatness Attack (AFA). Meanwhile, the AFA tackles the changed gradient sign that may occur during the iterative process. Additionally, to further strengthen our attack method, we also design MonteCarlo Adversarial Sampling (MCAS) by diversifying the inner sampling. Finally, extensive experiments on ImageNet-compatible dataset against six representative baselines demonstrate the superiority of the proposed method in various model settings. Additionally, the validation on the Baidu Cloud API also verifies our performance. In particular, following the integration of various input transformation attacks, our approach not only significantly enhances their effectiveness but also surpasses other baseline methods.

Formally, our main contributions are summarized as follows:
\begin{itemize}
\item Motivated by the issues of deceptive flatness in adversarial attacks, we first propose Adversarial Flatness (AF) as a feasible solution and provide a theoretical demonstration of its assurance for adversarial transferability.
\item To reduce the complexity of implementation, we employ an efficient approximation of the attack objective and instantiate our method as Adversarial Flatness Attack (AFA), meanwhile solving the problem of the altered gradient sign. 
\item To improve the inner-loop sampling efficiency and further enhance adversarial transferability, we design MonteCarlo Adversarial Sampling (MCAS) to augment the variety of inner sampling.
\item The extensive experiments on ImageNet-compatible dataset exhibit our superior performance compared to six representative baselines. Then, the results on the Baidu Cloud API also validate the effectiveness of our method. Additionally,  together with integrating various input transformation attacks, our method not only boosts their effectiveness significantly but also surpasses other baselines.
\end{itemize}
The remaining sections of the paper are structured as follows. Section 2 provides background information and discusses related works. The proposed method is presented in Section 3. In Section 4, the evaluation results demonstrate that our method surpasses the competitors. Finally, conclusions and discussions are drawn in Section 5.

\section{Related Work}
\subsection{Preliminary}
 The objective of adversarial attacks is to generate an adversarial example, leading to misclassifications of the target model. Given an input example $x$, any model $\boldsymbol{M}$, true label $y^{gt}$, loss function $\mathcal{L}$ (i.e., Cross-Entropy Loss) and adversarial loss $\mathcal{L}^{adv}$, the adversarial example $x^{adv}$ in white-box untargeted setting can be expressed as:
\begin{align}
\begin{split}
    x_{adv} &= x + \min\limits_{\|\sigma\|_p\leq\eta}\left(\mathcal{L}^{adv}(\boldsymbol{M}(x+\sigma),y^{gt})\right)\\
    &= x + \min\limits_{\|\sigma\|_p\leq\eta}\left(-\mathcal{L}(\boldsymbol{M}(x+\sigma),y^{gt})\right),
    \end{split}
    \label{untarget_adv}
\end{align}
where $\sigma$ is the adversarial perturbation, $\|.\|_p$ calculates $p$ norm and $\eta$ means perturbation constraint. It should be noted that our work focuses on the untargeted transferable adversarial attack. Additionally, adversarial attacks can be classified into digital and physical attacks based on their mode of implementation. Physical attacks have advanced in certain critical areas, such as face recognition. For example, a recent method \cite{ProjAttacker} projects a light adversarial mask onto the face to deceive the recognition system. However, this paper focuses on digital attacks.

\subsection{Gradient-based Attacks}
In the early stage, Goodfellow et al. \cite{FGSM} propose Fast Gradient Sign Method (FGSM) by adding perturbation along the ascent direction of loss gradient $\bigtriangledown_x\mathcal{L}(\boldsymbol{M}(x), y^{gt})$ and generating adversarial examples as follows:
\begin{align}
\begin{split}
    x_{adv} &= x + \epsilon\cdot\text{sign}(\bigtriangledown_x\mathcal{L}(\boldsymbol{M}(x), y^{gt})).
    \end{split}
    \label{FGSM}
\end{align}

By applying the concept of Stochastic Gradient Descent (SGD) with momentum, Dong et al. \cite{MI} propose Momentum Iterative Fast Gradient Sign Method (MI-FGSM) to deviate from poorer local optima and improve the adversarial transferability. Then, Wang et al. \cite{VMI} reduce gradient variance between the adversarial example's gradient and the average gradient of its neighbors from the previous iteration, which is Variance-tuning Momentum Iterative Fast Gradient Sign Method (VMI-FGSM). Wang et al. \cite{GI-MI} introduce the Global-momentum-Initialization Momentum Iterative method (GI-MI), providing global momentum knowledge to mitigate gradient elimination. Wang et al. \cite{FGSRA} design the Frequency-Guided Sample Relevance Attack (FGSRA) to avoid the sharpness of decision boundaries in model-sensitive regions. Ren et al. \cite{MuMoDIG} propose the Multiple Monotonic Diversified Integrated Gradients (MuMoDIG) attack by focusing on the integration path of Integral Gradient and refining this path. In addition, inspired by flat local domain generalization, Qin et al. \cite{RAP} introduce Reverse Adversarial Perturbation (RAP), which first minimizes $\mathcal{L}$ near $x_{adv}$ before maximizing $\mathcal{L}$, equivalent to minimizing the maximum difference of $\mathcal{L}^{adv}$ on $x_{adv}$ and its vicinity. Similarly, Qiu et al. \cite{NCS} propose Neighborhood Conditional Sampling (NCS), to achieve the maximization of $\mathcal{L}(x_{adv})$ and the minimization of $\mathcal{L}$ on any neighborhood of $x_{adv}$. Differently, Ge et al. \cite{PGN} devise Penalize Gradient Norm (PGN) to limit the gradient norm of $\mathcal{L}$ and aim to identify the plateau local maxima.

Currently, there is no solution to the issue of deceptive flatness, nor is there any theoretical evidence concerning the effect of such a solution on adversarial transferability. Our approach not only tackles this problem but also proposes a novel gradient-based black-box attack method that incorporates dual-order flatness optimization.

\subsection{Input transformation-based Attacks}
Xie et al. \cite{DIM} present Diverse Input Method (DIM), which involves applying random resizing and padding to inputs in every iteration, improving the transferable attack. Dong et al. \cite{TIM} convolve the gradients of the original images with a fixed Gaussian kernel to approximate the average gradient of a set of translated images, namely Translation-Invariant
Method (TIM). Inspired by the scale-invariant property in DNNs, Lin et al. \cite{SIM} propose Scale-Invariant Method (SIM), computing the mean gradient across images scaled by $1/2^i$. Building on Mixup \cite{mixup}, Wang et al. \cite{Admix} develop Admix, which modifies images by blending the initial input with small segments of images from other classes. Long et al. \cite{SSA} design Spectrum Simulation Attack (SSA) to enhance transferability by altering the input image in the frequency domain. Similarly, Qian et al. \cite{Mixed-Frequency-Input-attack} present Mixed-Frequency Inputs (MFI) to obtain a more stable gradient direction by aggregating the high-frequency components.

Although input transformation-based attacks help cut down on computational expenses and enhance transferability, their effectiveness is still constrained and frequently insufficient. Indeed, the input transformation-based attacks above can be combined with current gradient-based attacks. As a result, they can improve transferability and act as a strong basis for assessing the efficacy of our suggested approach.  

\subsection{Adversarial Defense Strategies}
To mitigate the harmful effects of adversarial examples and improve the robustness of the model, numerous studies have developed various defense strategies. Tram{\`{e}}r et al. \cite{adv-training_ens} enhance the robustness of the model by using adversarial examples generated from ensemble models. Zhou et al. \cite{Class-Flipping-adv-train} propose a class fairness-based adversarial training method, which uses the most misleading category as feedback to construct targeted adversarial training data. Wu et al. \cite{robust-Adversarial-Weight-Perturbation} devise Adversarial Weight Perturbation (AWP) to explicitly regularize the flatness of weight loss landscape, forming a double-perturbation mechanism in the adversarial training framework. Random Smoothing (RS), as proposed by Cohen et al. \cite{RS}, treats adversarial perturbations as regular noise and enhances robustness by incorporating random elements. To reduce error amplification in traditional denoisers, Liao et al. \cite{HGD} construct High-level-representation Guided Denoiser (HGD), which utilizes the output-layer loss between the original and adversarial images. Xie et al. \cite{FD} design a network for Feature Denoising (FD) to suppress features in semantically irrelevant regions. Naseer et al. \cite{NPR} present Neural Representation Purifier (NRP) in a self-supervised fashion to eliminate adversarial perturbation. Jia et al. \cite{comdefend} introduce a compression-reconstruction network that removes adversarial perturbations but preserves nearby pixel correlations. Guo et al. \cite{jpeg} employ JPEG compression on input images to remove adversarial perturbations before classification. Xu et al. \cite{bit-red} propose feature squeezing through pixel color Bit-depth Reduction (Bit-Red) to detect adversarial examples. Xie et al. \cite{R&P} suggest using random resizing and random padding (R\&P) to reduce the effects of adversarial examples. NIPS-r3 is similar to NRP \cite{NPR}. Zhou et al. \cite{zhou2025defending} defend against adversarial examples by modeling adversarial noise based on the learned transition relationship between adversarial and true labels.

Usually, assessing the attack capability through the lens of defense strategies can be seen as a more reliable gauge from the viewpoint of opponents, suggesting a higher level of difficulty. Hence, in order to thoroughly showcase the efficacy of our method, we will undertake a comparative analysis between our approach and other baseline methods under the defense strategies mentioned earlier. 

\subsection{Flat Local Domain Generalization}
Currently, domain generalization has been a more challenging problem than domain adaptation. Among the recent works, Foret et al. \cite{SAM} present Sharpness-Aware Minimization (SAM) to improve model generalization by simultaneously minimizing the loss and sharpness, and SAM can seek parameters in regions that consistently yield low loss values. Zhang et al. leverage first-order optimization \cite{GAM} or combine them with zeroth-order techniques \cite{FAM} to enhance the domain generalization capability.  

\begin{figure}[]
\centering
\includegraphics[width=7.805cm]{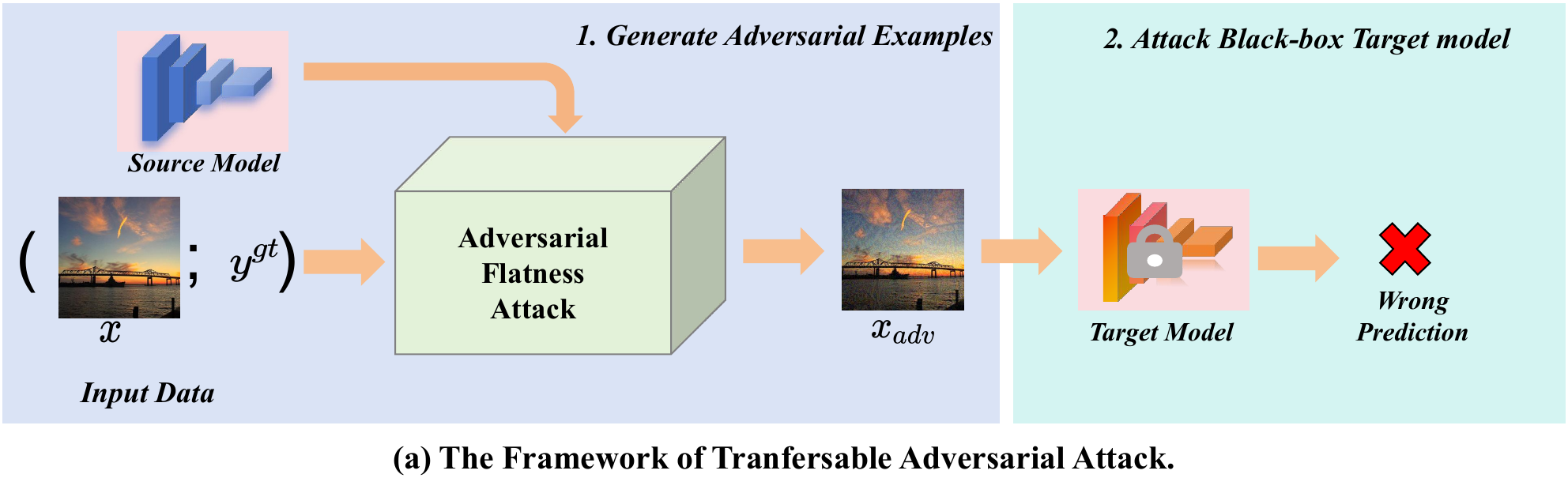}
\includegraphics[width=7.805cm]{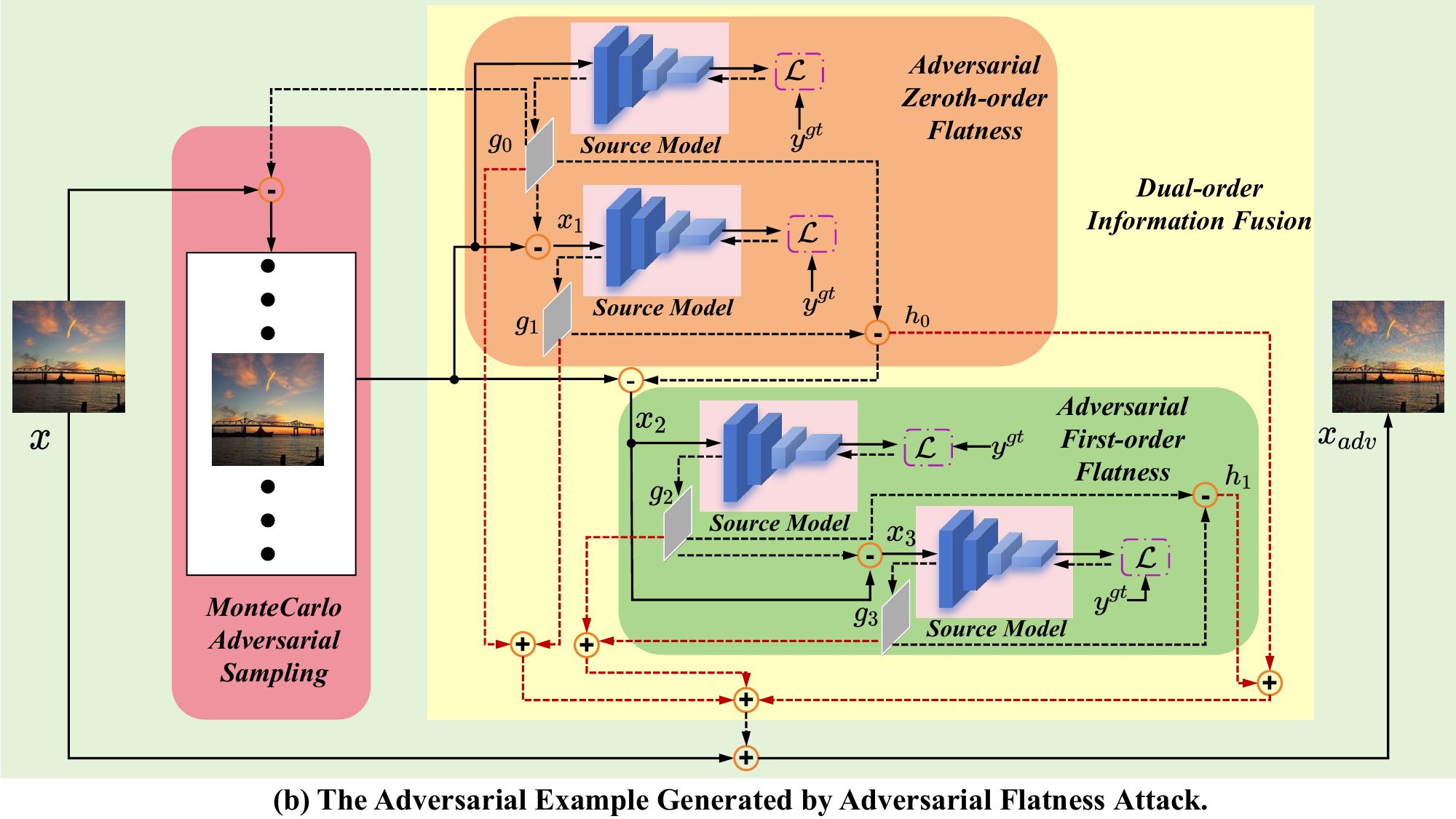}
\caption{(a) The process of the transferable adversarial attack. (b) The overview of generating the adversarial example by our proposed Adversarial Flatness Attack, corresponding to stage 1 in (a). In (b), our attack method consists of MonteCarlo Adversarial Sampling and Dual-order Information Fusion. The dashed red line denotes information fusion flow. The dashed black line indicates the backward gradient from the white-box source model.}
\label{fig: Framework}
\end{figure}

Notably, domain generalization and transferable attacks are two separate tasks. As discussed in \cite{SIM, VMI}, adversarial transferability can be analogized to model generalization, where the optimization of perturbations is likened to model training, and the implementation of transferable attacks on other black-box target models corresponds to testing procedures. Thus, our work can be naturally inspired by domain generalization.

\section{{Methodology}}
In this section, we first discuss the motivation of deceptive flatness, then introduce our dual-order solution with theoretical guarantees. We efficiently approximate the objective function and propose the Adversarial Flatness Attack (AFA) for black-box attacks. Finally, we present Monte Carlo Adversarial Sampling (MCAS) to further improve transferability. An overview is provided in Fig. \ref{fig: Framework}.

\subsection{Motivation}
Inspired by domain generalization \cite{SAM, GAM, FAM}, the recent works such as RAP \cite{RAP} and PGN \cite{PGN} are representative works that are respectively built upon low-loss differences and stable gradients around adversarial examples. It is worth noting that NCS \cite{NCS} bears resemblance to RAP \cite{RAP}. Subsequently, we formally elaborate on the mechanisms underlying \cite{RAP, PGN} as follows:

\begin{myDef}
	\label{AZF}
    (Adversarial Zeroth-order Flatness). For any radius $\xi>0$,  the adversarial zeroth-order flatness  $\Psi_0(x_{adv})$ of $\mathcal{L}_{src}^{adv}$ at a point $x_{adv}$ is formulated as
\begin{align}
   \begin{split}
   \Psi_0(x_{adv})&=\max\limits_{x_{adv}^{\prime}\in \mathcal{B}_{\xi}\left(x_{adv}\right)}\left(\mathcal{L}_{src}^{adv}\left(x_{adv}^{\prime}\right)-\mathcal{L}_{src}^{adv}\left(x_{adv}\right)\right), x_{adv}\in \mathcal{B}_{\epsilon}\left(x\right),
   \end{split}
\end{align}
where x is the original example of adversarial example $x_{adv}$, $\epsilon$ is the upper bound of the magnitude of perturbation on x, $\mathcal{L}_{src}^{adv}$ is adversarial loss of $x_{adv}$ on source model $\boldsymbol{M}_{src}$, and $\xi$ represents the upper bound of uniform sampling centered at adversarial example $x_{adv}$. $\mathcal{B}_{\epsilon}\left(x\right)$ indicates the neighborhood on point $x$ at radius of $\epsilon$. Similarly, $\mathcal{B}_{\xi}\left(x_{adv}\right)$ indicates the neighborhood on point $x_{adv}$ at radius of $\xi$.
\end{myDef}

\begin{myDef}
	\label{AFF}
    (Adversarial First-order Flatness). For any radius $\xi>0$,  the adversarial first-order flatness  $\Psi_1(x_{adv})$ of $\mathcal{L}_{src}^{adv}$ at a point $x_{adv}$ is formulated as
\begin{align}
   \begin{split}
   \Psi_1(x_{adv})&=\xi\cdot\max\limits_{x_{adv}^{\prime}\in \mathcal{B}_{\xi}\left(x_{adv}\right)}\left\|\bigtriangledown\mathcal{L}_{src}^{adv}\left(x_{adv}^{\prime}\right)\right\|_2, x_{adv}\in \mathcal{B}_{\epsilon}\left(x\right),
   \end{split}
\end{align}
where $\bigtriangledown\mathcal{L}_{src}^{adv}\left(x_{adv}^{\prime}\right)$ is the gradient of $\mathcal{L}_{src}^{adv}\left(x_{adv}^{\prime}\right)$ w.r.t. $x_{adv}^{\prime}$.
\end{myDef}

For simplified expression, we shorten Adversarial Zeroth-order Flatness and Adversarial First-order Flatness as AZF and AFF respectively.

Combined with the above definitions, we further analyze the limitations of \cite{RAP, PGN}. On the one hand, RAP \cite{RAP} assumes that an adversarial example will have a strong transferability if the loss difference within its vicinity is sufficiently small, falling under the category of AZF in Definition \ref{AZF}. However, as shown in Fig. \ref{example_both_AZF_AFF} (a), this method is prone to becoming trapped in a local suboptimal region if the search steps or radius are excessively small. On the other hand, PGN \cite{PGN} addresses this issue by generating adversarial examples in areas of the loss function with consistent gradients, known as AFF in Definition \ref{AFF}. Whereas, if there is a flat-yet-sharp area, a localized flat loss surface surrounding sharp losses near the current adversarial example, the AFF-based method (such as PGN \cite{PGN}) is prone to being fooled by this region, as illustrated in Fig. \ref{example_both_AZF_AFF} (b). Here, we refer to this shortcoming as \textbf{deceptive flatness}, leading to overfitting the surrogate model and degradation of adversarial transferability. In other words, we think the gradient flatness, which can get rid of the flat-yet-sharp areas, is important to adversarial transferability across diverse target models.

To this end, this paper focuses on addressing the issue of deceptive flatness at the gradient level and proposes a novel attack method to improve the effectiveness of attacks against other black-box models. 

\begin{figure}[htb]
\centering
\includegraphics[width=0.38\linewidth]{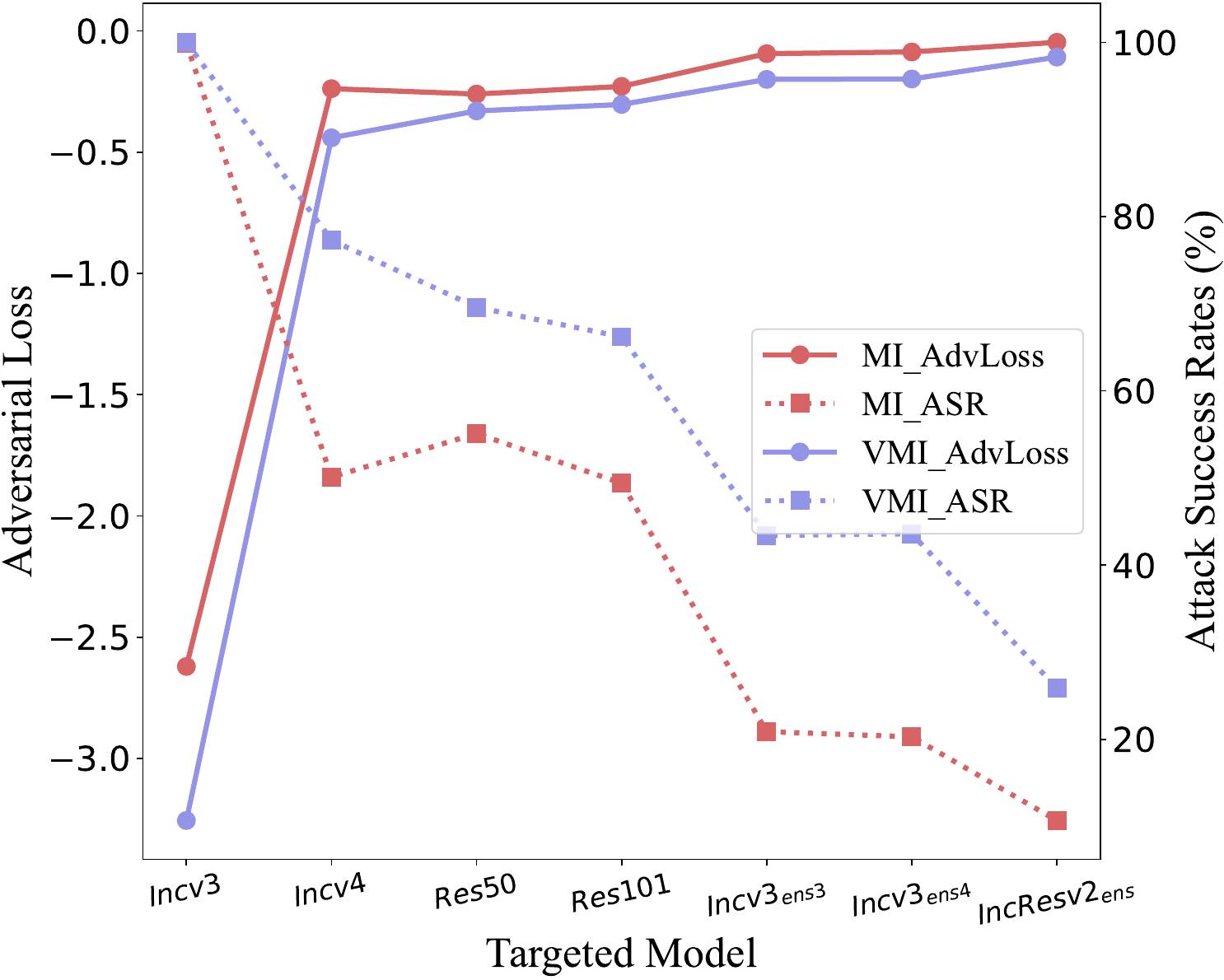}
\caption{The attack success rates (\%) and adversarial loss of MI and VMI on seven black-box models. The adversarial examples are generated on Inc-v3. The adversarial losses of MI-FGSM and VMI-FGSM are individually represented as MI\_AdvLoss and VMI\_AdvLoss. The attack success rates of MI-FGSM and VMI-FGSM are denoted as MI\_ASR and VMI\_ASR respectively.}
\label{fig_adv_loss}
\end{figure}

\subsection{Dual-order Solution}
\subsubsection{Adversarial Flatness}
As shown in Definition \ref{AZF}, in searching for an adversarial example in \cite{RAP}, AZF focuses more on the global loss information in its neighborhood. Meanwhile, AFF in Definition \ref{AFF} highlights the change of loss gradients in the neighborhood of the adversarial example. More precisely, AFF emphasizes the local information to some extent. Naturally, we suppose that fusion of the gradients information from AZF to AFF, can provide a global guidance to AFF, escaping from deceptive flatness, as expected in Fig. \ref{example_both_AZF_AFF} (c). Theoretically, these two fused gradient information, namely dual-order information fusion, can also help reduce overfitting of the adversarial examples into the surrogate model caused by deceptive flatness. However, there is currently a lack of literature exploring the dual-order information fusion for addressing deceptive flatness and its role in adversarial transferability.

Although adversarial transferability is analogous to model generalization, there remains a lack of concrete evidence demonstrating the connection between domain generalization and adversarial transferability. Therefore, a rigorous description is first given as follows:

\begin{assumption}
    Given white-box surrogate model $\boldsymbol{M}_{src}$, a series of black-box models $\boldsymbol{M}_{tar1}, \boldsymbol{M}_{tar2}, ..., \boldsymbol{M}_{tarn}$, and adversarial loss of $x_{adv}$ on model $\boldsymbol{M}_k$ represented as $\mathcal{L}_{k}^{adv} = \mathcal{L}_{k}^{adv}\left(x_{adv}\right)=\mathcal{L}^{adv}\left(x_{adv}, y^{gt}; \boldsymbol{M}_k\right)$, if $\mathcal{L}_{src}^{adv} \le \mathcal{L}_{tar1}^{adv} \le \mathcal{L}_{tar2}^{adv} \le ... \le \mathcal{L}_{tarn}^{adv}$, the attack transferability expressed as ASRs across surrogate and target models tends to satisfy the following inequality, i.e., $\textit{ASR}_{src} \ge \textit{ASR}_{tar1} \ge \textit{ASR}_{tar2} \ge ... \ge \textit{ASR}_{tarn}$. 
    \label{assumption1}
\end{assumption}

In Fig. \ref{fig_adv_loss}, when transferring adversarial example $x_{adv}$ generated on white-box Inc-v3 to other black-box targets, the higher ASRs correspond to the lower adversarial loss value. Such a remarkable phenomenon strongly supports Assumption \ref{assumption1}. This is the first instance of showing the numerical similarity between untargeted adversarial attacks and domain generalization, with a lower loss value usually indicating improved accuracy. Then, based on this observation and inspired by \cite{FAM} in domain generalization, we can feasibly propose our dual-order solution to deceptive flatness in our work, called Adversarial Flatness (AF), as below,
\begin{align}
   \begin{split}
   \Psi^{AF}(x_{adv})=\beta_f\Psi_0(x_{adv})+\left(1-\beta_f\right)\Psi_1(x_{adv}),
   \end{split}
   \label{AF}
\end{align}
where $\beta_f$ is the flatness balanced coefficient. After joining the basic adversarial objective as Eq. \ref{untarget_adv}, we derive our main objective as below.
\begin{align}
   \begin{split}
\mathcal{L}_{all}^{adv}=\mathcal{L}_{src}^{adv}+\lambda_f\Psi^{AF}(x_{adv}),
   \end{split}
   \label{overall_objective}
\end{align}
where $\lambda_f$ is the flatness item coefficient. Then, we explore the role of AF theoretically in adversarial transferability. 

\subsubsection{Theoretical Guarantees to Adversarial Transferability}
Given any adversarial example $x_{adv}$, the maximum radius of uniform sampling $\mathcal{U}$ centered at adversarial example $\xi$ and step size $\alpha$, we suppose that Eq. \ref{overall_objective} has a second-order gradient at least. According to Definition \ref{AZF} and Definition \ref{AFF}, $\vartheta\sim\mathcal{U}(-\xi, \xi)$, such that the adversarial loss on the surrogate model in the vicinity of $x_{adv}$ and AF can be related as below:
\begin{align}
\begin{split}
\mathcal{L}_{src}^{adv}\left(x_{adv}+\vartheta\right)
&\le\mathcal{L}_{src}^{adv}\left(x_{adv}\right)+\Psi^{AF}(x_{adv}).
\end{split}
   \label{R_inequal}
\end{align}
The detailed proof is provided in \ref{app1}. 

Finally, after integrating Eq. \ref{R_inequal} and the generalization boundary theory in theorem 4.2 \cite{nico++}, we obtain the expanded inequality between AF and adversarial transferability:

\begin{align}
\begin{split}
&\mathbb{E}_{x_{adv}^{\prime}\sim \mathcal{B}_{\xi}(x_{adv})}[\mathcal{L}_{tar}^{adv}\left(x_{adv}^{\prime}\right)]\le\mathcal{L}_{src}^{adv}\left(x_{adv}\right)+\Psi^{AF}(x_{adv})+BD_{upper},\\
&BD_{upper}= \sup\limits\|\mathcal{L}_{tar}(x^1, x^2)-\mathcal{L}_{src}(x^1, x^2) \|, x^1, x^2 \in \mathcal{B}_{\xi}(x_{adv}),
\end{split}
   \label{ATG}
\end{align}
where the term $\sup\limits\|\cdot\|$ represents the upper bound of the discrepancy distance \cite{covari_shift} between adversarial losses of $x_{adv}$ on the source model $\boldsymbol{M}_{src}$ and the target model $\boldsymbol{M}_{tar}$, mimicking the training domain and the unknown test domain in domain generalization. Similarly, $\mathcal{B}_{\xi}\left(x_{adv}\right)$ indicates the neighborhood on point $x_{adv}$ at radius of $\xi$. The detailed proof is provided in \ref{app2}. Clearly, the AF item $\Psi^{AF}(x_{adv})$ (the second term on the right side of the inequality) leads to better adversarial transferability on the black-box model $\boldsymbol{M}_{tar}$. To fully demonstrate the rationale of the proposed AF, we also conduct an additional visual experiment at \ref{Visual_AF}.

\subsection{Adversarial Flatness Attack}
Since our goal is to achieve a transferable untargeted attack, we can clarify our objective in Eq. \ref{overall_objective} as follows.
\begin{align}
    \begin{split}
        \max\limits_{x_{adv}\in \mathcal{B}_{\epsilon}(x)}[\mathcal{L}\left(x_{adv}, y^{gt}; \boldsymbol{M}_{src}\right)-\lambda_f\Psi^{AF}(x_{adv})],
    \end{split}
    \label{FAA}
\end{align}
where $\Psi^{AF}(x_{adv})$ includes the first-order gradient of $\ell_{src}^{tar}$. In addition, $\Psi^{AF}(x_{adv})$ is computationally expensive when directly deriving from the first-order gradient to the second-order gradient in generating the adversarial example $x_{adv}$. Hence, we approximate the second-order gradient using the first-order gradients. And similar to \cite{PGN}, we also take the search step $\triangle x = \alpha\cdot\frac{-\bigtriangledown_{x}\mathcal{L}(x,y^{gt};\boldsymbol{M}_{src})}{\|\bigtriangledown_{x}\mathcal{L}(x,y^{gt};\boldsymbol{M}_{src})\|_1}$ within the neighborhood size $\xi$, where $\|\cdot\|$ is the $L_1$ norm. Then the gradient of Eq. \ref{FAA} is approximated as
follows. Several gradient approximation methods exist \cite{survey_aprox}, and we have selected the Finite Difference Method because of its simplicity. Additional experiments comparing our method with those based on other gradient approximation techniques are presented in \ref{experiment_gradient_approx}.
\begin{figure}[!t]
\centering
    \includegraphics[width=0.35\linewidth]{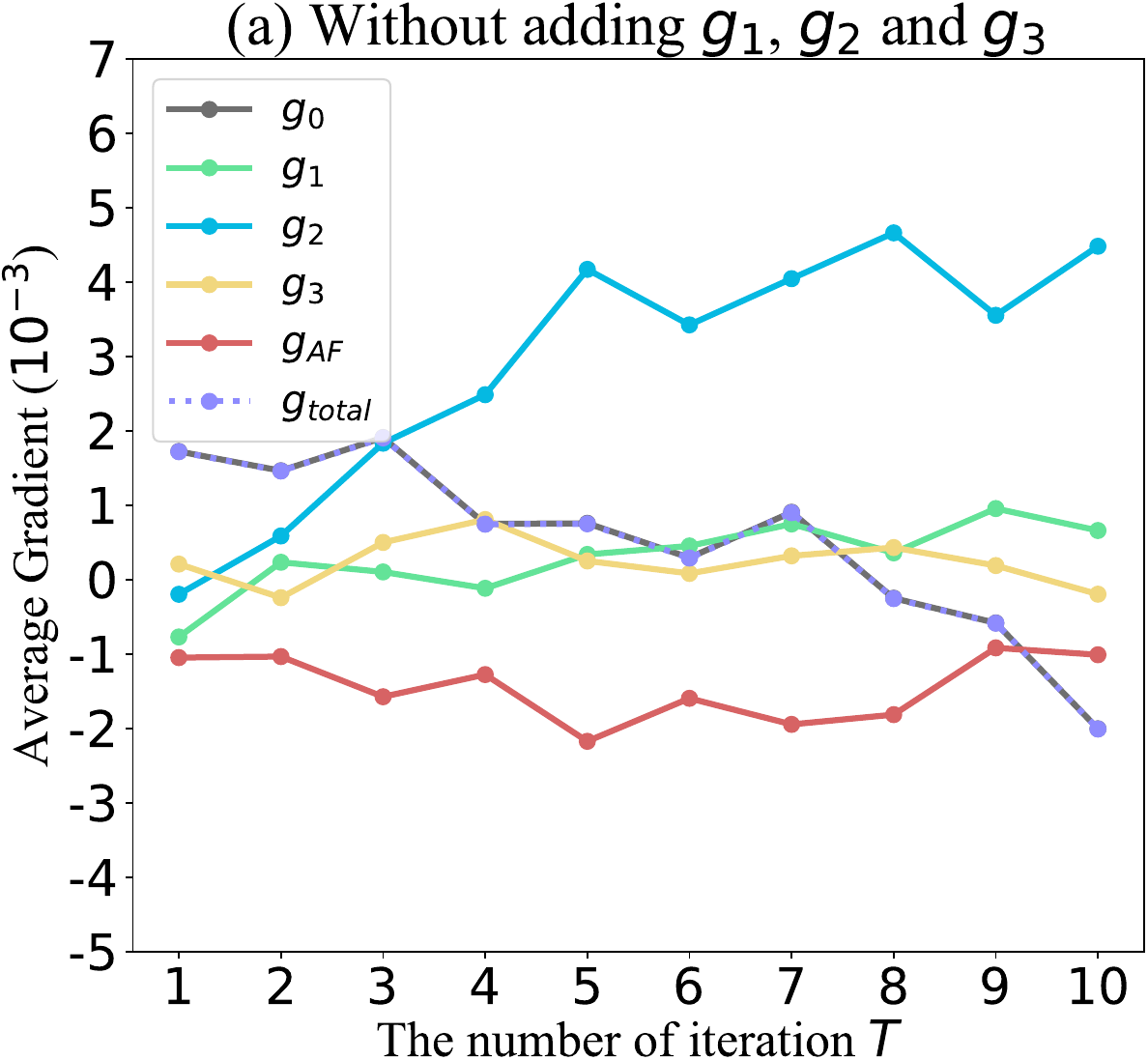}
    \hspace{2mm}
    \includegraphics[width=0.35\linewidth]{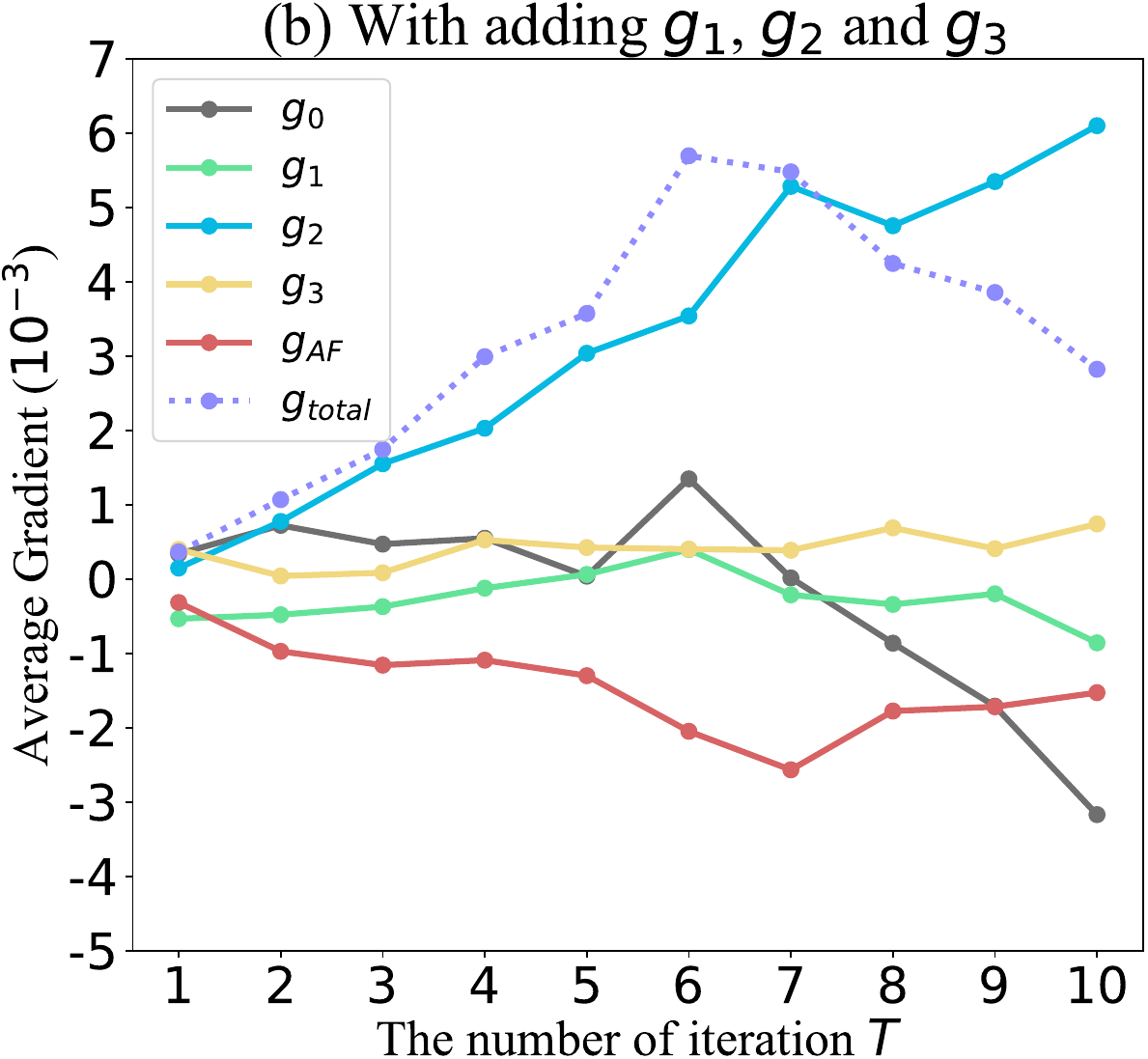}
    \caption{Illustrations of the average gradients of our approximated objective over $T$ iterations (a) without and (b) with adding $g_1$, $g_2$ and $g_3$. The dashed line representing $g_{total}$ is our focus.}
    \label{with_without_g123}
\end{figure}

\subsubsection{The Approximation of Gradients on $\Psi_0(x_{adv})$ and $\Psi_1(x_{adv})$} Let $x_0=x_{adv}$, $x_1 = x_0+\triangle x_0=x_0-\alpha\cdot\frac{\bigtriangledown_{x_0}\mathcal{L}(x_0,y^{gt};\boldsymbol{M}_{src})}{\|\bigtriangledown_{x_0}\mathcal{L}(x_0,y^{gt};\boldsymbol{M}_{src})\|_1}$. We can approximate the gradient of $\Psi_0(x_{adv})$ as follows:
\begin{align}
    \begin{split}
        &\bigtriangledown\Psi_0(x_{adv})=h_0\approx g_1^{\prime}-g_0^{\prime} =-(g_1-g_0), \text{    where  }\\
        &g_0^{\prime} = \bigtriangledown_{x_0}\mathcal{L}^{adv}_{src}(x_0)=-g_0=-\bigtriangledown_{x_0}\mathcal{L}(x_0,y^{gt};\boldsymbol{M}_{src}), \\
        &g_1^{\prime} = \bigtriangledown_{x_1}\mathcal{L}^{adv}_{src}(x_1)=-g_1=-\bigtriangledown_{x_1}\mathcal{L}(x_1,y^{gt};\boldsymbol{M}_{src}).
    \end{split}
    \label{h_0}
\end{align}

Here we use the smaller search radius $\alpha$ than $\xi$, so we reformulate gradient of $\Psi_1(x_{adv})$ as the following procedure:
\begin{align}
    \begin{split}
        &\bigtriangledown\Psi_1(x_{adv})=h_1\approx -(g_3-g_2), \text{where  }\\
     &g_2=\bigtriangledown_{x_2}\mathcal{L}(x_2,y^{gt};\boldsymbol{M}_{src}),\\
     &g_3=\bigtriangledown_{x_3}\mathcal{L}(x_3,y^{gt};\boldsymbol{M}_{src}),\\
&x_2=x_0-\alpha\cdot\frac{g_1-g_0}{\|g_1-g_0\|},\\
&x_3=x_2-\alpha\cdot\frac{g_2}{\|g_2\|}.
    \end{split}
    \label{h_1}
\end{align}
The detailed proof is provided in \ref{app3}. 

Next, after joining Eq. \ref{h_0} and Eq. \ref{h_1}, our objective in Eq. \ref{FAA} can be rewritten as:
\begin{align}
    \begin{split}
        \max\limits_{x_{adv}\in \mathcal{B}_{\epsilon}(x)}[g_0+\lambda_f(\beta_f(g_1-g_0)+(1-\beta_f)(g_3-g_2))].\\
    \end{split}
    \label{FAA_2}
\end{align}
The detailed proof is provided in \ref{app4}. 

\begin{figure}[htb]
\centering
\includegraphics[width=0.35\linewidth]{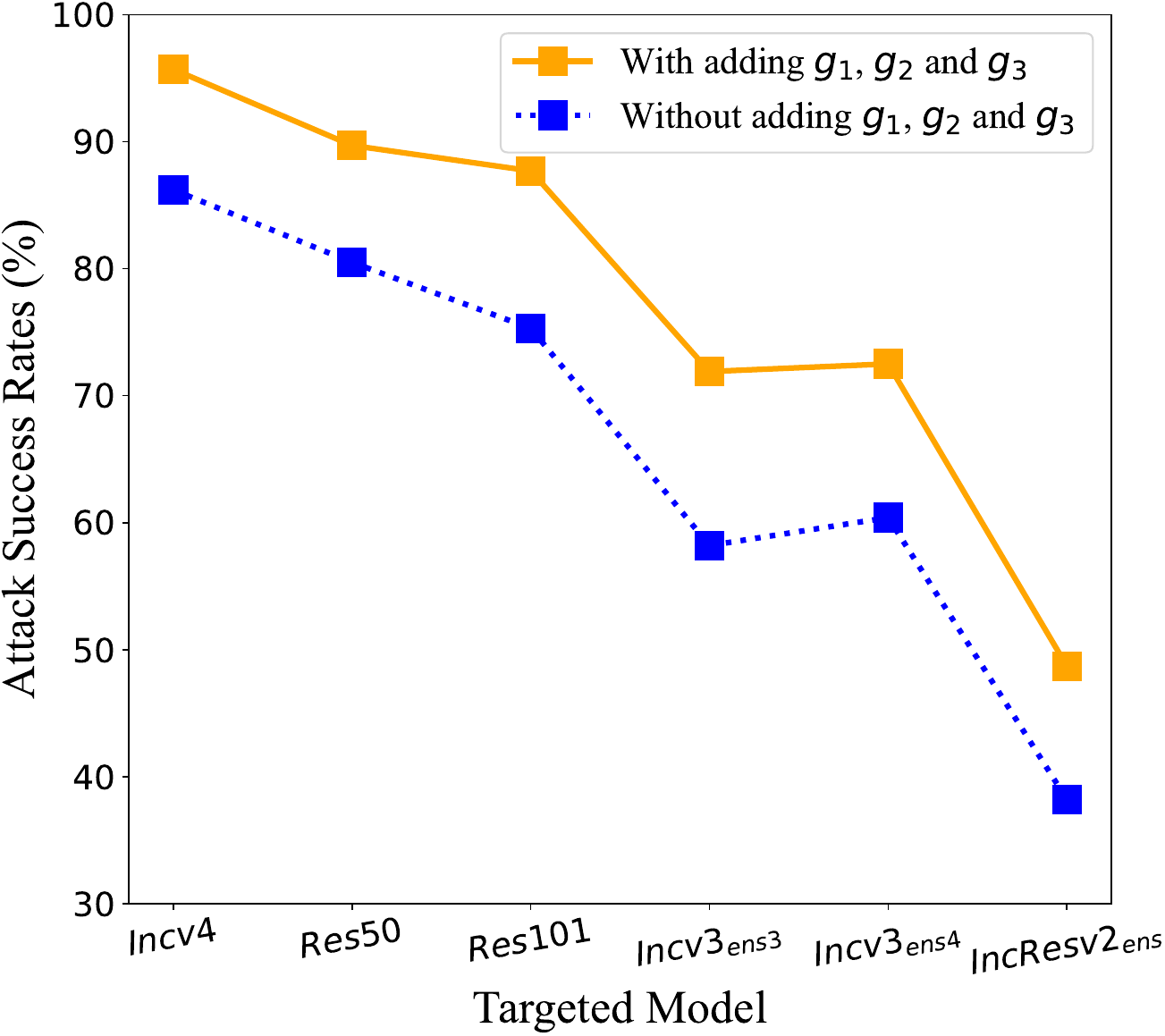}
\caption{The attack success rates (\%) of our AFA without or with adding $g_1$, $g_2$ and $g_3$ on six black-box models. The adversarial examples are generated on Inc-v3. Incv3$_{ens3}$, Incv3$_{ens4}$ and IncResv2$_{ens}$ are Inception-v3 and Inception-ResNet-v2 using ensemble adversarial training \cite{adv-training_ens}.}
\label{transferability_wo_g123}
\end{figure}

\subsubsection{Problem of Altered Gradient Sign}
However, our objective in Eq. \ref{FAA_2} represented as $g_{total}$ considers both $g_0$ and the other gradients of the three neighborhoods ($x_1$, $x_2$ and $x_3$), which means an intricate optimization process. Besides, it may lead to the unexpected direction alteration with $g_{total}<0$ after $T>7$, as illustrated in Fig. \ref{with_without_g123} (a). Since our objective is to find the adversarial example in the direction of gradient accent, this phenomenon can lead to a significant decline in transferable attack effectiveness. Moreover, based on our intuition, we can optimize $g_1$, $g_2$ and $g_3$ along the ascent direction simultaneously, as evidenced in Fig. \ref{with_without_g123} (b). Formally, we update Eq. \ref{FAA_2} explicitly for our objective as follows: 
\begin{align}
    \begin{split}
       \max\limits_{x_{adv}\in \mathcal{B}_{\epsilon}(x)}[&g_0+\lambda_f(\beta_f(g_1-g_0)+(1-\beta_f)(g_3-g_2))+g_1+g_2+g_3
        ].\\
    \end{split}
    \label{FAA_final}
\end{align}

In reality, it can be easily integrated with iterative attacks, aligned with existing works \cite{RAP, NCS, PGN, VMI}. Here, based on Eq. \ref{FAA_final} and MI-FGSM \cite{MI}, we formulate our Adversarial Flatness Attack (AFA). Hence, Algorithm \ref{alg:alg1} contains all the specifics of our AFA. This subsection mainly reflects on the rows except for the fourth, seventh and sixteenth.

In Fig. \ref{transferability_wo_g123}, it is evident that our AFA based on the enhanced objective in Eq. \ref{FAA_final} can achieve much better attack performance, compared to that based only on Eq. \ref{FAA_2}. It once confirms our assumption that the altered gradient sign of our objective can reduce the effectiveness of our attack, especially when the optimization process includes different gradients from multiple neighborhood examples. By explicitly maximizing the gradients in the vicinity of adversarial examples, the issue of the objective's gradient sign switching can be effectively reduced, preventing the decay of attack ability. Also, it can be shown in Fig. \ref{fig: Framework} (b).

\begin{algorithm}
\KwIn{A clean image $x$ with ground-truth label $y^{gt}$; the loss function $\mathcal{L}$ with source model $\boldsymbol{M}_{src}$; the magnitude of perturbation $\epsilon$; the maximum iterations number $T$; the momentum decay $\eta$; the sampling number $N$; the upper bound of neighborhood sampling $\xi$; the radius of MonteCarlo Adversarial Sampling $\gamma_{MCAS}$; the momentum decay of MonteCarlo Adversarial Sampling $\eta_{MCAS}$; the flatness balanced coefficient $\beta_{f}$; the flatness item coefficient $\lambda_{f}$.}      
\KwOut {An adversarial example $x_{adv}$.}       
\BlankLine
$m_0=0$, $x_{adv}^0=x$, $\alpha = \frac{\epsilon}{T}$\;

\For{$t\leftarrow 0$ \KwTo $T$-1}{
$\overline{g}=0$\; 

$\overline{g_{s}}=0$

\For{$i\leftarrow 0$ \KwTo $N$-1}{
\text{Randomly sample} $x^{\prime}=x+\vartheta$, $\vartheta\sim\mathcal{U}(-\xi, \xi)$\;

\text{MonteCarlo Adversarial Sampling} $x_0=x^{\prime}-\gamma_{MCAS}\cdot \text{sign}(\overline{g_{s}})$\;

\text{Calculate the gradient} $g_0=\bigtriangledown_{x_0}\mathcal{L}(x_0)$ \;

\text{Update sample} $x_1 = x_0-\alpha\cdot\frac{g_0}{\|g_0\|_1}$ \;

\text{Calculate the gradient} $g_1=\bigtriangledown_{x_1}\mathcal{L}(x_1)$ \;

\text{Update sample} $x_2 = x_0-\alpha\cdot\frac{g_1-g_0}{\|g_1-g_0\|_1}$ \;

\text{Calculate the gradient} $g_2=\bigtriangledown_{x_2}\mathcal{L}(x_2)$ \;

 \text{Update sample} $x_3 = x_2-\alpha\cdot\frac{g_2}{\|g_2\|_1}$ \;
 
\text{Calculate the gradient} $g_3=\bigtriangledown_{x_3}\mathcal{L}(x_3)$ \;

\text{Accumulate the gradient} $\overline{g} += \frac{g_0+\lambda_f(\beta_f(g_1-g_0)+(1-\beta_f)(g_3-g_2))+g_1+g_2+g_3}{N}$ \;

\text{Update MCAS momentum} $\overline{g_s}=\eta_{MCAS}\cdot\overline{g_s}-g_0$\;

}
$m_{g+1}=\eta\cdot m_t+\frac{\overline{g}}{\|\overline{g}\|_1}$\;

$x_{adv}^{t+1}=\Pi_{\mathcal{B}_{\epsilon}(x)}[x_{adv}^t+\alpha\cdot\text{sign}(m_{g+1})]$\;

}
$x_{adv}=x_{adv}^T$\;
\caption{Adversarial Flatness Attack.}
\label{alg:alg1}
\end{algorithm}

\begin{figure}[htb]
\centering
\includegraphics[width=0.35\linewidth]{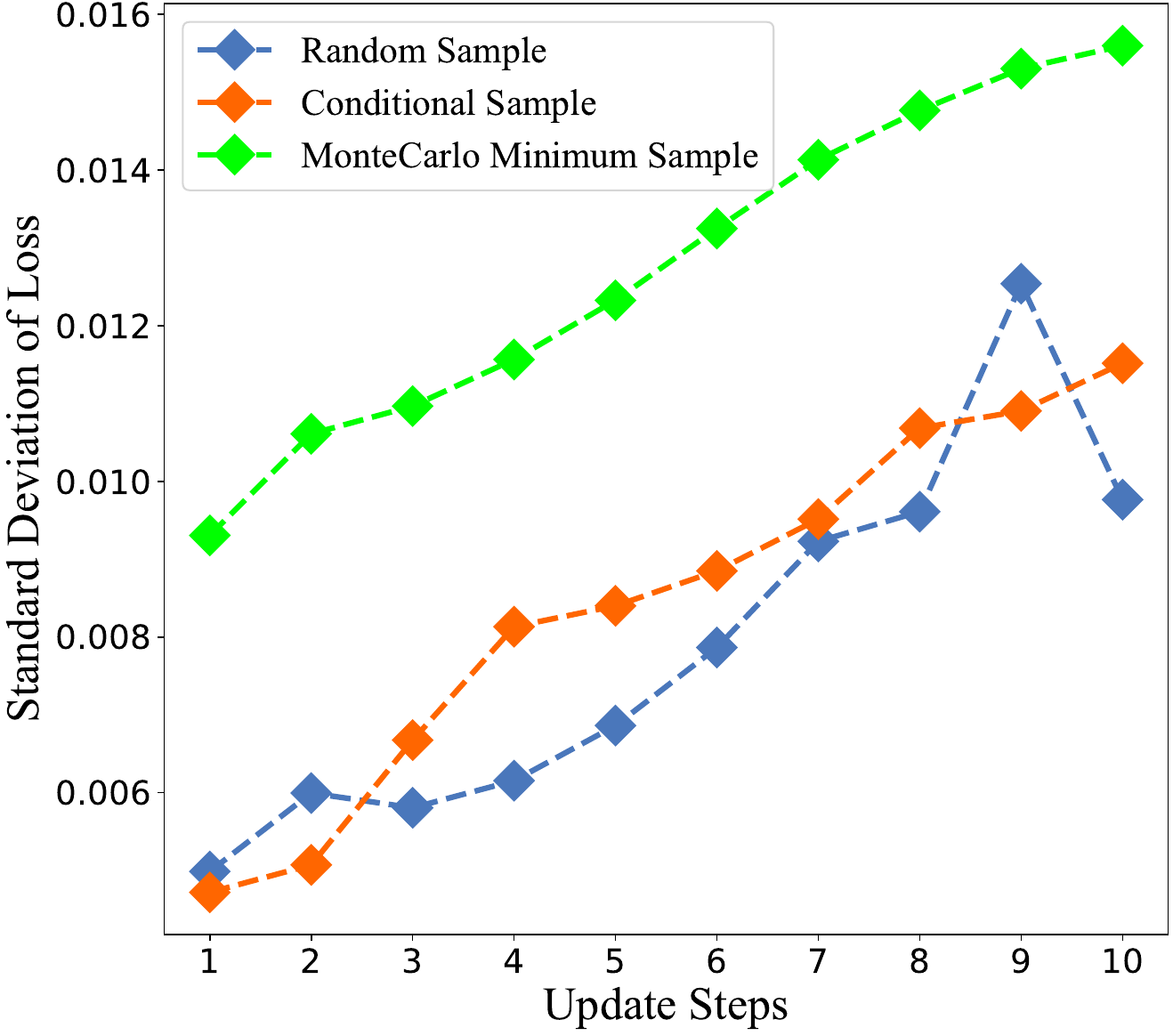}
\caption{Comparison of average loss standard deviation for every iteration $t$. The update steps are equivalent to the maximum iteration $T=10$.}
\label{sample_comparison}
\end{figure}

\subsection{MonteCarlo Adversarial Sampling}
As mentioned before, the sampling technique for the attack effectiveness is crucial. Naturally, we expect it to efficiently enhance our AFA, as shown in Fig. \ref{example_both_AZF_AFF} (d). Nevertheless, the above works largely overlook this aspect, where they merely use a uniform sampling around $x_{adv}^t$. 

Recently, Qiu et. al. \cite{NCS} have for the first time explained that diversifying the surrounding points centered on $x_{adv}$ can help perturbation optimization avoid suboptimal outcomes. They suggest utilizing the gradient information of the previous two sampling processes. Rather than concentrating on increasing diversity in sampling, we prioritize fixing its two weaknesses. Increasing the iteration number $T$ and the inner sampling number $N$ will require more hardware memory and computational resources, which is not feasible. However, conditional sampling only avoids repeating the same sampling as the previous two iterations, without ensuring dissimilarity with the previous ones.

Based on MonteCarlo estimation in reinforcement learning \cite{MCRL}, we introduce MonteCarlo Adversarial Sampling (MCAS) to broaden the surroundings of $x_{adv}^t$, which strategically eliminates the most recent sampling information (gradients) in a momentum manner, as shown in Fig. \ref{fig: Framework} (b). Mathematically, the proposed MCAS is formulated as:
\begin{align}
    \begin{split}
        &x_{0}=x_{adv}^{t}-\gamma_{MCAS}\cdot \text{sign}(\overline{g_{s}}), \text{where } \overline{g_s}=\eta_{MCAS}\cdot\overline{g_s}-\bigtriangledown_{x_0}\mathcal{L}(x_0),
    \end{split}
    \label{MCAS}
\end{align}
where $\overline{g_{s}}=0$ when $t=0$. Besides, $\gamma_{MCAS}$ and $\eta_{MCAS}$ indicate the radius and the momentum decay of MCAS, respectively. To showcase the diversity of our MCAS compared to conditional sampling and random sampling, we calculate the average loss standard deviation for each iteration $t$. To initially validate the effectiveness of MCAS, we conduct a toy experiment, where random sampling, conditional sampling and our MCAS are individually equipped with MI-FGSM \cite{MI}. As shown in Fig. \ref{sample_comparison}, among ten update steps, the toy example demonstrates that our MCAS can achieve a higher average standard deviation of loss. It indicates our method can produce more diverse samples in the vicinity of $x_{adv}^t$. Additionally, when generating an adversarial example, our MCAS requires 412 MB less memory than NCS and has an inference time 0.4s shorter than NCS. Furthermore, a more detailed discussion of our method is provided in \ref{computational efficiency}. Finally, our proposed MCAS is allocated to the fourth, seventh and sixteenth rows as specified in Algorithm \ref{alg:alg1}.

\section{Experiments}
\label{experiment}
\noindent {\bf{Datasets.}} Our validation is conducted on ImageNet-compatible dataset from the NIPS 2017 adversarial competition, commonly used in earlier studies. It includes 1,000 images sized 299 $\times$ 299 $\times$ 3, along with ground-truth labels and target labels for targeted attacks. It notes that our AFA focuses on transferable untargeted attacks, disregarding the target label in the dataset.

\noindent {\bf{Models.}} To showcase the efficacy of our proposed methods, we evaluate the attack performance on both normally trained models and robust models. For normally trained models, we use seven classical Convolutional Neural Networks (CNNs), including Inception-v3 (Inc-v3), Inception-v4 (Inc-v4), Inception-ResNet-v2 (IncRes-v2), ResNet-50 (Res-50), ResNet-101 (Res-101), DenseNet-121 (Dense-121) and VGG-19bn (VGG-19) \cite{cnn_survey}. Aside from the CNNs above, we also test on models with more diverse architectures: MobileNet-v2 (MobileNet) \cite{cnn_survey}, PASNet-5-Large (PASNet-L) \cite{PASNet-L}, ViT-Based/16 (ViT-B/16), PiT-S, Swin-T, MLP-mixer and ResMLP \cite{Transformer}. For robust models, we employ methods such as adversarial training, random smoothing, active defense and more. Precisely, we take Inc-v3$^*$, Inc-v3$_{ens}$, Inc-v4$_{ens}$ and IncRes-v2$_{ens}$ \cite{adv-training_ens} as adversarially trained models. Other defense strategies under attack contain RS \cite{RS}, HGD \cite{HGD}, FD \cite{FD}, NRP \cite{NPR}, ComDefend \cite{comdefend}, JPEG \cite{jpeg}, Bit-Red \cite{bit-red}, R\&P \cite{R&P} and NIPS-r3 \cite{NPR}. 

\noindent {\bf{Baselines.}} Our baselines consist of six gradient-based iterative adversarial attacks, including GI-MI \cite{GI-MI}, FGSRA \cite{FGSRA}, MuMoDIG \cite{MuMoDIG}, RAP \cite{RAP}, PGN \cite{PGN} and NCS \cite{NCS}. Additionally, we test the effectiveness of our AFA by incorporating various input transformations such as DIM \cite{DIM}, TIM \cite{TIM}, SIM \cite{SIM}, Admix \cite{Admix}, and SSA \cite{SSA}.

\noindent {\bf{Implementation details.}} Our method is applied using PyTorch on a NVIDIA GeForce GTX 3090 GPUs. We typically choose a maximum perturbation as $\epsilon = 16.0/255$ with iterations $T = 10$, a step size of $\alpha=\epsilon/T$ and a momentum decay of $\eta =1.0$. For GI-MI, we set the number of sampled examples $N = 20$ and the pre-attack epochs $P = 5$. For RAP, we set the step size $\alpha = 2.0/255$, the number of iterations $K = 400$, the inner iteration number $T = 10$, the late-start $K_{LS} = 100$, the size of neighborhoods $\epsilon_{n} = 16.0/255$. It should be noted that RAP in this paper is equipped with momentum decay for a fair comparison. For PGN, we set the number of sampling $N = 20$, the balanced coefficient $\delta = 0.5$, and the upper bound of $\xi = 3.0 \times \epsilon$. For NCS, we set the
upper bound of neighborhood sampling $\xi = 2 \times \epsilon$, the number of sampling $N = 20$, the upper bound of sub-regions $\gamma = 0.15 \times \epsilon$, and the balanced coefficient $\lambda = \alpha/T$. For MuMoDIG, the epoch $T$ and the inner interpolation times $N_i$ are set to 10 and 20, respectively, while the transformation number $N_{trans}$ is set to 0 to ensure a fair comparison. It should be noted that the baselines are initially excluded from all integrated transformation methods, in accordance with our approach, especially for FGSRA and MuMoDIG. For our proposed AFA, we set the maximum radius of uniform sampling centered at adversarial example $\xi = 3 \times \epsilon$, the number of sampling $N = 20$, the radius of MonteCarlo Adversarial Sampling $\gamma_{MCAS} = 0.15 \times \epsilon$, the momentum decay of MonteCarlo Adversarial Sampling $\eta_{MCAS} = 0.9$, the flatness balanced coefficient $\beta_{f}=0.5$ and the flatness item coefficient $\lambda_{f} = \alpha\times\beta_{f}$.

\noindent {\bf{Evaluation metrics.}} Given that our proposed AFA targets transferable attacks, we use the fool rate to measure the success of the attacks. More specifically, the fooling rate indicates the proportion of adversarial examples that successfully deceive the target model compared to all the adversarial examples created.

\begin{figure*}[htb]
\centering
\includegraphics[width=\linewidth,height=5.1cm]{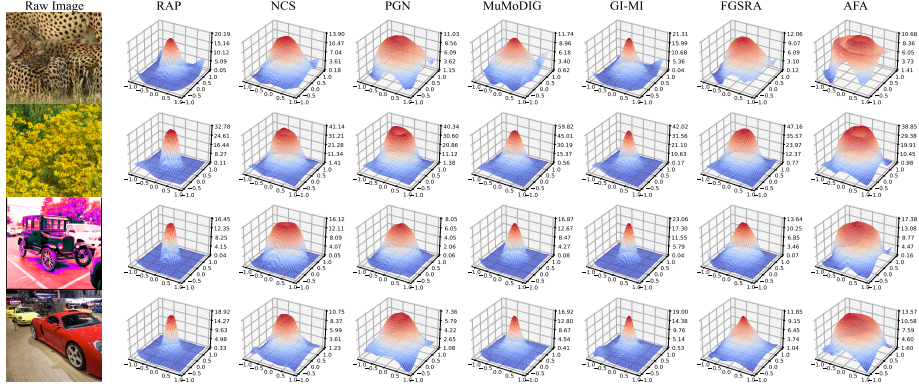}
\caption{Visualization of loss surfaces along two random directions for two randomly sampled
adversarial examples on the surrogate Inc-v3. The center of each 2D graph corresponds to
the adversarial example generated by different attack methods. The $x$ and $y$ axes represent the random noises added to $x_{adv}$ twice in succession. The $z$ axis indicates the loss value.}
\label{loss_surface}
\end{figure*}

\begin{table}[!t]
\centering
\caption{The untargeted attack success rates of various gradient-based
attacks in the normal model setting. * indicates the results on the white-box model. \textbf{BOLD} indicates the best.}
\resizebox{8.4cm}{4.4cm}{
\begin{tabular}{c|c|ccccccc}
		\hline
		\multicolumn{1}{c|}{Model}  
		 &\multicolumn{1}{c|}{Attack} & \multicolumn{1}{c}{Inc-v3} & \multicolumn{1}{c}{Inc-v4} &\multicolumn{1}{c}{Res-50} & \multicolumn{1}{c}{Res-101} & \multicolumn{1}{c}{InRes-v2} &\multicolumn{1}{c}{VGG-19} & \multicolumn{1}{c}{Dense-121}  \\ \hline
 \multirow{7}{*}{Inc-v3} &\multicolumn{1}{c|}{RAP} & 99.90* & 80.60 & 84.50 & 76.70 & 80.10 & 82.30 & 69.50 \\&\multicolumn{1}{c|}{NCS} & \textbf{100.0*} & 82.50 & 77.80 & 73.80 & 79.80 & 81.00 & 77.60 \\&\multicolumn{1}{c|}{PGN} & \textbf{{100.0*}} & {90.50} & {85.60} & {81.60} & {89.60} & {87.20} & {87.30} \\&\multicolumn{1}{c|}{MuMoDIG} & \textbf{100.0*} & 78.30 & 74.00 & 71.80 & 76.10 & 76.10 & 74.50 \\&\multicolumn{1}{c|}{GI-MI} & \textbf{100.0*} & 66.40 & 67.50 &  59.70 &  64.30 & 67.60 & 56.30 \\ &\multicolumn{1}{c|}{FGSRA} & \textbf{100.0*} & 87.00 & 80.70 & 77.20 &  85.90 & 84.00 & 82.80 \\&\multicolumn{1}{c|}{\bf{AFA}} & \textbf{{100.0*}} & \textbf{{95.20}} & \textbf{{89.90}} & \textbf{{87.70}} & \textbf{{95.40}} & \textbf{{91.50}} & \textbf{{91.90}} \\ \hline

 \multirow{7}{*}{Inc-v4} &\multicolumn{1}{c|}{RAP} & 78.00 & 99.80* & 80.90 & 74.80 & 70.60 & 80.80 & 63.50 \\&\multicolumn{1}{c|}{NCS} & 86.20 & 99.50* & 78.40 & 75.40 & 81.80 & 84.50 & 78.60 \\&\multicolumn{1}{c|}{PGN} & 91.50 & 99.40* & 85.80 & 82.10 & 87.80 & 90.40 & 87.00 \\&\multicolumn{1}{c|}{MuMoDIG} & 85.30 & 99.90* & 76.90 & 74.30 & 81.00 & 81.70 & 77.00 \\&\multicolumn{1}{c|}{GI-MI} & 71.40 & \textbf{100.0*} & 67.00 & 60.80 & 60.00 & 72.50 & 56.90  \\ &\multicolumn{1}{c|}{FGSRA} & 88.40 &  99.30* & 82.80 & 79.50 & 85.20 & 86.00 & 83.50 \\&\multicolumn{1}{c|}{\bf{AFA}} & \textbf{95.90} & \textbf{100.0*} & \textbf{91.80} & \textbf{89.80} & \textbf{93.70} & \textbf{95.70} & \textbf{92.30} \\
 \hline

 \multirow{7}{*}{Res-50} &\multicolumn{1}{c|}{RAP} & 73.70 & 87.00 & 99.90* & 97.90 & 62.90 & 87.90 & 76.70 \\&\multicolumn{1}{c|}{NCS} & 84.20 & 82.70 & \textbf{100.0*} & 99.50 & 76.90 & 94.60 & 91.90 \\&\multicolumn{1}{c|}{PGN} & 85.20 & 83.20 & \textbf{100.0*} & 99.20 & 78.10 & 95.00 & 92.90 \\&\multicolumn{1}{c|}{MuMoDIG} & 79.20 & 74.60 & \textbf{100.0*} & 99.00 & 65.40 & 90.20 & 82.70 \\&\multicolumn{1}{c|}{GI-MI} & 60.60 & 55.10 & \textbf{100.0*} & 99.00 & 45.10 & 81.30 & 65.80 \\ &\multicolumn{1}{c|}{FGSRA} & 87.00 & 83.60 & \textbf{100.0*} & 99.40 & 78.20 & 95.00 & 93.10 \\&\multicolumn{1}{c|}{\bf{AFA}} & \textbf{89.50} & \textbf{88.20} & \textbf{100.0*} & \textbf{99.90} & \textbf{82.60} & \textbf{97.90} & \textbf{95.10} \\
 \hline

 \multirow{7}{*}{Res-101} &\multicolumn{1}{c|}{RAP} & 77.00 & 69.50 & 98.80 & \textbf{100.0*} & 64.10 & 87.70 & 79.10 \\&\multicolumn{1}{c|}{NCS} & 85.10 & 82.60 & 99.40 & \textbf{100.0*} & 77.20 & 93.80 & 91.40 \\&\multicolumn{1}{c|}{PGN} & 87.30 & 83.00 & 99.50 & \textbf{100.0*} & 80.00 & 94.90 & 93.70 \\&\multicolumn{1}{c|}{MuMoDIG} & 80.20 & 74.90 & 99.60 & \textbf{100.0*} & 67.90 & 89.20 & 85.00 \\&\multicolumn{1}{c|}{GI-MI} & 63.80 & 56.20 & 99.60 &  \textbf{100.0*} & 47.30 & 78.00 & 68.20 \\ &\multicolumn{1}{c|}{FGSRA} & 86.80 & 83.70 & 99.40 & \textbf{100.0*} & 79.60 & 94.90 & 93.50 \\&\multicolumn{1}{c|}{\bf{AFA}} & \textbf{89.40} & \textbf{87.60} & \textbf{99.80} & \textbf{100.0*} & \textbf{83.80} & \textbf{96.20} & \textbf{95.80} \\
 \hline
 
  \end{tabular}
 \label{single_normal}
 }
\end{table}

\begin{table}[!t]
\centering
\caption{The untargeted attack success rates of various gradient-based
attacks in the diverse model architectures setting. \textbf{BOLD} indicates the best.}
\resizebox{7.8cm}{4.4cm}{

\begin{tabular}{c|c|cccccc}
		\hline
		\multicolumn{1}{c|}{Model}  
		 &\multicolumn{1}{c|}{Attack} & \multicolumn{1}{c}{MobilNet} & \multicolumn{1}{c}{PASNet-L} &\multicolumn{1}{c}{ViT-B/16} & \multicolumn{1}{c}{PiT-S} & \multicolumn{1}{c}{MLP-mixer} &\multicolumn{1}{c}{ResMLP} \\ \hline
 \multirow{7}{*}{Inc-v3} &\multicolumn{1}{c|}{RAP} & 89.10 & 71.70 & 23.70 & 26.60 & 50.30 & 43.30 \\&\multicolumn{1}{c|}{NCS} & 79.20 & 73.30 & 39.40 & 39.10 & 56.30 & 54.10 \\&\multicolumn{1}{c|}{PGN} & 85.90 & 84.00 & 50.70 & 51.20 & 63.30 & 66.30 \\&\multicolumn{1}{c|}{MuMoDIG} & 79.00 & 70.60 & 30.70 & 30.90 & 53.80 & 47.40 \\&\multicolumn{1}{c|}{GI-MI} & 76.50 & 58.40 & 20.40 & 22.10 & 46.30 & 35.60 \\ &\multicolumn{1}{c|}{FGSRA} & 82.90 & 78.90 & 44.80 & 47.00 & 60.10 & 60.30 \\&\multicolumn{1}{c|}{\bf{AFA}}  & \textbf{91.10} & \textbf{87.40} & \textbf{56.70} & \textbf{59.20} & \textbf{65.30} & \textbf{72.30} \\ \hline

 \multirow{7}{*}{Inc-v4} &\multicolumn{1}{c|}{RAP} & 89.90 & 69.70 & 22.50 & 25.20 & 48.70 & 36.20 \\&\multicolumn{1}{c|}{NCS} & 80.10  & 79.30 & 43.10 & 50.00 & 57.60 & 55.70 \\&\multicolumn{1}{c|}{PGN} & 86.30  & 88.30 & 54.70 & 54.90 & 62.30 & 66.00 \\&\multicolumn{1}{c|}{MuMoDIG} & 82.00 & 77.50 & 31.80 & 38.30 & 51.50 & 47.90 \\&\multicolumn{1}{c|}{GI-MI} & 76.30 & 64.30 & 20.80 & 23.80 & 44.10 & 33.10 \\ &\multicolumn{1}{c|}{FGSRA} & 82.90 & 83.30 & 50.00 & 54.70 & 59.50 & 61.10 \\&\multicolumn{1}{c|}{\bf{AFA}} & \textbf{92.20} & \textbf{92.00} & \textbf{62.80} & \textbf{68.20} & \textbf{67.00} & \textbf{74.10} \\
 \hline

 \multirow{7}{*}{Res-50} &\multicolumn{1}{c|}{RAP} & 95.20 & 70.70 & 25.20 & 31.80 & 49.20 & 43.70 \\&\multicolumn{1}{c|}{NCS} & 97.00 & 84.30 & 40.90 & 45.20 & 54.90 & 59.20 \\&\multicolumn{1}{c|}{PGN} & 96.80 & 88.70 & 47.70 & 45.10 & 62.20 & 65.20 \\&\multicolumn{1}{c|}{MuMoDIG} & 96.20 & 73.40 & 21.90 & 29.10 & 44.80 & 39.60 \\&\multicolumn{1}{c|}{GI-MI} & 92.50 & 59.90 & 17.20 & 19.20 & 41.70 & 31.10 \\ &\multicolumn{1}{c|}{FGSRA} & 97.50 & 88.00 & 45.00 & 45.60 & 59.70 & 65.20  \\&\multicolumn{1}{c|}{\bf{AFA}} & \textbf{98.20} & \textbf{91.10} & \textbf{48.70} & \textbf{49.90} & \textbf{62.30} & \textbf{68.70} \\
 \hline

 \multirow{7}{*}{Res-101} &\multicolumn{1}{c|}{RAP} & 96.00 & 74.00 & 28.30 & 31.70 & 52.60 & 47.30 \\&\multicolumn{1}{c|}{NCS} & 96.20 & 83.60 & 44.10 & 46.00 & 57.00 & 61.40 \\&\multicolumn{1}{c|}{PGN} & 97.00 & 88.30 & 50.80 & 48.30 & 62.20 & 68.40 \\&\multicolumn{1}{c|}{MuMoDIG} & 95.40 & 72.00 & 26.50 & 30.40 & 47.30 & 42.50 \\&\multicolumn{1}{c|}{GI-MI} & 90.30 & 59.70 & 19.60 & 20.40 & 43.40 & 31.30 \\ &\multicolumn{1}{c|}{FGSRA} & 97.40 & 86.80 & 49.50 & 47.70 & 60.00 & 66.90 \\&\multicolumn{1}{c|}{\bf{AFA}} & \textbf{98.10} & \textbf{89.50} & \textbf{55.10} & \textbf{52.40} & \textbf{64.30} & \textbf{70.30} \\
 \hline
 
  \end{tabular}
  
 \label{single_diverse_arch}
 }
\end{table}

\subsection{Visualization of Loss Surfaces for Adversarial Example} To validate the efficacy of our proposed AFA in identifying adversarial examples within a flat maxima region, we compare the loss surface maps of adversarial examples generated by different baselines on the surrogate Inc-v3 model. Each 2D graph corresponds to an adversarial example, with the adversarial example located in the middle. Four images are randomly selected from the dataset and their loss surfaces are compared in Fig. \ref{loss_surface}, with each row displaying the visualization of one image across all baselines. As shown in Fig. \ref{loss_surface}, it is clear that our AFA method helps adversarial examples reach flatter peaks compared to all baseline methods. It demonstrates that our approach is able to generate adversarial examples within the smooth maximum region. AFA demonstrates stronger transferability compared to these baselines due to the presence of adversarial examples in flatter local maxima, a fact further confirmed in subsequent experiments.

\begin{table}[!t]
\centering
\caption{The untargeted attack success rates of various gradient-based
attacks in the adversarially trained model setting. \textbf{BOLD} indicates the best.}
\resizebox{9cm}{2.2cm}{
\begin{tabular}{c|c|cccc|c|cccc}
		\hline
		\multicolumn{1}{c|}{Attack} & \multicolumn{1}{c|}{Model}  
		 & \multicolumn{1}{c}{Inc-v3*} & \multicolumn{1}{c}{Inc-v3$_{ens}$} &
   \multicolumn{1}{c}{Inc-v4$_{ens}$}  & \multicolumn{1}{c|}{InRes-v2$_{ens}$} & \multicolumn{1}{c|}{Model}  
		 & \multicolumn{1}{c}{Inc-v3*} & \multicolumn{1}{c}{Inc-v3$_{ens}$} &
   \multicolumn{1}{c}{Inc-v4$_{ens}$}  & \multicolumn{1}{c}{InRes-v2$_{ens}$}  \\ \hline
 
 \multicolumn{1}{c|}{RAP} & \multirow{7}{*}{Inc-v3} & 34.80 & 17.00 & 17.10 & 8.30 & \multirow{7}{*}{Inc-v4} & 31.90 & 18.20 & 17.00 & 7.80 \\ \multicolumn{1}{c|}{NCS} & & 55.40 & 53.70 & 53.60 & 35.10 & & 53.40 & 54.50 & 54.80 & 40.40  \\ \multicolumn{1}{c|}{PGN} & & 71.30 & 64.40 & 66.50 & 46.20 & & 64.80 & 67.30 & 64.60 & 48.20 \\ \multicolumn{1}{c|}{MuMoDIG} & & 44.50 & 40.70 & 39.90 & 20.60  & & 41.70 & 41.70 & 38.60 & 23.60 \\ \multicolumn{1}{c|}{GI-MI} & & 30.60 & 22.20  & 20.80  & 9.80 & & 28.40 & 20.30 & 18.90 & 10.50 \\ \multicolumn{1}{c|}{FGSRA} & & 67.50 & 61.50 & 63.00 & 43.30 & & 62.50 & 63.70 & 63.10 & 48.00 \\ \multicolumn{1}{c|}{\bf{AFA}} & & \textbf{75.70} & \textbf{73.00} & \textbf{73.20} & \textbf{50.90} & & \textbf{73.20} & \textbf{75.10} & \textbf{75.30} & \textbf{57.10} \\ \hline
 
  \multicolumn{1}{c|}{RAP} &\multirow{7}{*}{Res-50} & 32.80 & 19.50 & 17.10 & 11.10 &\multirow{7}{*}{Res-101} &35.50 & 20.90 & 19.30 & 9.30 \\ \multicolumn{1}{c|}{NCS} & &52.00 & 50.40 & 48.70 & 35.50  & & 55.90 & 54.40 & 55.40 & 42.10 \\ \multicolumn{1}{c|}{PGN} & & 59.90 & 57.50 & 57.10 & 43.50  & & 65.10 & 63.10 & 62.50 & 51.10 \\ \multicolumn{1}{c|}{MuMoDIG} & & 30.40 & 24.90 & 23.40 & 12.50 & & 33.70 & 26.50 & 26.20 & 15.40 \\  
  \multicolumn{1}{c|}{GI-MI} & & 23.60 & 17.80 & 17.10 & 9.10 & & 24.80 & 18.10 & 16.70 & 9.30 \\ 
  \multicolumn{1}{c|}{FGSRA} & & 56.40 & 56.70 & 55.90 & 42.30 & & 63.10 & 60.70 & 60.20 & 48.40 \\
  \multicolumn{1}{c|}{\bf{AFA}} & & \textbf{61.10} & \textbf{59.30} & \textbf{59.90} & \textbf{44.70} & & \textbf{67.20} & \textbf{63.70} & \textbf{63.90} & \textbf{51.80}  \\ \hline

  \end{tabular}
 \label{single_adv}
 }
\end{table}

\begin{table}[!t]
\centering
\caption{The untargeted attack success rates of various gradient-based
attacks in the defense model setting. \textbf{BOLD} indicates the best.}
\resizebox{8.8cm}{4.4cm}{
\begin{tabular}{c|c|ccccccccc}
		\hline
		\multicolumn{1}{c|}{Model}  
		 &\multicolumn{1}{c|}{Attack} & \multicolumn{1}{c}{RS} & \multicolumn{1}{c}{HGD} &\multicolumn{1}{c}{FD} & \multicolumn{1}{c}{NRP} & \multicolumn{1}{c}{ComDefend} &\multicolumn{1}{c}{JPEG} & \multicolumn{1}{c}{Bit-Red}& \multicolumn{1}{c}{R\&P}& \multicolumn{1}{c}{NIPS-r3}  \\ \hline

 \multirow{7}{*}{Inc-v3} &
 \multicolumn{1}{c|}{RAP} & 25.20 & 2.20 & 63.40 & 57.10 & 55.70 & 61.20 & 71.40 & 77.40 & 12.90 \\&\multicolumn{1}{c|}{NCS} & 31.30 & 32.30 & 72.70 & 66.90 & 66.80 & 71.20 & 75.00 & 75.70 & 41.50 \\&\multicolumn{1}{c|}{PGN} & 42.20 & 39.00 & 82.50 & 72.00 & 77.90 & 81.40 & 83.50 & 86.00 & 54.10 \\ &\multicolumn{1}{c|}{MuMoDIG} & 23.50 & 18.80 & 69.60 & 61.80 & 64.30 & 64.50 & 70.00 & 73.40 & 27.40 \\ &\multicolumn{1}{c|}{GI-MI} & 20.50 & 4.30 & 54.50 & 57.00 & 49.80 & 48.80 & 56.00 & 57.60 & 14.70\\ &\multicolumn{1}{c|}{FGSRA} & 39.50 & 39.40 & 78.80 & 68.50 & 74.10 & 77.20 & 79.60 & 81.00 & 50.90 \\&\multicolumn{1}{c|}{\bf{AFA}} & \textbf{44.00} & \textbf{43.10} & \textbf{86.70} & \textbf{74.80} & \textbf{82.40} & \textbf{87.60} & \textbf{89.80} & \textbf{91.30} & \textbf{59.80}\\ \hline

 \multirow{7}{*}{Inc-v4}  &\multicolumn{1}{c|}{RAP} & 19.00 & 5.10 & 48.50 & 58.00 & 40.00 & 43.30 & 53.90 & 77.90 & 12.80 \\&\multicolumn{1}{c|}{NCS} & 32.70 & 37.70 & 72.10 & 67.10 & 67.50 & 70.10 & 74.60 & 79.00 & 45.40 \\&\multicolumn{1}{c|}{PGN} & 41.50 & 39.00 & 81.50 & 75.20 & 75.40 & 80.70 & 82.70 & 85.40 & 56.20 \\&\multicolumn{1}{c|}{MuMoDIG} & 25.20 & 21.80 & 66.90 & 62.70 & 61.70 & 64.80 & 70.10 & 77.00 & 29.90 \\ &\multicolumn{1}{c|}{GI-MI} & 21.80 & 3.30 & 51.40 & 57.80 & 45.70 & 47.70 & 54.60 & 57.80 & 13.20 \\ &\multicolumn{1}{c|}{FGSRA} & 38.50 & 42.00 & 77.40 & 70.70 & 72.40 & 75.70 & 79.10 & 82.60 & 53.40 \\&\multicolumn{1}{c|}{\bf{AFA}} & \textbf{45.30} & \textbf{48.20} & \textbf{88.40} & \textbf{77.80} & \textbf{81.70} & \textbf{87.10} & \textbf{89.40} & \textbf{91.90}& \textbf{63.70}\\
 \hline

 \multirow{7}{*}{Res-50}  &\multicolumn{1}{c|}{RAP} & 29.30 & 5.20 & 77.10 & 81.90 & 64.60 & 89.20 & 94.40 &  66.70 & 13.90 \\&\multicolumn{1}{c|}{NCS} & 39.60 & 63.10 & 91.20 & 89.40 & 81.80 & 98.50 & 99.60 & 76.90 & 44.60 \\&\multicolumn{1}{c|}{PGN} & 47.70 & 63.30 & 92.30 & 93.70 & 85.20 & 98.00 & 98.70 & 79.20 & 54.30 \\&\multicolumn{1}{c|}{MuMoDIG} & 23.40 & 36.30 & 79.50 & 79.50 & 64.70 & 91.30 & 97.30 & 67.90 & 18.40 \\&\multicolumn{1}{c|}{GI-MI} & 21.60 & 11.70 & 63.80 & 76.10 & 47.80 & 81.70 & 92.40 & 47.40 & 11.70\\ &\multicolumn{1}{c|}{FGSRA} & 45.50 & 62.80 & 92.60 & 91.90 & 85.30 & 99.00 & 99.70 & 77.80 & 52.20 \\ &\multicolumn{1}{c|}{\bf{AFA}} & \textbf{49.30} & \textbf{67.80} & \textbf{94.40} & \textbf{94.90} & \textbf{87.60} & \textbf{99.40} & \textbf{99.90} & \textbf{83.40} & \textbf{56.30} \\
 \hline

 \multirow{7}{*}{Res-101} &\multicolumn{1}{c|}{RAP} & 30.70 & 14.20 & 95.60 & 79.50 & 82.90 & 75.40 & 82.40 & 67.40 & 15.60 \\&\multicolumn{1}{c|}{NCS} & 43.30 & 73.50 & 98.60 & 88.40 & 97.00 & 90.30 & 92.80 & 75.70 & 49.70 \\&\multicolumn{1}{c|}{PGN} & 54.80 & 71.30 & 98.50 & 91.50 & 97.60 & 92.10 & 94.10 &77.90 & 58.20 \\&\multicolumn{1}{c|}{MuMoDIG} & 24.60 & 44.50 & 98.30 & 79.20 & 90.40 & 75.40 & 82.40 & 68.90 & 20.20 \\&\multicolumn{1}{c|}{GI-MI} & 22.20 & 13.10 & 94.20 & 74.00 & 73.90 & 58.70 & 69.00 & 48.90 & 12.70 \\ &\multicolumn{1}{c|}{FGSRA} & 51.10 & 72.20 & 99.40 & 89.90 & 97.70 & 92.20 & 94.00 & 78.70 & 55.90 \\ &\multicolumn{1}{c|}{\bf{AFA}} & \textbf{55.20} & \textbf{78.70} & \textbf{99.90} & \textbf{92.80} & \textbf{98.50} & \textbf{94.60} & \textbf{96.00} & \textbf{92.30}& \textbf{61.60}\\
 \hline
 
  \end{tabular}
 \label{single_defense}
 }
\end{table}

\subsection{Attack a Single Model}
\noindent {\bf{Under normally trained models.}}
Initially, we compared the ASR of our fully automated aircraft with baselines from seven normally trained models. Specifically, we choose one of four models (such as Inc-v3, Inc-v4, Res-50, and Res-101) to serve as the white-box surrogate model, while the remaining models trained normally (six CNNs and six advanced models with various architectures) are considered the black-box target models. After creating adversarial examples on the substitute model, they can be used to test how well adversarial attacks transfer to the black-box target models. As illustrated in Table \ref{single_normal}, our approach consistently achieves better ASR performance compared to other methods. Our AFA clearly shows much better transferable attack performance than all other comparison models. Besides, as shown in Table \ref{single_diverse_arch}, some different model architectures (especially ViT-B/16, PiT-S and MLP-mixer) might significantly degrade the transferable ASRs compared to CNN-like architectures. It can be attributed to the shift of attention areas in these models, which has been demonstrated in \cite{TIM} and can usually be alleviated by the input transformation strategies such as \cite{DIM, SIM, TIM}. These input transformation strategies typically increase the diversity of the backward gradients and reduce the likelihood that the generated perturbations will overfit to high-frequency information and the current surrogate model. Whereas, our suggested approach consistently outperforms the last four different architectures. In general, our AFA method demonstrates a greater level of versatility in its ability to handle adversarial attacks on different architectures. For the page limitation, the experiments of generating adversarial examples on transformers (ViT-B/16, Swin-T) are also provided at \ref{single_transformer}.

\noindent {\bf{Under models equipped with defense strategies.}} By using improved robustness obtained from adversarial training, we confirm the effectiveness of our method on four different models equipped with various defense strategies, including adversarial techniques (namely, Inc-v3$^*$, Inc-v3$_{ens}$, Inc-v4$_{ens}$, and IncRes-v2$_{ens}$) and other defense models like random smoothing, active defenses and so on. Moreover, the white-box surrogate models share similarities with normally trained models, whereas these models with defense methods are considered black-box target models. As shown in Table \ref{single_adv}, the ASRs of all methods on target models consistently decrease. Nevertheless, our method outperforms others in terms of effectiveness. Our method effectively preserves the high level of attack transferability in the scenario of adversarially trained models. Besides, as illustrated in Table \ref{single_defense}, various defense models have different levels of resilience when facing adversarial attacks. The substantial decreases in ASR for RS and NIPS-r3 across all methods are especially notable. However, our suggested method consistently shows impressive deceptive abilities even in the face of strong defenses. For the page limitation, the validation on transformers-based surrogates (ViT-B/16, Swin-T) is also shown at \ref{defend_transformer}.

\begin{figure}[htbp]
\centering
\includegraphics[width=0.45\linewidth]{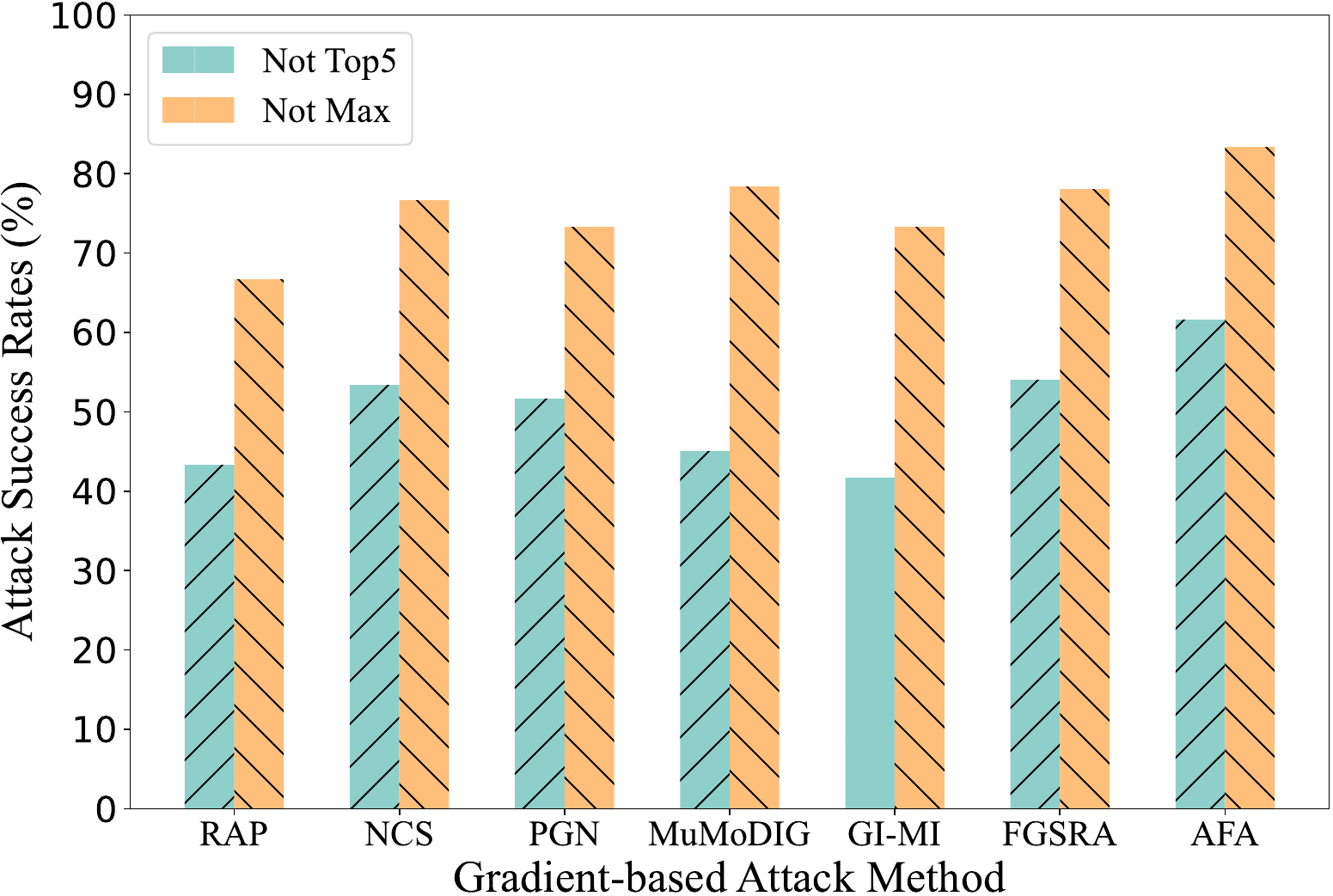}
\caption{Comparison of the average untargeted attack success rates between various gradient-based
attacks on the Baidu Cloud API. The white-box source model is Inc-v3.}
\label{bdc_comparison}
\end{figure}

\begin{figure}[htbp]
\centering
\includegraphics[width=1\linewidth]{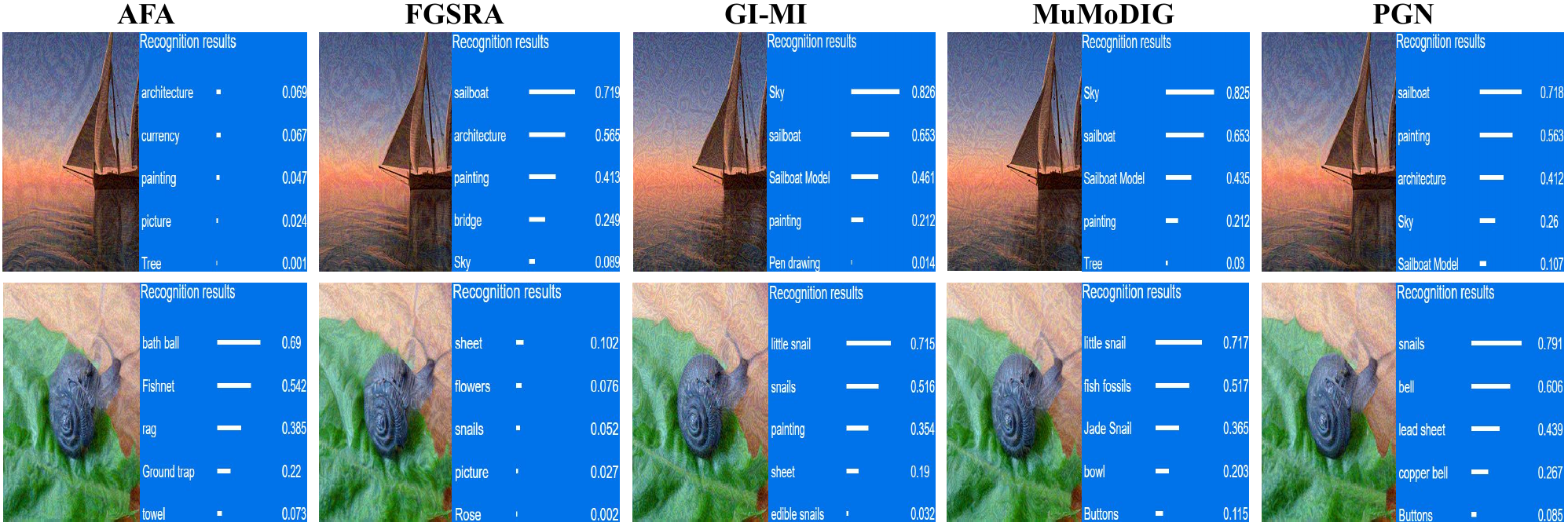}
\caption{Samples of the recognition results of adversarial examples generated by ours, FGSRA, GI-MI, MuMoDIG and PGN on the Baidu Cloud API. The white-box source model is Inc-v3.}
\label{bdc_sample_comparison}
\end{figure}

\subsection{Transfer to the commercial API}
To further demonstrate the broad applicability of our method, we select the Baidu Cloud system as a target. Besides, it is a commercial API capable of recognizing over 100,000 categories and thus exhibits strong representativeness. In practice, we randomly choose 60 images from the datasets and set Inc-v3 as the surrogate. Additionally, we focus not only on the max prediction score, but also on the top five predictions. In other words, we set the successful attack conditions as the commonly used wrong prediction at the maximum prediction score (simplified as ``Not Max''), and the more restrictive wrong classification within the top five predictions (simplified as ``Not Top5''). As shown in Fig. \ref{bdc_comparison}, our method achieves at least 5\% higher ASRs compared to other baselines under both conditions. Besides, we also display two groups of recognition results from the Baidu Cloud system. It is suggested that our method can more successfully deceive the Baidu Cloud system in Fig. \ref{bdc_sample_comparison}, while these images with added perturbations can still be classified correctly by human eyes. These results have verified our superiority preliminarily.

\begin{figure*}[!t]
    \centering
    \includegraphics[width=0.295\linewidth]{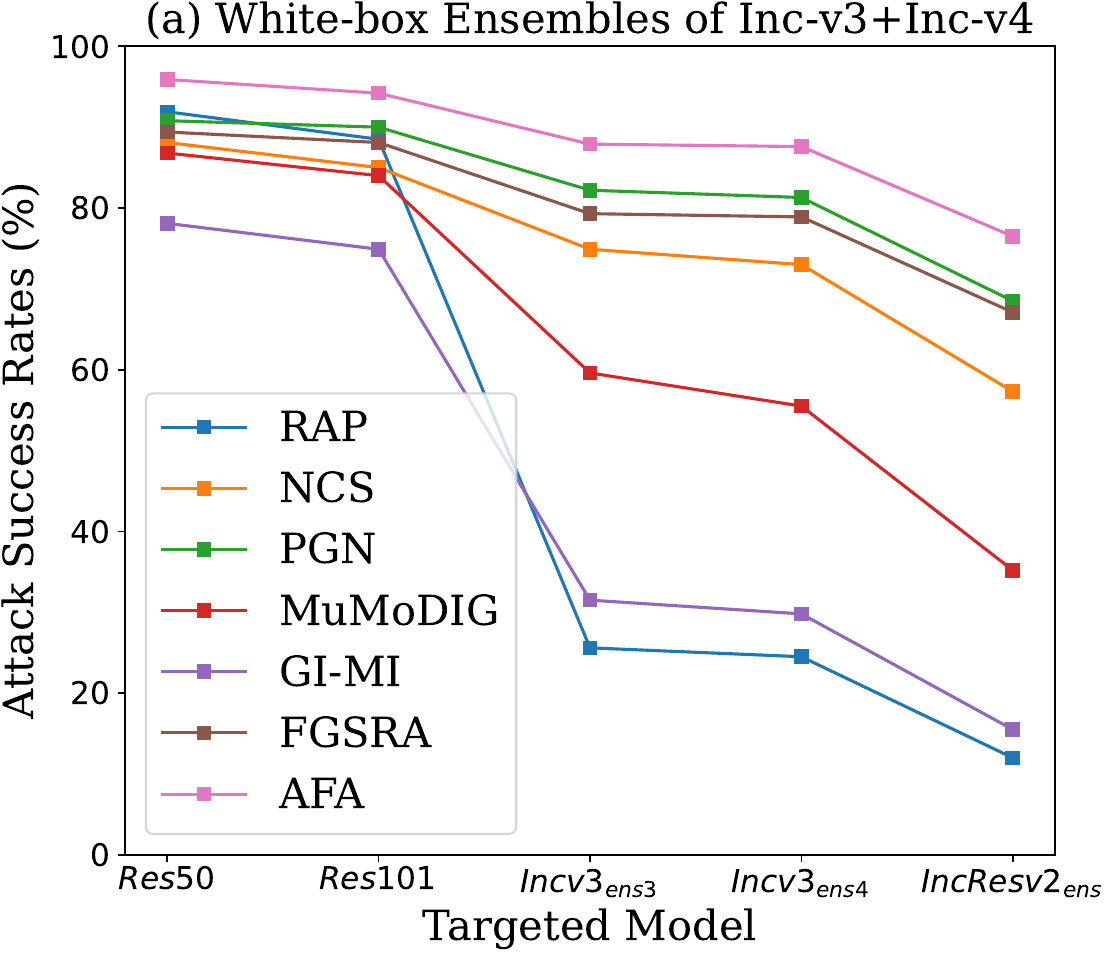}
    \hspace{-2mm}
    \includegraphics[width=0.295\linewidth]{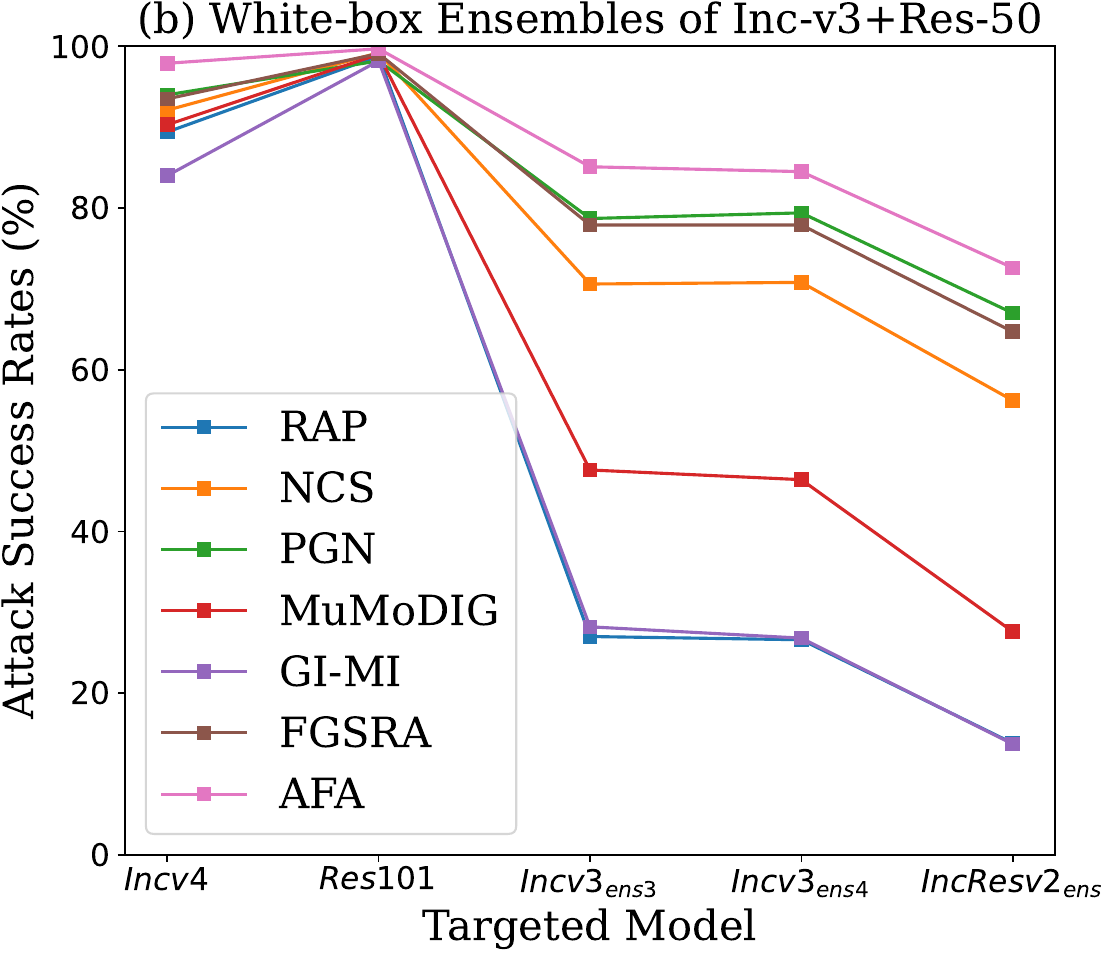}
    \hspace{-2mm}
    \includegraphics[width=0.295\linewidth]{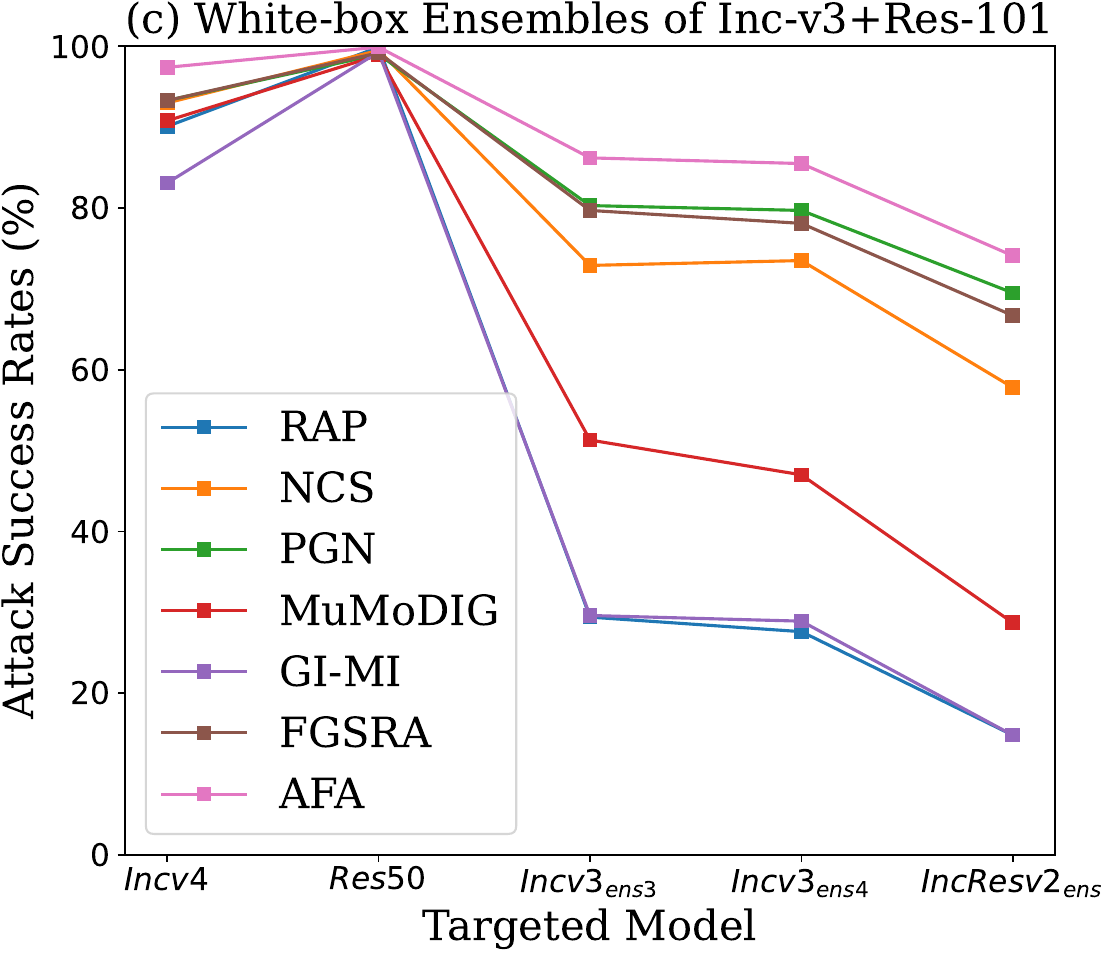}
    \vskip 1pt
    
    \centering
    \includegraphics[width=0.295\linewidth]{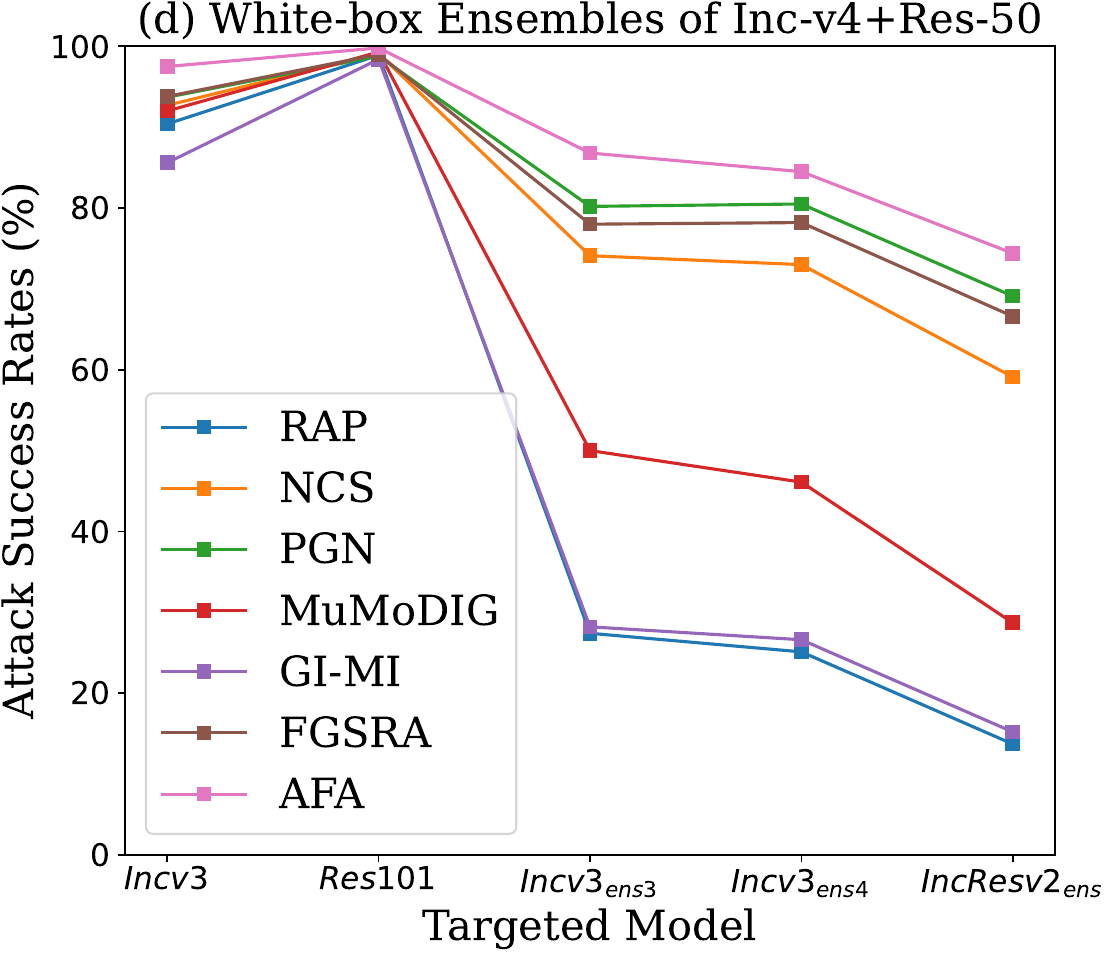}
   \hspace{-2mm}
    \centering
   \includegraphics[width=0.295\linewidth]{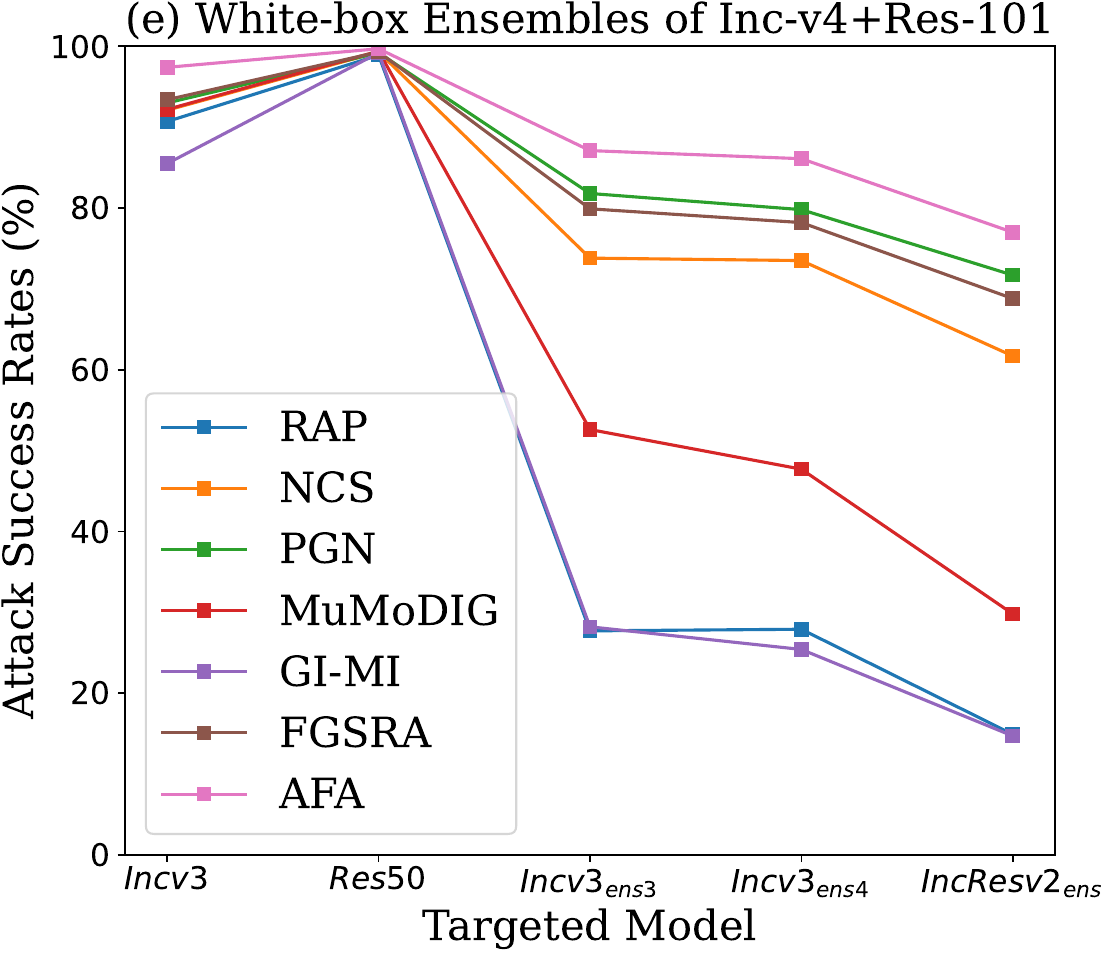}
  \hspace{-2mm}
  \includegraphics[width=0.295\linewidth]{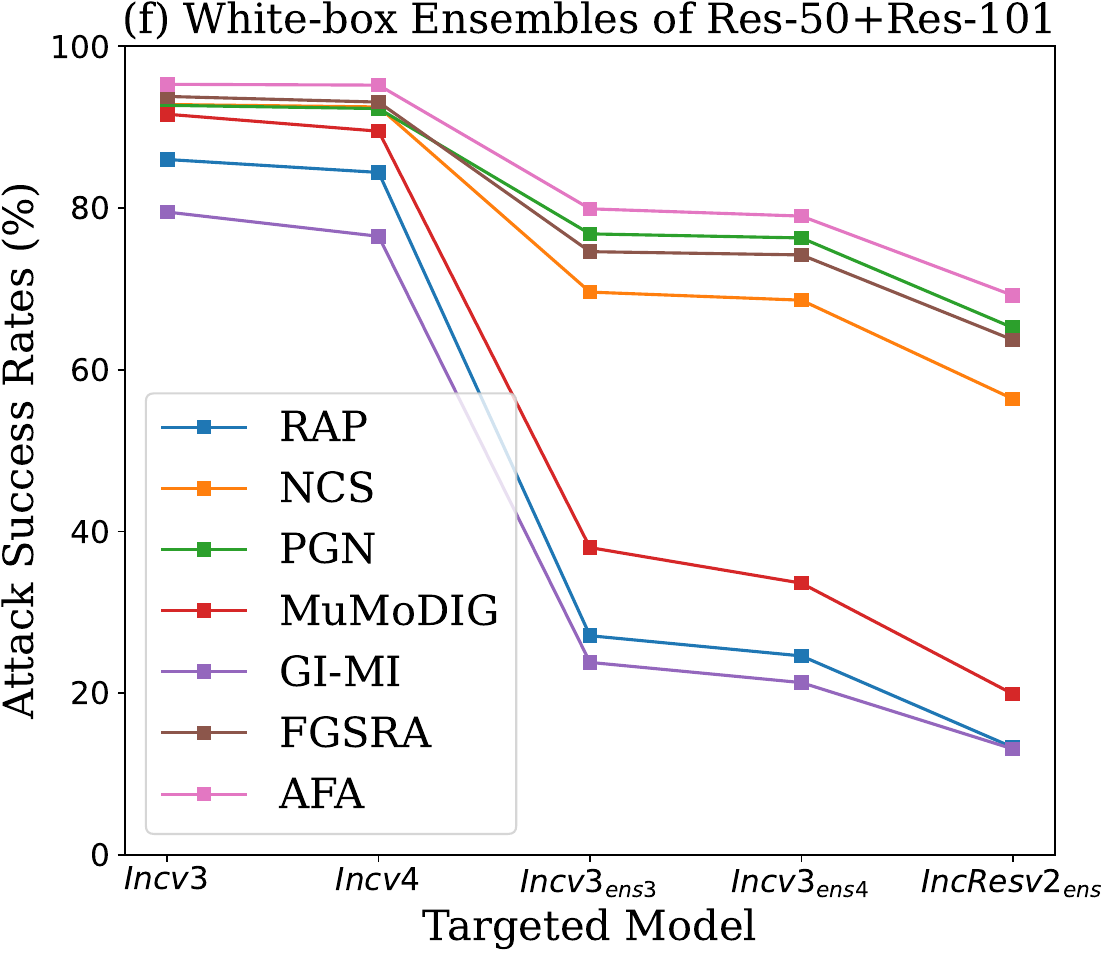}
  
    \caption{Untargeted attack success rates between various gradient-based attacks in the ensemble models setting. The white-box source ensemble models are in the title. The black-box target models are on $x$ axis.}
    \label{ensemble}
\end{figure*}

\subsection{Attack an Ensemble of Models}
Furthermore, we assess the effectiveness of our AFA in an ensemble-model setting, in addition to testing on an individual model. In this section, we use two models as the white-box surrogate \cite{MI}, creating an ensemble by averaging the logit outputs of both models. The adversaries are generated by randomly combining two models from four normally trained models: Inc-v3, Inc-v4, Res-50 and Res-101. All the ensemble models are assigned equal weights and we test the performance of transferability on both two normally trained models and three adversarially trained models. The results presented in Fig. \ref{ensemble} indicate that our AFA method consistently achieves the highest attack success rates among the six ensemble settings. Our AFA achieves significantly more stable ASRs compared to previous gradient-based attacks. Furthermore, our method demonstrates remarkable improvements even against adversarially trained models. These results validate the effectiveness of integrating both adversarial zeroth-order flatness and adversarial first-order flatness in enhancing the transferability of adversarial attacks, as opposed to solely optimizing either one or using only adversarial first-order flatness.

\begin{figure}[htbp]
    \centering
    \includegraphics[width=0.48\linewidth]{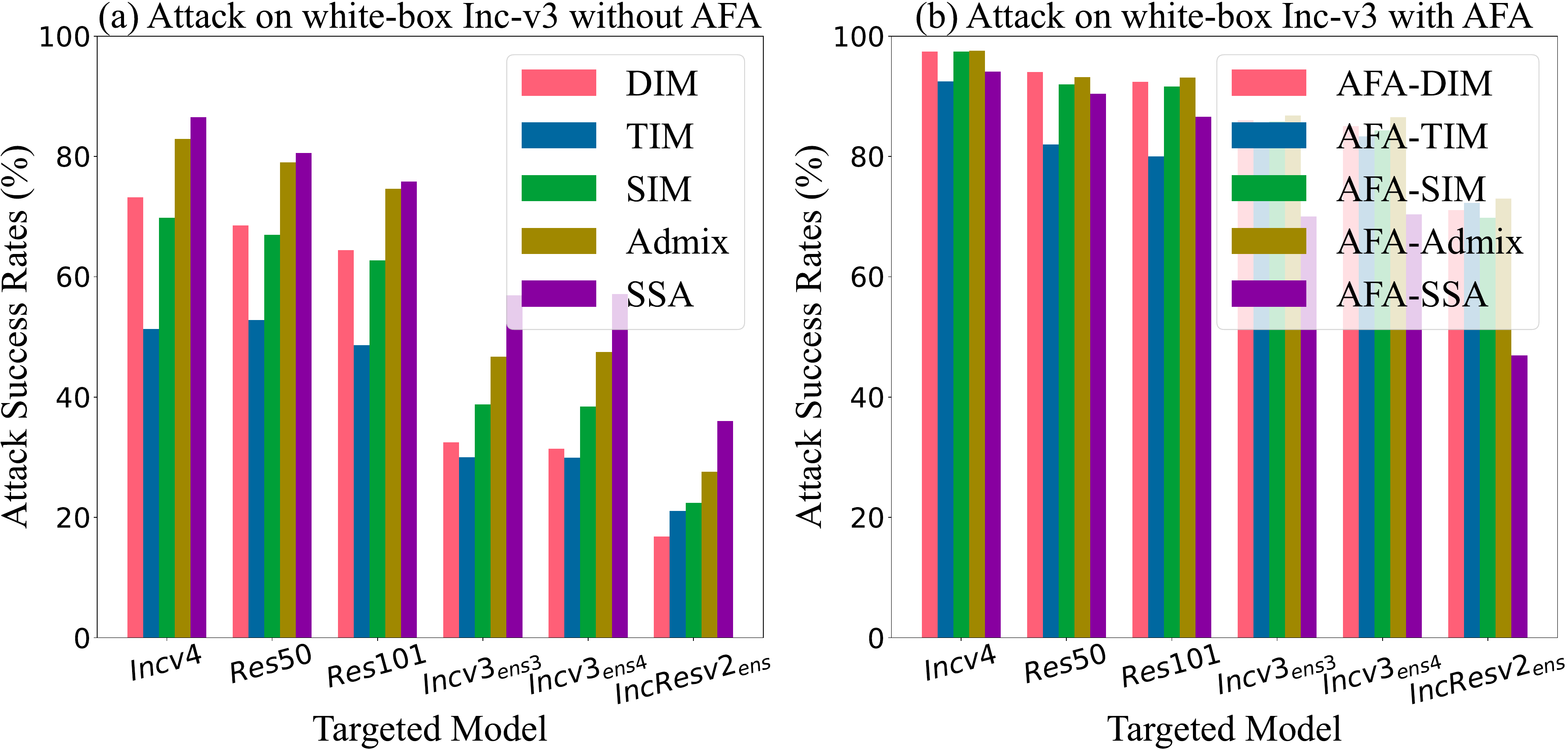}
    \hspace{-2mm}
    \includegraphics[width=0.48\linewidth]{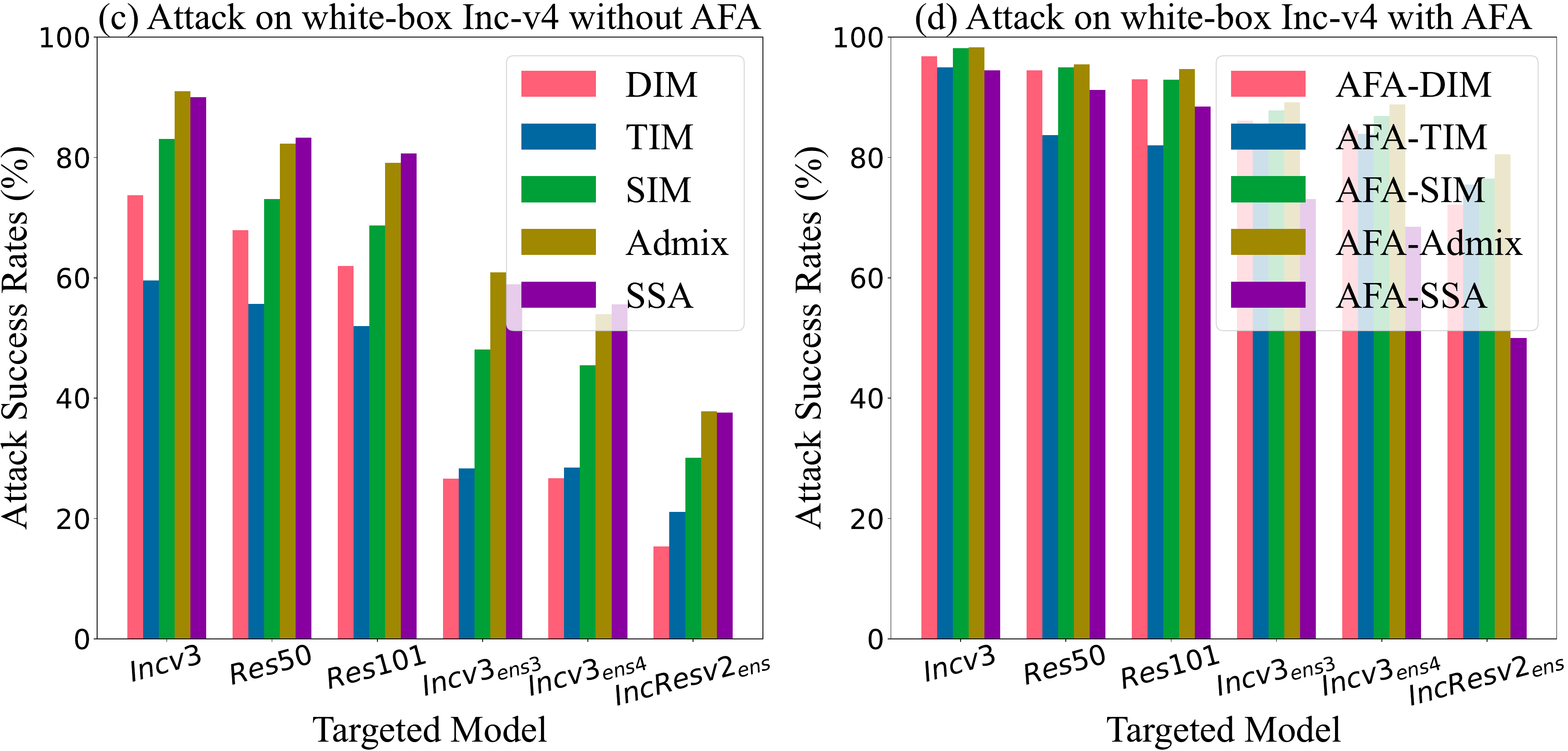}
 \vskip 1pt
 
	\centering
   \includegraphics[width=0.48\linewidth]{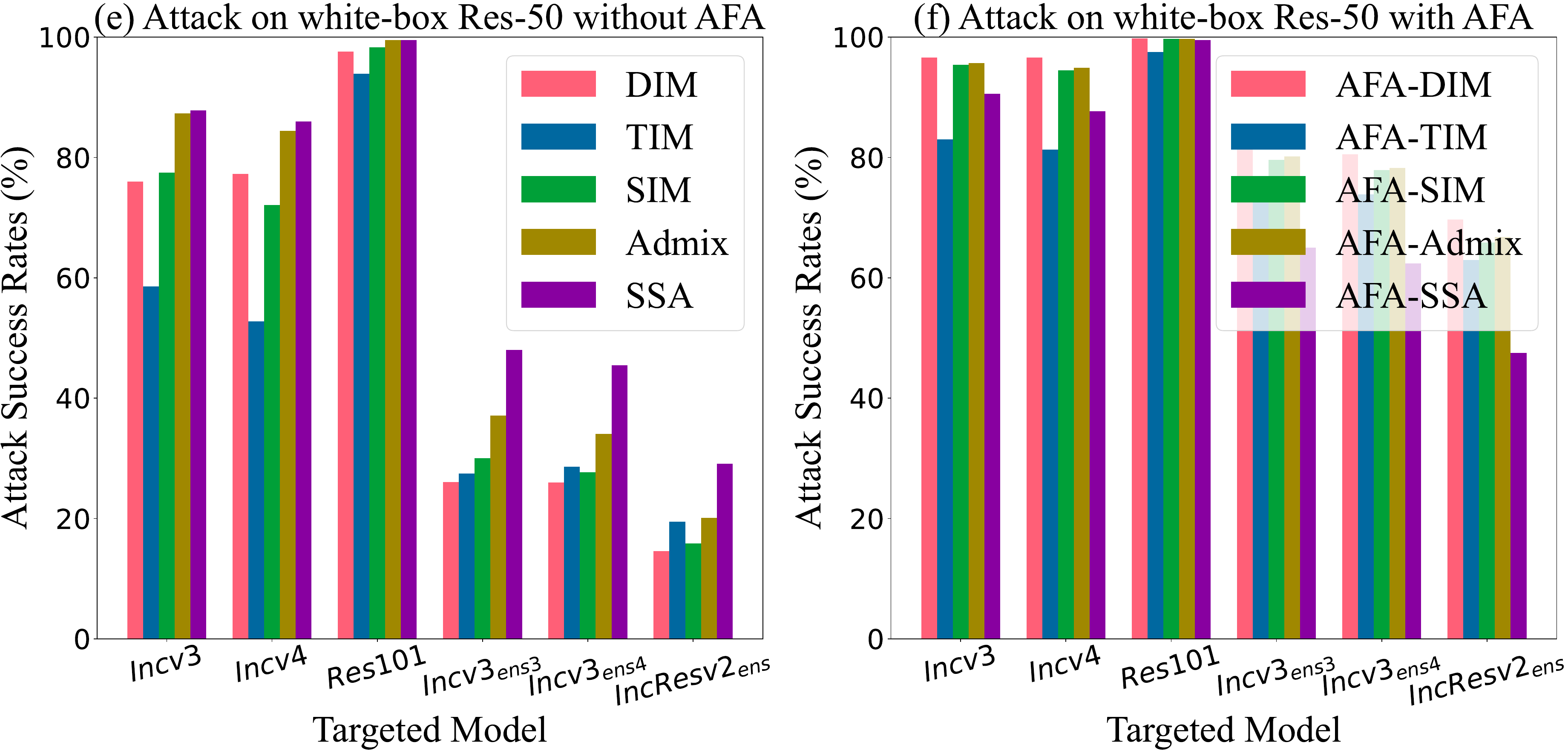}
    \hspace{-2mm}
  \includegraphics[width=0.48\linewidth]{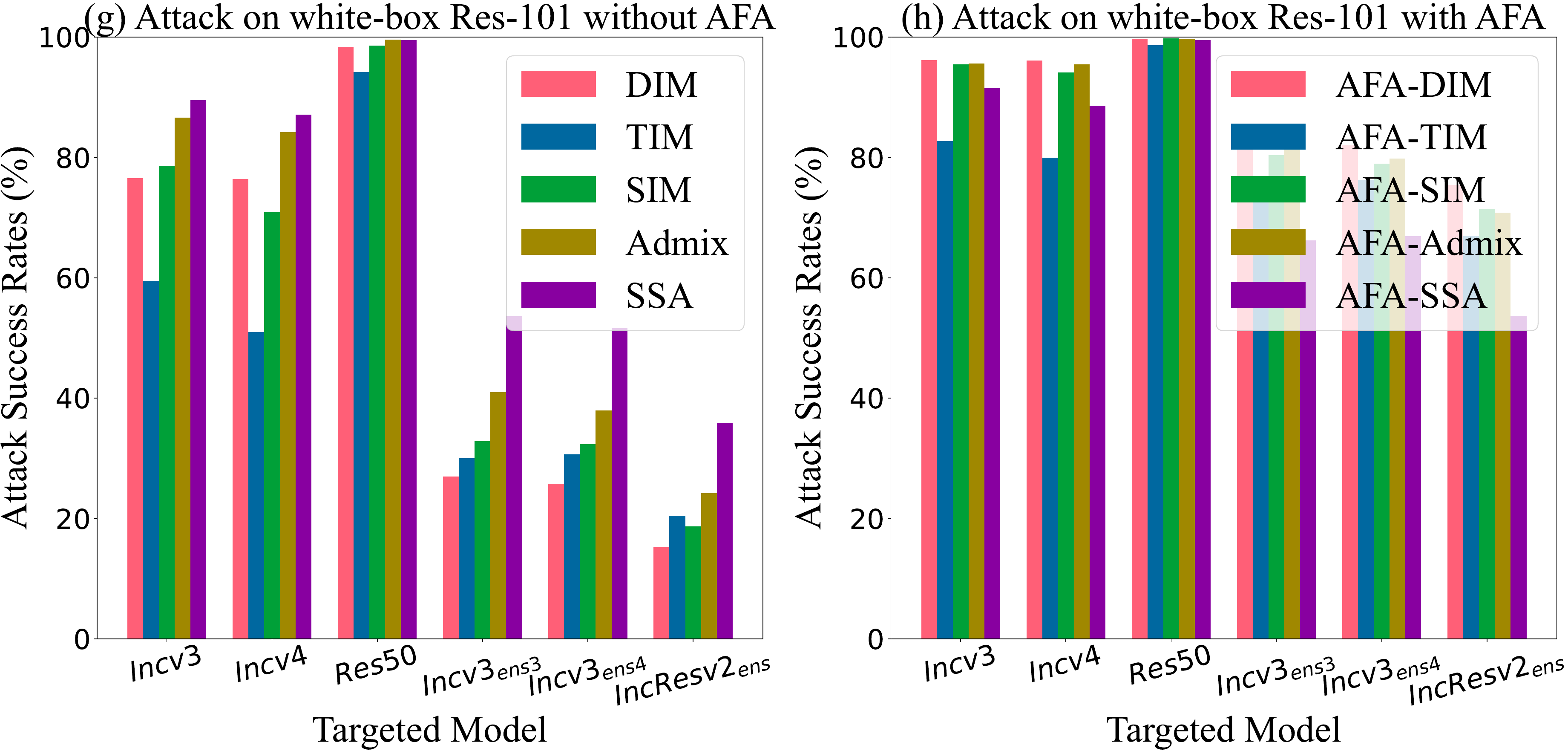}
  
    \caption{Untargeted attack success rates between various gradient-based attacks under the augmentation setting without/with AFA. The white-box source model is in the title. The black-box target models are on $x$ axis.}
    \label{FAA+input}
\end{figure}

\subsection{Combined with Input Transformation Attacks}
Currently, input transformation attacks have shown substantial compatibility with each other. Given that our AFA method generates adversarial examples utilizing gradient-level information, it naturally complements these input transformation-based methods to boost the adversarial transferability. To further validate the effectiveness of our AFA, we integrate it into various input transformations such as DIM, TIM, SIM, Admix and SSA. Subsequently, we generate adversarial examples individually on Inc-v3, Inc-v4, Res-50 and Res-101 models. Finally, we evaluate the transferability of these adversarial examples on six black-box models, including three normally trained models and three adversarially trained models. It should be mentioned that we compare the ASRs of input transformation-based attacks with and without our AFA, as well as calculate the ASRs of input transformation-based attacks utilizing all baselines. The experimental results can be observed in Fig. \ref{FAA+input}. When integrated with our gradient-based approach, it can greatly improve the transferability of adversarial attacks relying on input transformations towards black-box target models. Moreover, our approach exhibits significant enhancements in attack performance when integrated with these input transformation-based techniques. For instance, input transformation-based methods alone yield an average ASRs below 35\% on IncRes-v2$_{ens}$ in white-box Res-50. Nonetheless, when combined with our AFA method, the average ASRs surpass 60\%, representing 20\% increase as depicted in Fig. \ref{FAA+input}. More results can be referred in \ref{app5}.

\begin{table}[!t]
\centering
\caption{Ablation study conducted under normal models, adversarially trained models and diverse architectures. The white-box source models are IncRes-v2 and Inc-v3 respectively. The \checkmark and \XSolidBrush symbols indicate our method with and without the corresponding component respectively.}
\resizebox{8.4cm}{1.2cm}
{
\begin{tabular}{c|c|ccccccc}
		\hline
		\multicolumn{1}{c|}{Model}  
		 &\multicolumn{1}{c}{AF} & \multicolumn{1}{c|}{MCAS} & \multicolumn{1}{c}{Res-50} &\multicolumn{1}{c}{Res-101} & \multicolumn{1}{c}{Inc-v3$_{ens}$} & \multicolumn{1}{c}{IncRes-v2$_{ens}$} &\multicolumn{1}{c}{ViT-B/16} & \multicolumn{1}{c}{PiT-S}  \\ \hline
 \multirow{3}{*}{IncRes-v2}  & \multicolumn{1}{c}{\XSolidBrush} &\multicolumn{1}{c|}{\XSolidBrush} & 79.50 & 74.40 & 69.40 & 62.30 & 49.90 & 52.40 \\ &\multicolumn{1}{c}{\checkmark} &\multicolumn{1}{c|}{\XSolidBrush} & 90.40 & 88.10 & 80.20 & 71.30 & 60.10 & 67.40\\&\multicolumn{1}{c}{\checkmark} &\multicolumn{1}{c|}{\checkmark} & \textbf{91.90} & \textbf{90.40} & \textbf{81.60} & \textbf{71.60} & \textbf{60.80} & \textbf{68.10}\\ \hline

 \multirow{3}{*}{Inc-v3} & \multicolumn{1}{c}{\XSolidBrush} &\multicolumn{1}{c|}{\XSolidBrush} & 79.90 & 75.50 & 62.10 & 43.20 & 46.00 & 44.30 \\ &\multicolumn{1}{c}{\checkmark} &\multicolumn{1}{c|}{\XSolidBrush} & 87.20 & 87.20 & 70.50 & 48.60 & 54.30 & 57.40\\&\multicolumn{1}{c}{\checkmark} &\multicolumn{1}{c|}{\checkmark} & \textbf{90.10} & \textbf{88.60} & \textbf{71.10} & \textbf{50.20} & \textbf{55.40} & \textbf{58.70}\\ \hline
  \end{tabular}
 \label{ablation}
 }
\end{table}

\subsection{Ablation Study}
\noindent {\bf{Ablation for each component.}}
To thoroughly examine how components affect our method, we perform ablation studies on various variants to target two normally trained models, two adversarially trained models and two diverse architectures using white-box surrogates Inc-v3 and IncRes-v2: (1) Without AF and MCAS, which equal to MI equipped with $N$ inner samplings and is a basic framework of most relevant works. We treat the variant as a foundation to demonstrate the improvement of each component. (2) Utilizing AF but only sampling uniformly. (3) Both AF and MCAS are taken into consideration. As presented in Table \ref{ablation}, several noteworthy observations emerge when comparing the results of various versions. Together with the results of previous studies, it is clear that the basic MI method combined with $N$ inner random sampling shows improved transferability compared to its traditional version, especially for models incorporating adversarially trained weights and different architectures. Given the surrogate model as Inc-v3, the average ASRs of MI with $N$ inner samplings are 35\% larger than MI \cite{MI} in Fig. \ref{fig_adv_loss}. This discovery confirms that our deliberate selection of the inner loop sampling design as it significantly mitigates the decline in attack effectiveness. Then when only integrating our proposed AF to generate adversarial examples on Inc-v3, our average ASRs can reach 4\%, 8\%, 14\%, 22\%, 28\% and 33\% more higher than PGN,  FGSRA, NCS, MuMoDIG, RAP and GI-MI. It is evident to show the superiority and effectiveness of our proposed AF. Finally, equipped with our MCAS, compared to the version with only AF, it is clear that our new performance can be further enhanced, demonstrating our success in the new neighborhood sampling. To further demonstrate the effectiveness of our proposed components, we conduct the ablation study on the defense model setting in \ref{app7}. Finally, the ablation study results about more hyperparameters are provided in \ref{app6}.

\section{Conclusion}
In this paper, motivated by the deceptive flatness problem, we propose a novel black-box gradient-based adversarial attack method to improve adversarial transferability. First, we feasibly fuse the dual-order flatness to construct the solution to deceptive flatness, and theoretically prove its assurance of adversarial transferability. Subsequently, by efficiently approximating our optimization objective based on the proposed dual-order information fusion, we develop the Adversarial Flatness Attack (AFA) with addressing the gradient sign alteration, which achieves the best attack performance on ImageNet-compatible dataset under various attack and defense settings. Additionally, we introduce MonteCarlo Adversarial Sampling (MCAS) to diversify inner sampling and further bolster adversarial transferability, resulting in less memory and more diverse sampling. These findings indicate that the current advanced models are still vulnerable and highlight the importance of researching more robust defense methods.

While this approach has yielded significant results on commonly used benchmarks, it does have specific limitations. Our method is built on CNN-like surrogate models and lacks direct consideration of attention shifts between CNN and Transformer architectures. Although this issue can be mitigated by input transformation strategies, further exploration from the perspective of gradient flatness is required. Hence, our future goal is to match the attack effectiveness on Transformer-based models with that on CNNs.

\section*{CRediT authorship contribution statement}
\textbf{Zhixuan Zhang}: Conceptualization, Methodology, Software, Writing - original draft. \textbf{Pingyu Wang}: Supervision, Conceptualization, Writing-review \& editing, Methodology. \textbf{Xingjian Zheng}: Writing - review \& editing. \textbf{Linbo Qing}: Writing - review \& editing. \textbf{Qi Liu}: Writing - review \& editing.

\section*{Declaration of competing interest}
The authors declare that they have no known competing financial interests or personal relations that could have appeared to influence their work reported in this paper.

\section*{Acknowledgments}
This work was supported by Science and Technology Projects of Xizang Autonomous Region (No. XZ202501ZY0064), the National Natural Science Foundation of China (No. 62301346 and No. 62202174), Sichuan Science and Technology Program (No. 2024NSFSC1424), Chengdu Technology Innovation Research and Development Project (No. 2024-YF05-00652-SN), Chengdu Major Technology Application Demonstration Project (No. 2023-YF09-00019-SN), the Fundamental Research Funds for the Central Universities (No. YJ202326 and No. 2025ZYGXZR053) and the GJYC program of Guangzhou (No. 2024D01J0081) and the ZJ program of Guangdong (No. 2023QN10X455).

\section*{Data and Code availability}
The data and code will be made available on request.

\bibliographystyle{elsarticle-num}
\bibliography{reference}
\clearpage

\appendix

\section{Proof of the relationships between the
adversarial loss on the surrogate model in the vicinity of
$x_{adv}$ with AZF and AFF}
\label{app1}

Given any adversarial example $x_{adv}$, the maximum radius of uniform sampling $\mathcal{U}$ centered at adversarial example $\xi$ and step size $\alpha$, we suppose that Eq. \ref{overall_objective} has the second-order gradient at least. On the one hand, according to the mean value theorem and Cauchy–Schwarz inequality, $\vartheta\sim\mathcal{U}(-\xi, \xi)$, there exists a constant 0 $\le$ c $\le$ 1, such that the adversarial loss on the surrogate model in the vicinity of $x_{adv}$ and AFF can be related as below:
\begin{align}
\begin{split}
\mathcal{L}_{src}^{adv}\left(x_{adv}+\vartheta\right)
&=\mathcal{L}_{src}^{adv}\left(x_{adv}\right)+(\bigtriangledown\mathcal{L}_{src}^{adv}\left(x_{adv}+c\vartheta\right))^{\top}\vartheta\\
&\le\mathcal{L}_{src}^{adv}\left(x_{adv}\right)+\|\bigtriangledown\mathcal{L}_{src}^{adv}\left(x_{adv}+c\vartheta\right))\|\|\vartheta\|\\
&\le\mathcal{L}_{src}^{adv}\left(x_{adv}\right)+\Psi_1\left(x_{adv}\right)
\end{split}
   \label{R1_inequal}
\end{align}
On the other hand, as expressed in Definition \ref{AZF}, we can easily reason the relationship of the adversarial loss on the surrogate model in the vicinity of $x_{adv}$ and AZF as follows:
\begin{align}
\begin{split}
\mathcal{L}_{src}^{adv}\left(x_{adv}+\vartheta\right)
&\le\mathcal{L}_{src}^{adv}\left(x_{adv}\right)+\Psi_0\left(x_{adv}\right)
\end{split}
   \label{R0_inequal}
\end{align}

Thus, by combining Eq. \ref{R1_inequal} and Eq. \ref{R0_inequal}, $0 \le \beta_f \le 1$, we can associate the adversarial loss on the surrogate model in the vicinity of $x_{adv}$ with AF, deriving Eq. \ref{R_inequal}.

\section{Proof of the relationship between AF and adversarial transferability}
\label{app2}
Subsequently, inspired by the generalization boundary theory in Theorem 4.2 \cite{nico++}, we further reformulate the adversarial transferability as below:
\begin{align}
\begin{split}
&\mathcal{L}_{tar}^{adv}\left(x_{adv}\right)\le\mathcal{L}_{src}^{adv}\left(x_{adv}\right)+BD_{upper},\\
&BD_{upper}= \sup\limits\|\mathcal{L}_{tar}(x^1, x^2)-\mathcal{L}_{src}(x^1, x^2) \|, x^1, x^2 \in \mathcal{B}_{\xi}(x_{adv})
\end{split}
   \label{DG_boundary}
\end{align}
where the term $\sup\limits\|\cdot\|$ corresponds to the upper bound of the discrepancy distance \cite{covari_shift}, which measures the covariate shift between adversarial losses of $x_{adv}$ on source model $\boldsymbol{M}_{src}$ and target model $\boldsymbol{M}_{tar}$, which are similar to the training domain and the unknown test domain in domain generalization respectively. Similarly, $\mathcal{B}_{\xi}\left(x_{adv}\right)$ indicates the neighborhood on point $x_{adv}$ at radius of $\xi$.

Finally, after integrating Eq. \ref{R_inequal} and Eq. \ref{DG_boundary}, we obtain the expanded inequality between AF and adversarial transferability:
\begin{align}
\begin{split}
&\mathbb{E}_{x_{adv}^{\prime}\sim \mathcal{B}_{\xi}(x_{adv})}[\mathcal{L}_{tar}^{adv}\left(x_{adv}^{\prime}\right)]\\
&\le\mathbb{E}_{x_{adv}^{\prime}\sim \mathcal{B}_{\xi}(x_{adv})}[\mathcal{L}_{src}^{adv}\left(x_{adv}^{\prime}\right)]+BD_{upper}\\
&\le\mathcal{L}_{src}^{adv}\left(x_{adv}\right)+\Psi^{AF}(x_{adv})+BD_{upper}
\end{split}
   \label{ATG1}
\end{align}

We can obtain Eq. \ref{ATG} by simplifying Eq. \ref{ATG1}.

\section{Proof of the gradient of Adversarial First-order Flatness}
\label{app3}
Based on Corollary 1 in \cite{PGN}, $\bigtriangledown_x\|\bigtriangledown_x\mathcal{L}^{adv}_{src}(x)\|=\frac{\bigtriangledown_{x^{\prime}}\mathcal{L}^{adv}_{src}(x^{\prime})|_{x^{\prime}=x+\triangle x}-\bigtriangledown_x\mathcal{L}^{adv}_{src}(x)}{\alpha}$. Further, we can first drive $x_2$, $x_3 = x_2+\triangle x_2$ and their corresponding gradients as follows:
\begin{align}
    \begin{split}  
    g_2^{\prime} &=-g_2=-\bigtriangledown_{x_2}\mathcal{L}(x_2,y^{gt};\boldsymbol{M}_{src}), \text{ where  }\\
    x_2&=x_0+\alpha\cdot\frac{\bigtriangledown_{x_0}\|\bigtriangledown_{x_0}\mathcal{L}_{src}^{adv}(x_0)\|}{\|\bigtriangledown_{x_0}\|\bigtriangledown_{x_0}\mathcal{L}_{src}^{adv}(x_0)\|\|}\\
    &=x_0-\alpha\cdot\frac{g_1-g_0}{\|g_1-g_0\|}
    \end{split}
    \label{x2_g_2}
\end{align}
\begin{align}
    \begin{split}  
    g_3^{\prime} &=-g_3=-\bigtriangledown_{x_3}\mathcal{L}(x_3,y^{gt};\boldsymbol{M}_{src}), \text{ where  }\\
    x_3&=x_2+\triangle x_2\\
    &=x_2-\alpha\cdot\frac{g_2}{\|g_2\|}
    \end{split}
    \label{x3_g_3xxx}
\end{align}

Combine above Eq. \ref{x2_g_2} and Eq. \ref{x3_g_3xxx}, gradient of $\Psi_1(x_{adv})$ can be approximated as below:
\begin{align}
    \begin{split}
        &\bigtriangledown\Psi_1(x_{adv})=h_1\approx \alpha\cdot\bigtriangledown\|\bigtriangledown\mathcal{L}_{src}^{adv}(x_2)\|\\
        &=\alpha\cdot\frac{\bigtriangledown_{x_3}\mathcal{L}^{adv}_{src}(x_3)-\bigtriangledown_{x_2}\mathcal{L}^{adv}_{src}(x_2)}{\alpha}\\
        &=g_3^{\prime}-g_2^{\prime}\\
        &=-(g_3-g_2)
    \end{split}
    \label{h_1xxxxx}
\end{align}

To this end, we prove Eq. \ref{h_1} as above all.

\section{Proof of the gradient of the approximated objective function of AFA}
\label{app4}
After joining Eq. \ref{h_0} and Eq. \ref{h_1}, our AFA can be written as:
\begin{align}
    \begin{split}
        &\min\limits_{x_{adv}\in \mathcal{B}_{\epsilon}(x)}[g_0^{\prime}+\lambda_f\Psi^{AF}(x_{adv})]\\
       \Rightarrow  &\min\limits_{x_{adv}\in \mathcal{B}_{\epsilon}(x)}[-g_0-\lambda_f(\beta_f(g_1-g_0)+(1-\beta_f)(g_3-g_2))]\\
       \Rightarrow  &\max\limits_{x_{adv}\in \mathcal{B}_{\epsilon}(x)}[g_0+\lambda_f(\beta_f(g_1-g_0)+(1-\beta_f)(g_3-g_2))]\\
    \end{split}
    \label{FAA_2xxxx}
\end{align}

Therefore, we prove Eq. \ref{FAA_2}.

\section{Visual experiment for the rationale of the combined Adversarial Flatness}
\label{Visual_AF}
\begin{figure*}[htb]
\centering
\includegraphics[width=8cm,height=5.1cm]{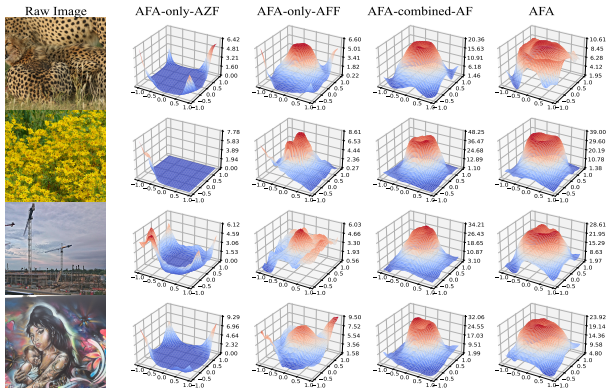}
\caption{Visualization of loss surfaces along two random directions for two randomly sampled adversarial examples on the surrogate Inc-v3. The center of each 2D graph corresponds to the adversarial example generated by our method, and three variants with only AZF, only AFF and only AF. The $x$ and $y$ axes represent the random noises added to $x_{adv}$ twice in succession. The $z$ axis indicates the loss value.}
\label{loss_surface_AF}
\end{figure*}

In fact, relying solely on extensive abstract mathematical derivations is insufficient to intuitively uncover the rationale behind the combined Adversarial Flatness. Therefore, we present a visualization of loss surfaces for adversarial examples generated by our method solving the problem of altered gradient sign and three variants with only AZF, only AFF and only AF. As shown in Fig. \ref{loss_surface_AF}, it is obvious that the variant with the combined AF can achieve a much broader and flatter top plane, which is more beneficial to adversarial transferability. It verifies the rationale of the combined Adversarial Flatness again. Moreover, it is suggested that the last column of results also demonstrates the effectiveness of our solution to the problem of altered gradient sign.

\section{Experiments based on different gradient approximation methods}
\label{experiment_gradient_approx}
As for the gradient approximation strategies used in our method, we mainly consider two aspects: precision and simplicity. It is easy to decide on the Finite Difference Method, especially the commonly used Forward Difference Method (FDM), which is also utilized in our method. However, there are other variants in the family of the Finite Difference Method, such as the Center Difference Method (CDM) and the Backward Difference Method (BDM). BDM is similar to FDM, although CDM is slightly more complex. We compare our method with its variants based on the two gradient approximation methods in terms of ASRs on different models. As shown in Table \ref{single_approx_gredient}, it is clear that the variety of gradient approximation methods significantly influences their effectiveness. Furthermore, the FDM-based implementation achieves superior performance compared to the other two implementations, which demonstrates the feasibility of the FDM in our method.

\begin{table}[!t]
\centering
\caption{The untargeted attack success rates based on different gradient approximation methods. \textbf{BOLD} indicates the best.}
\resizebox{12cm}{1.cm}{
\begin{tabular}{c|c|cccccccccccc}
		\hline
		\multicolumn{1}{c|}{Model}  
		 &\multicolumn{1}{c|}{Approx} & \multicolumn{1}{c}{Inc-v4} &\multicolumn{1}{c}{Dense-121} & \multicolumn{1}{c}{VGG-19} & \multicolumn{1}{c}{PASNet-L} & \multicolumn{1}{c}{MLP-mixer} & \multicolumn{1}{c}{ResMLP} & \multicolumn{1}{c}{Inc-v3$_{ens}$} &
   \multicolumn{1}{c}{Inc-v4$_{ens}$}  & \multicolumn{1}{c}{InRes-v2$_{ens}$} & \multicolumn{1}{c}{NRP} & \multicolumn{1}{c}{JPEG} & \multicolumn{1}{c}{Bit-Red}  \\ \hline

 \multirow{3}{*}{Inc-v3} &\multicolumn{1}{c|}{BDM} & 91.00 & 87.20 & 87.80 & 81.20 & 60.70 & 61.90 & 63.50 & 65.00 & 43.10 & 72.40 & 80.20 & 84.80 \\&\multicolumn{1}{c|}{CDM} & 94.20 & 90.00 & 90.80 & 85.80 & 63.30 & 67.50 & 68.90 & 69.50 & 46.50 & 73.80 & 84.50 & 88.00\\&\multicolumn{1}{c|}{\bf{FDM}} & \textbf{95.20} & \textbf{91.90} & \textbf{91.50} & \textbf{87.40}& \textbf{65.30}& \textbf{72.30}& \textbf{73.00}& \textbf{73.20}& \textbf{50.90} & \textbf{74.80}& \textbf{87.60}& \textbf{89.80}\\ \hline
  
 \multirow{3}{*}{Swin-T} &\multicolumn{1}{c|}{BDM} & 81.90 & 93.30 & 93.50 &  89.40 & 91.20 & 97.10 & 73.30 &71.70 &63.10 & 89.00 & 90.10 & 90.00\\&\multicolumn{1}{c|}{CDM} & 81.70 & 93.10 & 93.70 & 88.90 & 90.80 & 96.90 & 72.10 & 71.30& 61.50 & 89.30 & 90.00 & 89.70 \\&\multicolumn{1}{c|}{\bf{FDM}} & \textbf{85.70} & \textbf{94.30} & \textbf{94.70} & \textbf{89.90}& \textbf{93.20}& \textbf{97.60}& \textbf{75.40}& \textbf{74.50}& \textbf{65.30} & \textbf{90.80}& \textbf{90.60}& \textbf{91.10}\\ \hline
 
  \end{tabular}
 \label{single_approx_gredient}
 }
\end{table}

\begin{table}[!t]
\centering
\caption{The untargeted attack success rates based on transformer surrogates among various gradient-based
attacks in the normal model setting. \textbf{BOLD} indicates the best.}
\resizebox{6.3cm}{2.6cm}{
\begin{tabular}{c|c|cccc}
		\hline
		\multicolumn{1}{c|}{Model}  
		 &\multicolumn{1}{c|}{Attack} & \multicolumn{1}{c}{Inc-v3} & \multicolumn{1}{c}{Res-101} &\multicolumn{1}{c}{Dense-121} & \multicolumn{1}{c}{VGG-19}   \\ \hline

 \multirow{7}{*}{ViT-B/16} &\multicolumn{1}{c|}{RAP} & 73.50 & 82.90 & 84.00 & 83.80  \\&\multicolumn{1}{c|}{NCS} & 75.20 & 78.90 & 81.10 &  79.50 \\&\multicolumn{1}{c|}{PGN} & 79.80 & 84.40 & 85.80 & 84.60  \\&\multicolumn{1}{c|}{MuMoDIG} & 71.30 & 76.60 & 80.40 &  77.90 \\&\multicolumn{1}{c|}{GI-MI} & 66.60 & 75.60 & 77.80 & 80.00 \\ &\multicolumn{1}{c|}{FGSRA} & 74.80 & 78.90 & 81.70 & 79.40 \\&\multicolumn{1}{c|}{\bf{AFA}} & \textbf{84.10} & \textbf{88.20} & \textbf{90.80} & \textbf{89.80} \\ \hline

 \multirow{7}{*}{Swin-T} &\multicolumn{1}{c|}{RAP} & 53.80 & 67.70 & 69.80 & 77.30  \\&\multicolumn{1}{c|}{NCS} & 81.80 & 86.50 & 89.40 & 90.30  \\&\multicolumn{1}{c|}{PGN} & 85.20 & 89.70  & 91.10 & 91.90  \\&\multicolumn{1}{c|}{MuMoDIG} & 56.20 & 61.90 & 68.70 & 72.30 \\&\multicolumn{1}{c|}{GI-MI} & 49.50 & 60.70 & 64.60 & 72.40 \\ &\multicolumn{1}{c|}{FGSRA} & 83.60 & 86.80 & 89.90 & 90.60 \\&\multicolumn{1}{c|}{\bf{AFA}} & \textbf{87.10} & \textbf{{91.70}} & \textbf{94.30} & \textbf{{94.70}} \\
 \hline
  \end{tabular}
 \label{single_transformer_normal}
 }

\end{table}

\begin{table}[!t]
\centering
\caption{The untargeted attack success rates based on transformer surrogates among various gradient-based attacks in the diverse model architectures setting. \textbf{BOLD} indicates the best.}
\resizebox{6.3cm}{2.6cm}{
\begin{tabular}{c|c|cccc}
		\hline
		\multicolumn{1}{c|}{Model}  
		 &\multicolumn{1}{c|}{Attack} & \multicolumn{1}{c}{MobilNet} & \multicolumn{1}{c}{PASNet-L} &\multicolumn{1}{c}{MLP-mixer} & \multicolumn{1}{c}{ResMLP}   \\ \hline

 \multirow{7}{*}{ViT-B/16} &\multicolumn{1}{c|}{RAP} & 90.20 & 82.30 & 92.30 & 97.10  \\&\multicolumn{1}{c|}{NCS} & 86.70 & 79.10 & 92.60 & 97.50  \\&\multicolumn{1}{c|}{PGN} & 89.60 & 84.30 & 94.00 & 98.10  \\&\multicolumn{1}{c|}{MuMoDIG} & 86.00 & 79.10 & 89.10 & 96.70 \\&\multicolumn{1}{c|}{GI-MI} & 87.60 & 75.50 & 89.80 & 96.70 \\ &\multicolumn{1}{c|}{FGSRA} & 86.30 & 79.20 & 91.30 & 96.30 \\&\multicolumn{1}{c|}{\bf{AFA}} & \textbf{91.70} & \textbf{{88.60}} & \textbf{96.70} & \textbf{{99.50}} \\ \hline

 \multirow{7}{*}{Swin-T} &\multicolumn{1}{c|}{RAP} & 85.00 & 69.30 & 71.40 & 72.50  \\&\multicolumn{1}{c|}{NCS} & 91.40 & 84.90 & 85.90 & 94.10  \\&\multicolumn{1}{c|}{PGN} & 93.80 & 88.40 & 91.00 & 95.20  \\&\multicolumn{1}{c|}{MuMoDIG} & 80.70 & 67.30 & 68.30 & 75.80 \\&\multicolumn{1}{c|}{GI-MI} & 81.60 & 63.00 & 67.20 & 69.30 \\ &\multicolumn{1}{c|}{FGSRA} & 92.90 & 86.20 & 88.10 & 94.80 \\&\multicolumn{1}{c|}{\bf{AFA}} & \textbf{94.70} & \textbf{89.90} & \textbf{93.20} & \textbf{97.60} \\
 \hline
 
  \end{tabular}
 \label{single_transformer_arch}
 }
\end{table}

\section{Experiments based on transformer-based surrogates and attacking normally trained models}
\label{single_transformer}
For comprehensively validating our performance on a single model, we also consider the advanced transformer-based models as the surrogate models. Because traditional CNNs differ from Transformers in terms of information extraction, it is essential to conduct these experiments to fully explore our superiority. Wherein, we set the surrogates as ViT-B/16 and Swin-T. And the experiments are divided into two sub-experiments: (1) attacking the normal models with CNN architectures and (2) attacking the models with diverse architectures. As for the target models, the former includes Inc-v3, Res-101, Dense-121 and VGG-19 while the latter employs MobilNet, PASNet-L, MLP-mixer and ResMLP. When comparing Table \ref{single_transformer_normal} with Table \ref{single_diverse_arch}, it is evident that when adversarial examples generated on Inc-v3 are used to attack ViT-B/16, the ASRs of all methods are significantly lower than those observed when attacking Inc-v3 from ViT-B/16. And, the same phenomenon also exists in Table \ref{single_transformer_arch}. In particular, for MLP-Mixer and ResMLP, attacks launched by ViT-B/16 and Swin-T achieve significantly higher average ASRs compared to those from CNNs. These results may derive from the difference in the information extraction process between CNNs and Transformers. However, our method still obtains the best performance in both Table \ref{single_transformer_normal} and Table \ref{single_transformer_arch}.

\begin{table}[!t]
\centering
\caption{The untargeted attack success rates based on transformer surrogates among various gradient-based attacks in the adversarially trained model setting. \textbf{BOLD} indicates the best.}
\resizebox{6.3cm}{2.6cm}{
\begin{tabular}{c|c|cccc}
		\hline
		\multicolumn{1}{c|}{Model}  
		 &\multicolumn{1}{c|}{Attack} & \multicolumn{1}{c}{Inc-v3*} & \multicolumn{1}{c}{Inc-v3$_{ens}$} &
   \multicolumn{1}{c}{Inc-v4$_{ens}$}  & \multicolumn{1}{c}{InRes-v2$_{ens}$}   \\ \hline
 \multirow{7}{*}{ViT-B/16} &\multicolumn{1}{c|}{RAP} & 61.30 & 50.90 & 52.50 & 39.80  \\&\multicolumn{1}{c|}{NCS} & 65.40 & 62.90 & 62.80 &  56.10 \\&\multicolumn{1}{c|}{PGN} & 73.20 & 71.20 & 71.30 & 63.90  \\&\multicolumn{1}{c|}{MuMoDIG} & 63.40 & 58.10 & 59.10 &  50.30 \\&\multicolumn{1}{c|}{GI-MI} & 56.80 & 52.60 & 50.90 & 42.10 \\ &\multicolumn{1}{c|}{FGSRA} & 67.90 & 66.50 & 66.20 & 58.70 \\&\multicolumn{1}{c|}{\bf{AFA}} & \textbf{79.30} & \textbf{75.40} & \textbf{75.70} & \textbf{69.40} \\ \hline

 \multirow{7}{*}{Swin-T} &\multicolumn{1}{c|}{RAP} & 36.40 & 30.80 & 32.00 & 21.00  \\&\multicolumn{1}{c|}{NCS} & 66.10 & 66.60 & 57.90 & 62.00  \\&\multicolumn{1}{c|}{PGN} & 78.00  & 74.50 & 72.10 & 64.80  \\&\multicolumn{1}{c|}{MuMoDIG} & 39.30 & 34.10 & 34.20 & 25.70 \\&\multicolumn{1}{c|}{GI-MI} & 31.30 & 29.40 & 30.50 & 21.00 \\ &\multicolumn{1}{c|}{FGSRA} & 75.40 & 71.10 & 71.20 & 61.90 \\&\multicolumn{1}{c|}{\bf{AFA}} & \textbf{80.00} & \textbf{75.40} & \textbf{74.50} & \textbf{65.30} \\
 \hline
  \end{tabular}
 \label{single_transformer_adv}
 }
\end{table}

\begin{table}[!t]
\centering
\caption{The untargeted attack success rates based on transformer surrogates among various gradient-based
attacks in the defense model setting. \textbf{BOLD} indicates the best.}
\resizebox{6.3cm}{2.6cm}{

\begin{tabular}{c|c|cccc}
		\hline
		\multicolumn{1}{c|}{Model}  
		 &\multicolumn{1}{c|}{Attack} & \multicolumn{1}{c}{HGD} & \multicolumn{1}{c}{NRP} &
   \multicolumn{1}{c}{JPEG}  & \multicolumn{1}{c}{Bit-Red}   \\ \hline
 \multirow{7}{*}{ViT-B/16} &\multicolumn{1}{c|}{RAP} & 48.40 & 75.60 & 79.30 & 78.40  \\&\multicolumn{1}{c|}{NCS} & 61.00 & 80.40 & 79.80 & 76.00  \\&\multicolumn{1}{c|}{PGN} & 69.20 & 85.20 & 84.90 & 81.70  \\&\multicolumn{1}{c|}{MuMoDIG} & 55.90 & 79.90 & 75.20 & 73.40 \\&\multicolumn{1}{c|}{GI-MI} & 48.90 & 76.50 & 73.10 & 72.30 \\ &\multicolumn{1}{c|}{FGSRA} & 64.00 & 83.00 & 78.20 & 76.90 \\&\multicolumn{1}{c|}{\bf{AFA}} & \textbf{73.20} & \textbf{{86.00}} & \textbf{87.60} & \textbf{{86.50}} \\ \hline

 \multirow{7}{*}{Swin-T} &\multicolumn{1}{c|}{RAP} & 20.90 & 80.10 & 64.90 & 66.40  \\&\multicolumn{1}{c|}{NCS} & 68.00 & 85.60 & 84.40 & 84.80  \\&\multicolumn{1}{c|}{PGN} & 75.30 & 89.00 & 89.60 & 89.60  \\&\multicolumn{1}{c|}{MuMoDIG} & 30.70 & 77.10 & 60.10 & 62.70 \\&\multicolumn{1}{c|}{GI-MI} & 23.20 & 76.50 & 56.80 & 60.70 \\ &\multicolumn{1}{c|}{FGSRA} & 72.20 & 86.50 & 85.90 & 86.90 \\&\multicolumn{1}{c|}{\bf{AFA}} & \textbf{76.60} & \textbf{90.80} & \textbf{90.60} & \textbf{91.10} \\
 \hline
  \end{tabular}

 \label{single_transformer_defend}
 }
\end{table}

\begin{figure}[htbp]
\centering
\includegraphics[width=0.53\linewidth]{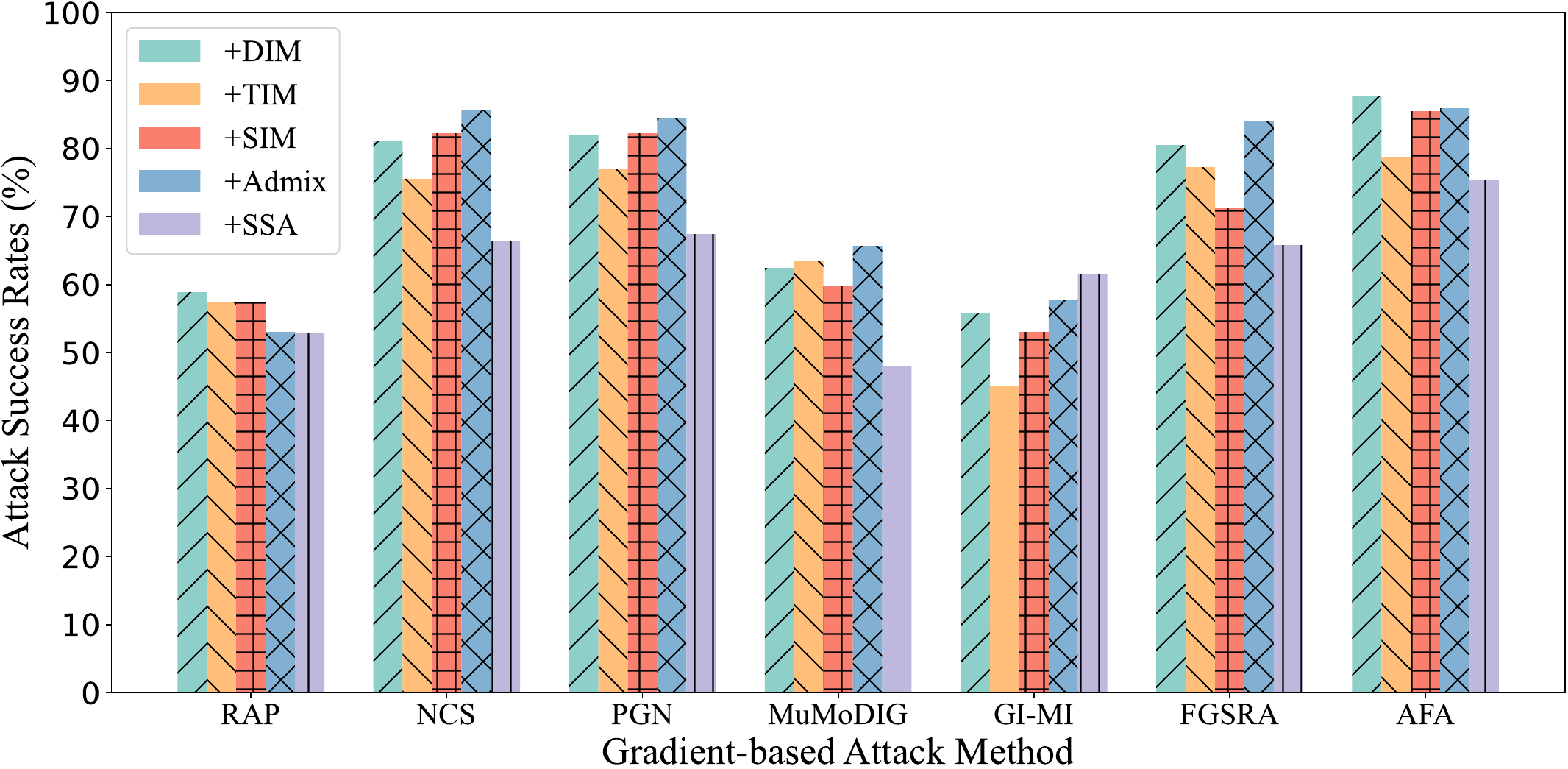}
\caption{Comparison of the average untargeted attack success rates between various gradient-based
attacks with the various augmentation-based settings. The white-box source model is Res-50.}
\label{aug_comparison}
\end{figure}

\section{Experiments based on transformer-based surrogates and attacking defense models}
\label{defend_transformer}
Meanwhile, we also test our effectiveness on the transformer-based models for the defense models. Except for the defense strategies as adversarial training, HGD, NRP, JPEG and Bit-Red, the other settings are identical to \ref{single_transformer}. We group this validation as two experiments: (1) aiming at the models equipped with the adversarial training strategy, including Inc-v3*, Inc-v3$_{ens}$, Inc-v4$_{ens}$ and InRes-v2$_{ens}$. Then, (2) attacking the models with the other defense strategies. In terms of attacking models with adversarial training, compared to Table \ref{single_adv}, our method gains a 10.30\% increase in ASRs while maintaining the best effectiveness. In addition, in Table \ref{single_transformer_defend}, there exists a 2\% attack capability increment under the defense strategies compared with Table \ref{single_defense}. And, our proposed method can still pose the state-of-the-art performance in Table \ref{single_transformer_defend}.

\begin{figure}[]
    \centering
    \includegraphics[width=0.3\linewidth]{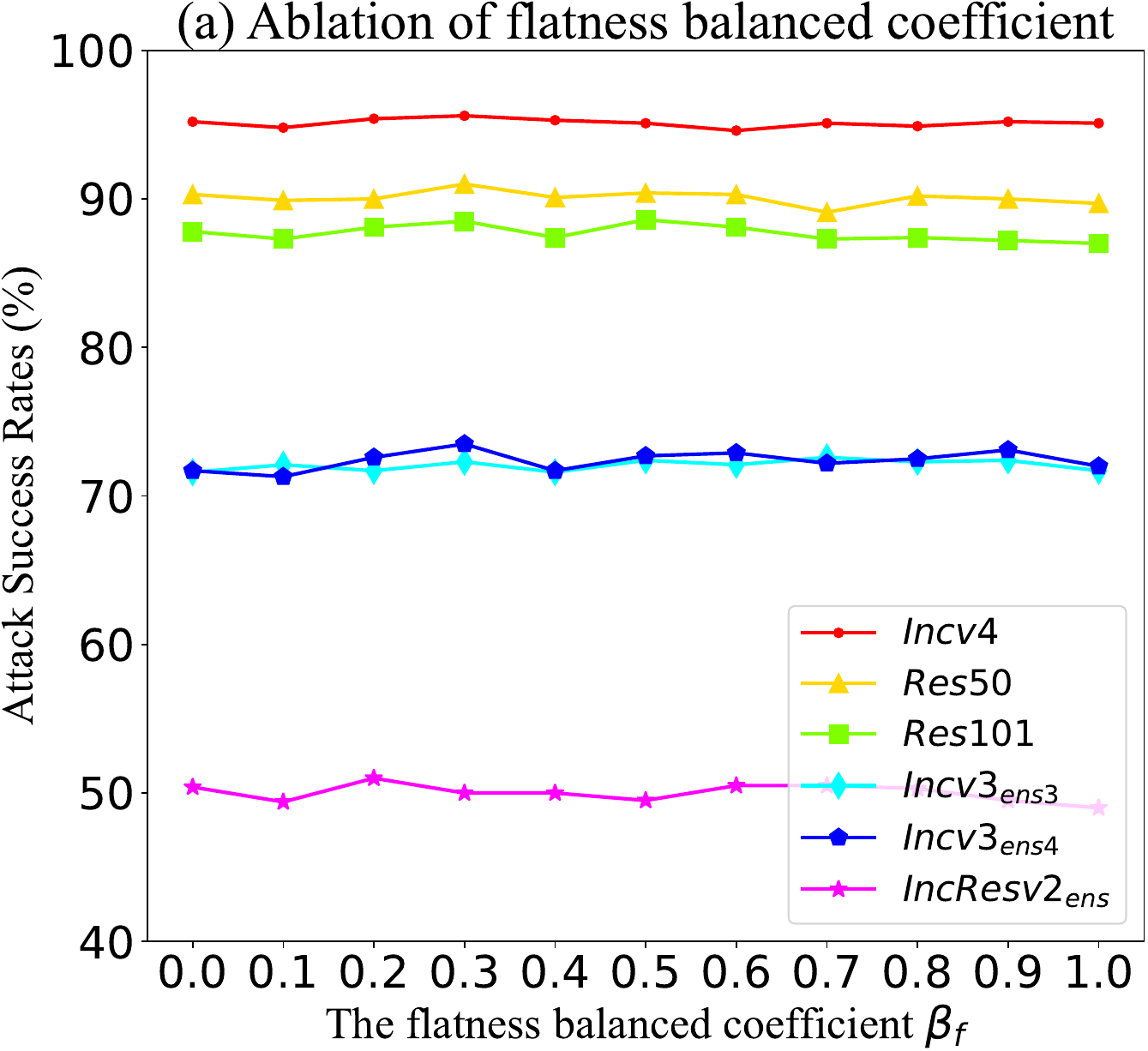}
    \hspace{-0.1mm}
    \includegraphics[width=0.3\linewidth]{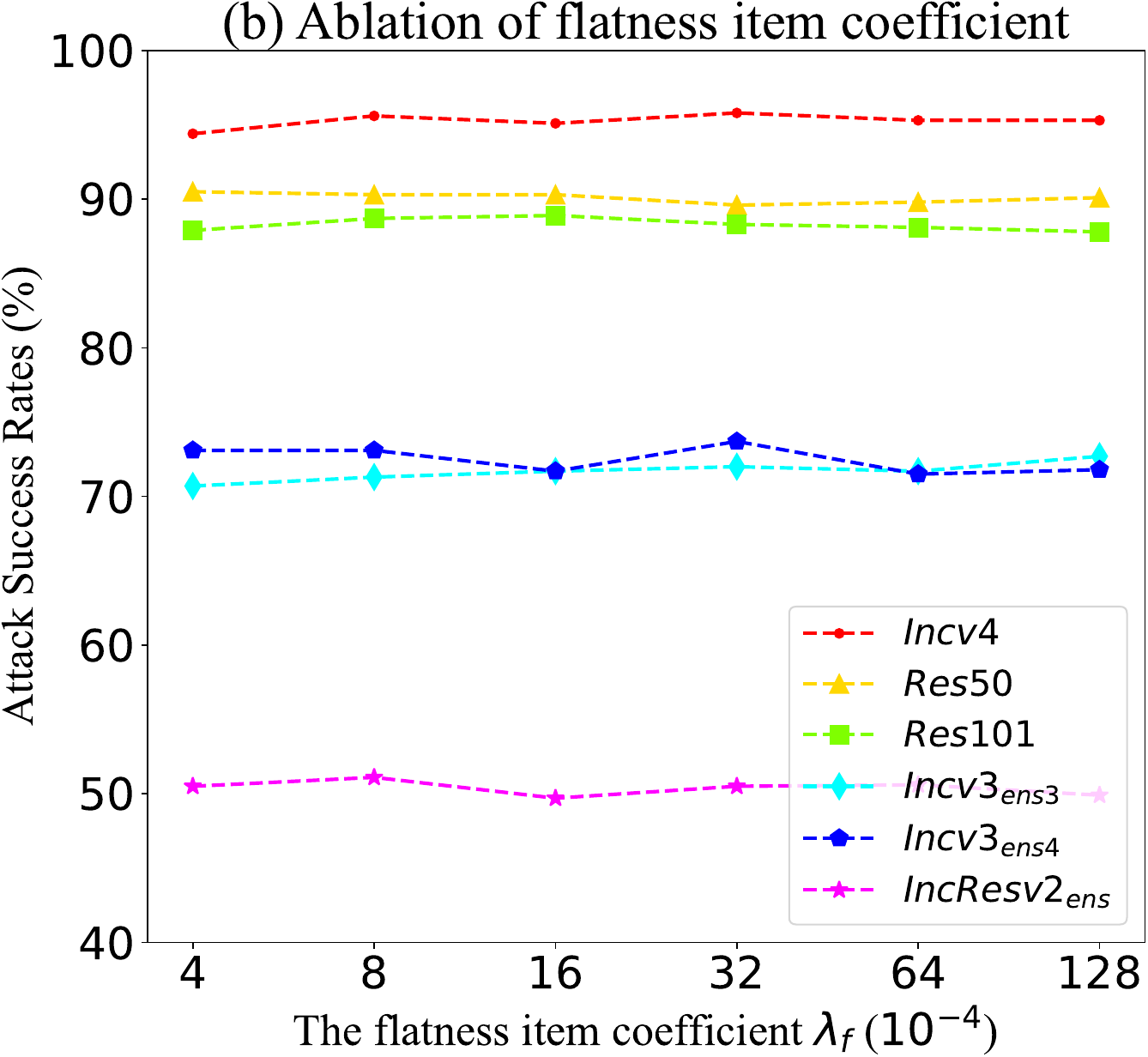}
    \hspace{-0.1mm}
    \includegraphics[width=0.3\linewidth]{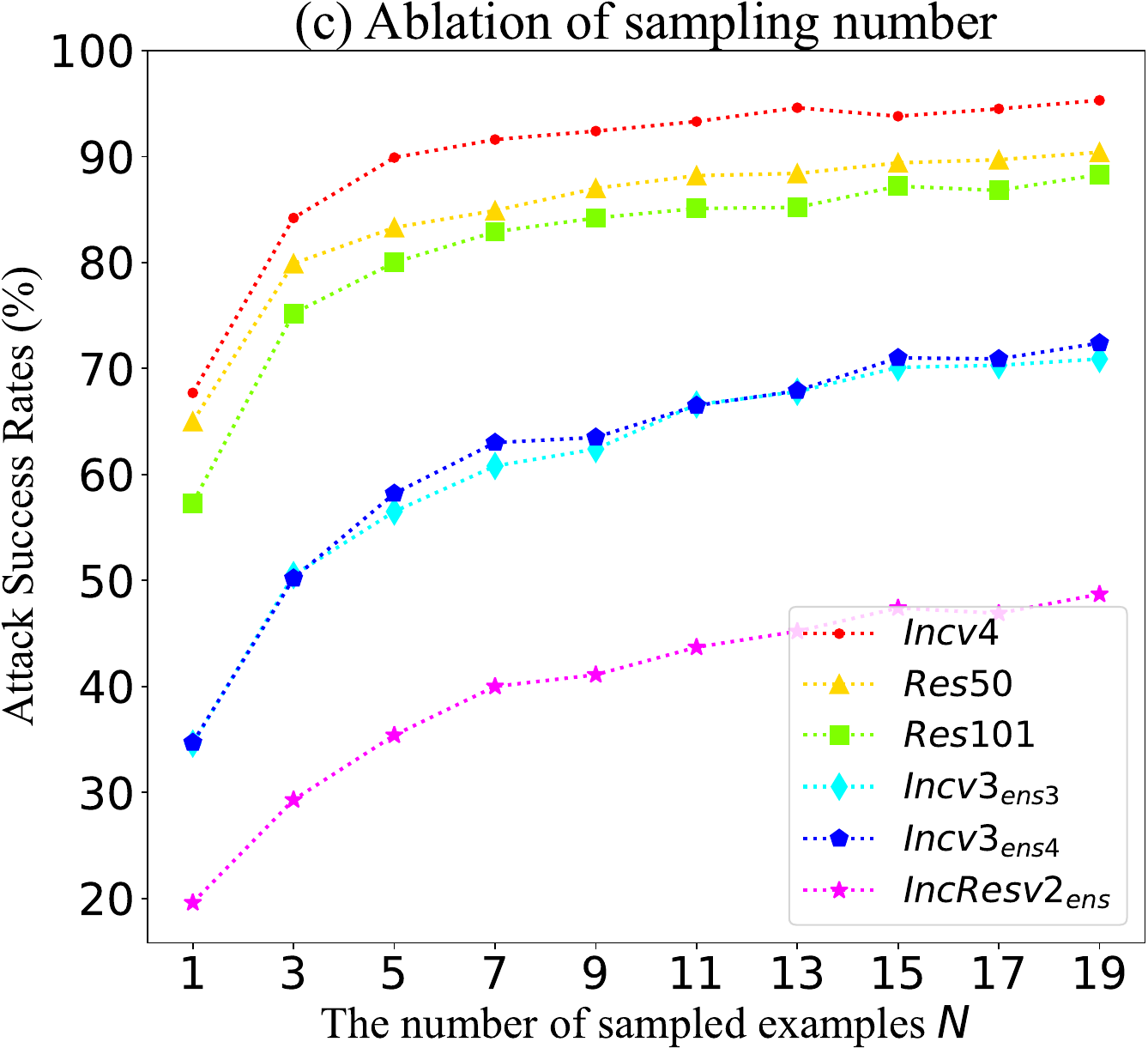}
    \centering
\vskip 5pt    
    \centering
    \includegraphics[width=0.3\linewidth]{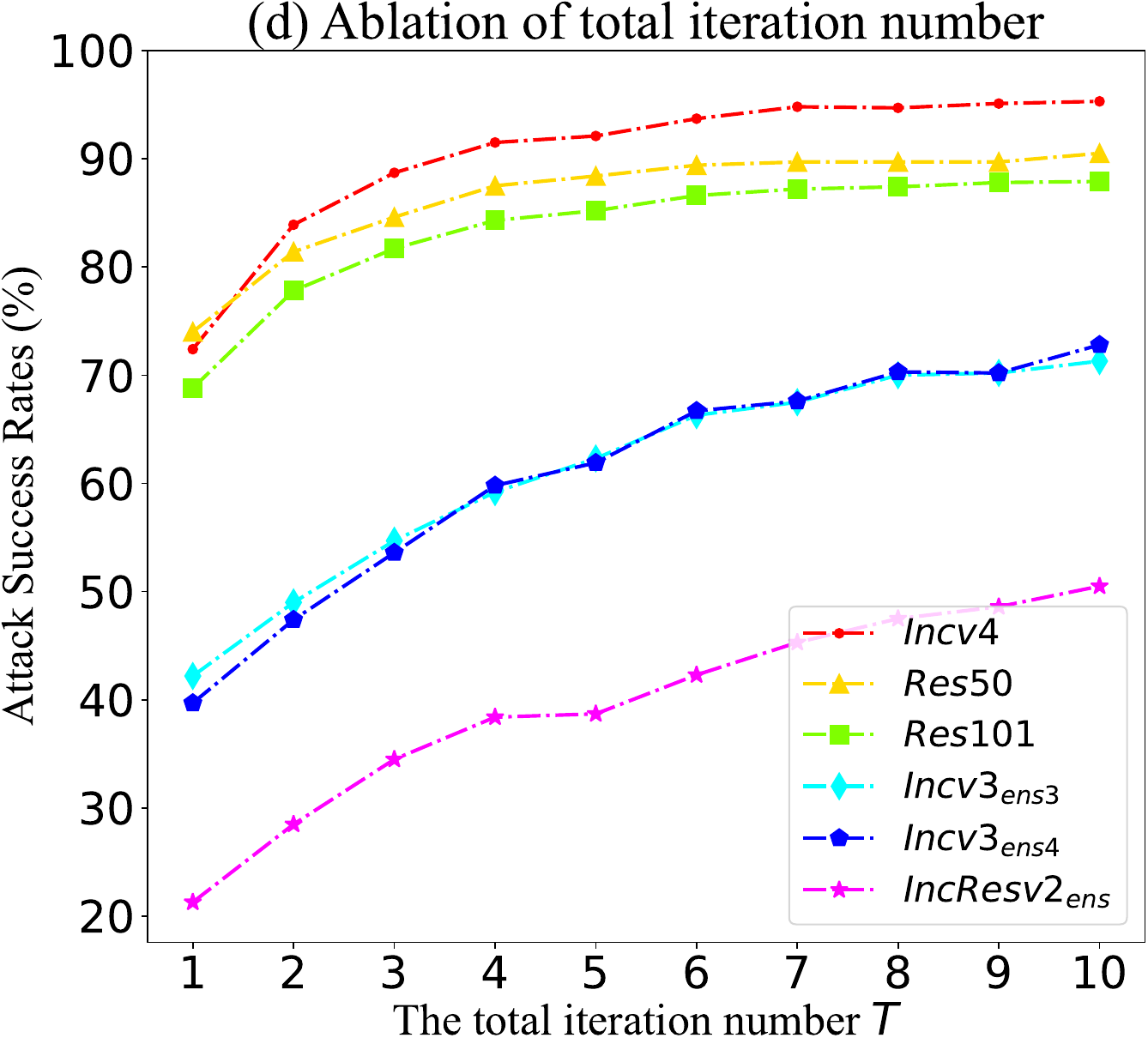}
    \hspace{-0.1mm}
    \includegraphics[width=0.3\linewidth]{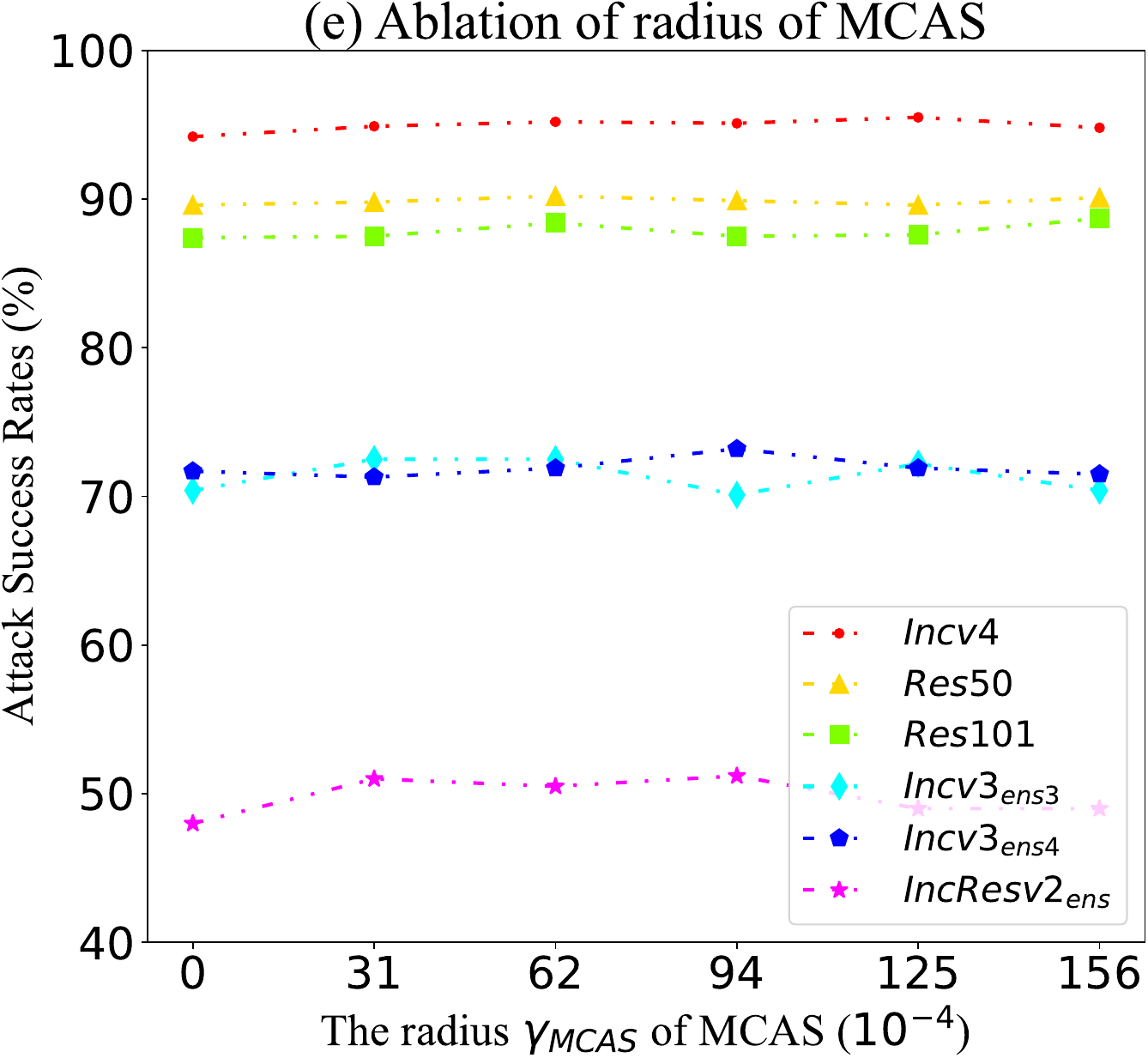}
    \hspace{-0.1mm}
    \includegraphics[width=0.3\linewidth]{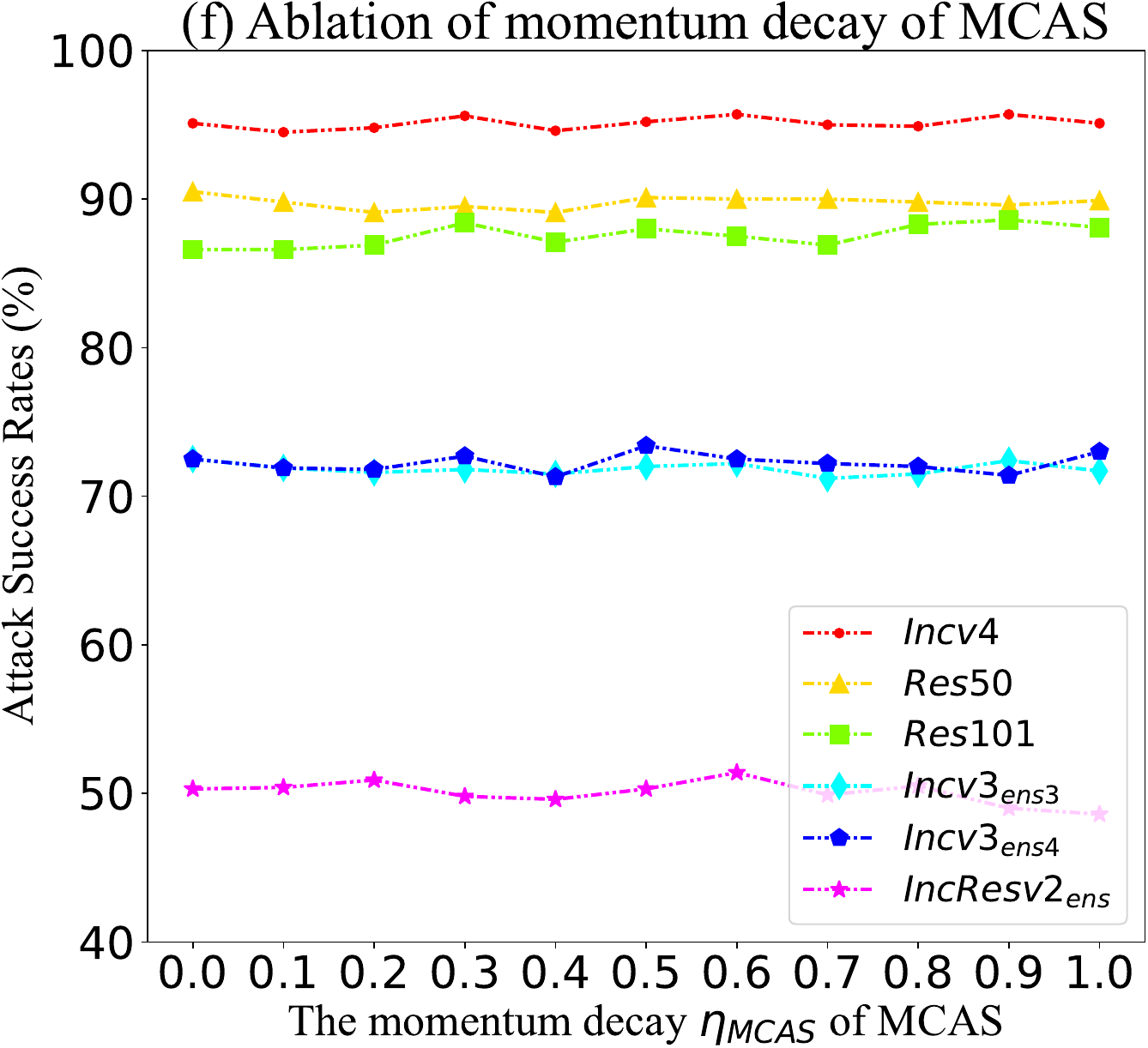}
    \caption{Untargeted attack success rates (\%) in six black box models with different hyperparameter ablation settings. The adversarial examples are generated by our AFA on Inc-v3.}
    \label{ablation_hyperparam}
\end{figure}

\section{Additional experiments combined with input transformation attacks}
\label{app5}
We also display the results of the additional experiments combined with input transformation attacks. In Fig. \ref{aug_comparison}, it further shows that these input transformation-based methods can benefit more from our AFA than other baselines. Thereby, combined with current input transformation-based methods, our approach can further improve the success rate of transfer-based black-box attacks.

\section{Ablation for the more hyperparameters}
\label{app6}
Furthermore, we conduct ablation experiments to analyze the impact of these hyperparameters, such as the flatness balanced coefficient $\beta_{f}$, the flatness item coefficient $\lambda_{f}$, the number of sampled examples $N$, the total iteration number $T$, the ablations towards the radius $\gamma_{MCAS}$ and the momentum decay $\eta_{MCAS}$ of MCAS. As default, they are set as 0.5, 0.0032, 20, 10, 0.0094 and 0.9 respectively. In this validation, we attempt different values of one parameter while keeping the other three parameters unchanged. As shown in Fig. \ref{ablation_hyperparam} (c) and (d), the attack ability of our proposed AFA is influenced by $N$ and $T$ more than the two other hyperparameters. As $N$ and $T$ increased, the transferability exhibits rapid improvement until $N=13$ and $T=7$, after which it gradually converges for normally trained models. Notably, when $N\ge13$ and $T\ge7$, a slight performance improvement can still be achieved by increasing the number of sampled examples and the number of iterations in our AFA method. Meanwhile, as depicted in Fig. \ref{ablation_hyperparam} (a), the best transferability to other models occurs at the intermediate value of the hyperparameter $\beta_f$, rather than at $\beta_f=0$ or $\beta_f=1$, which correspond to using only AFF or AZF independently. It confirms the efficacy of our dual-order information fusion. Further, Fig. \ref{ablation_hyperparam} (b) suggests that the diverse range of the flatness item coefficient $\lambda_f$ has varying effects on adversarial transferability. Finally, Fig. \ref{ablation_hyperparam} (e) presents that the transferable ASRs among the adv-trained models might decrease with the radius $\gamma_{MCAS}$ of MCAS larger than 0.0094. As shown in Fig. \ref{ablation_hyperparam} (f), the smaller momentum decay $\eta_{MCAS}$ of MCAS slightly affects the adversarial transferability, but $\eta_{MCAS}=0.9$ is a suitable choice.

\section{Ablation of the hyperparameters on robustness and applicability}
\label{ablation_hyperparam_robustness_application}
\begin{figure}[]
    \centering
    \includegraphics[width=0.30\linewidth]{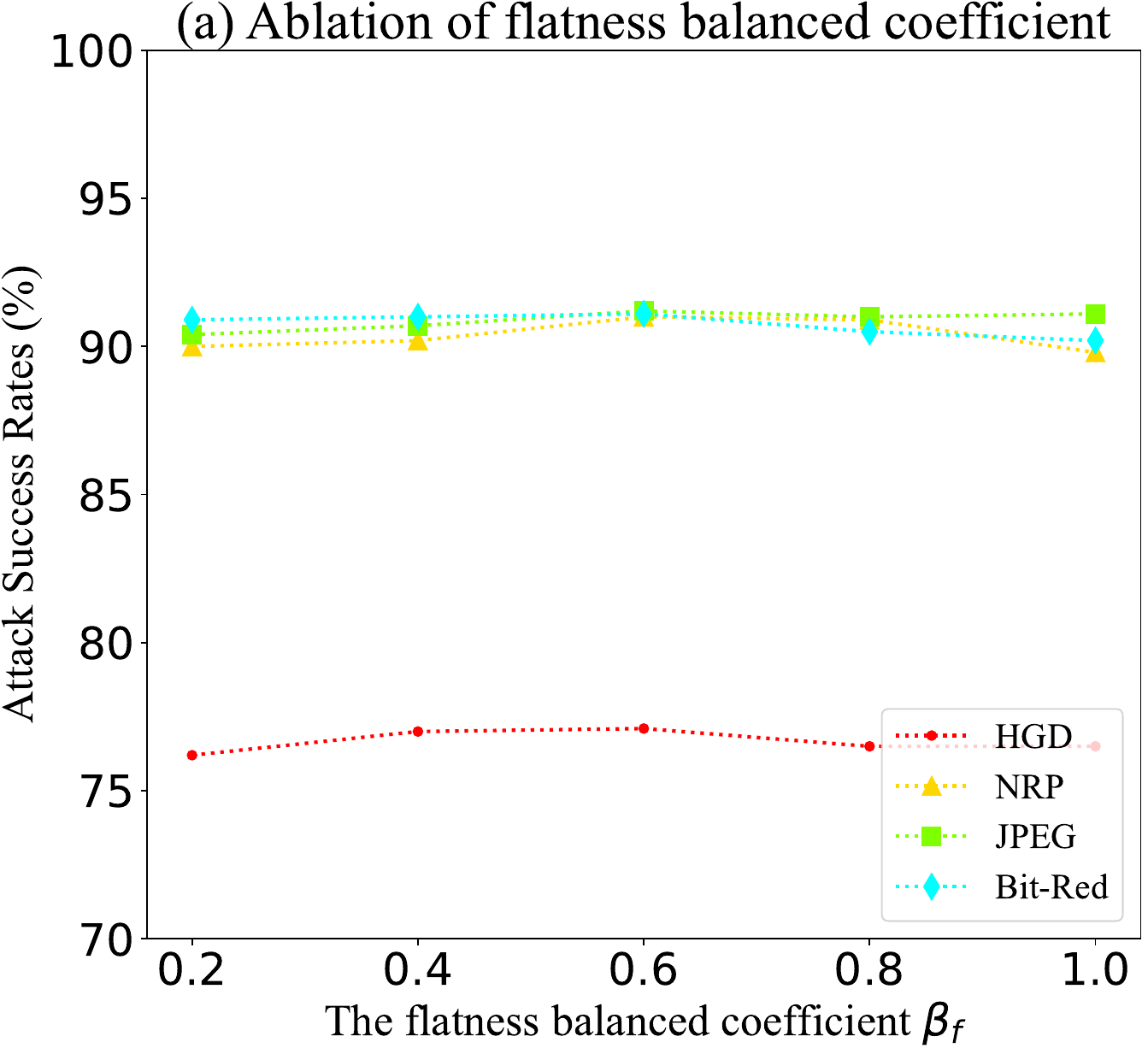}
    \hspace{-0.1mm}
    \includegraphics[width=0.30\linewidth]{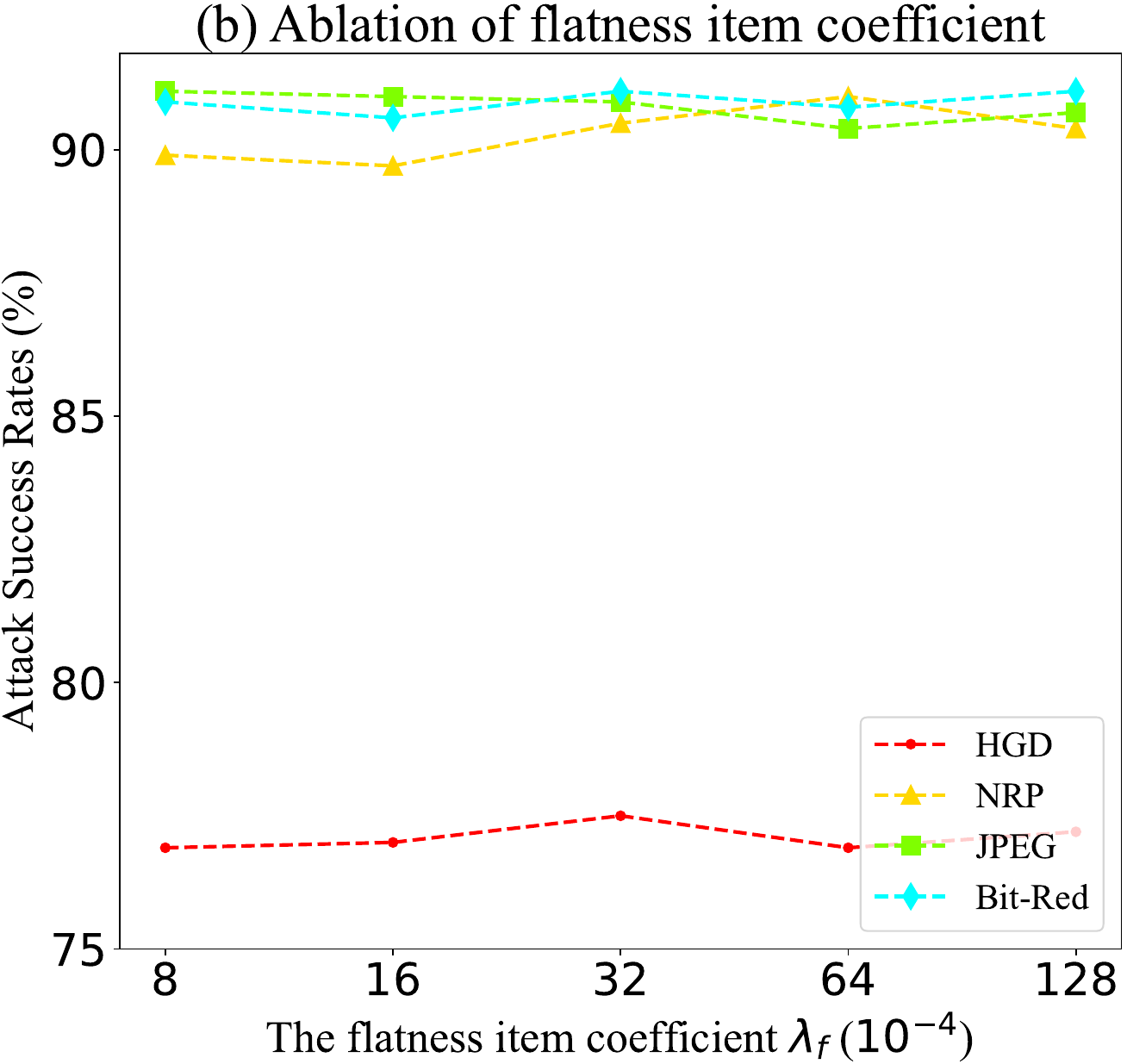}
    \vskip 5pt    
    \centering
    \includegraphics[width=0.30\linewidth]{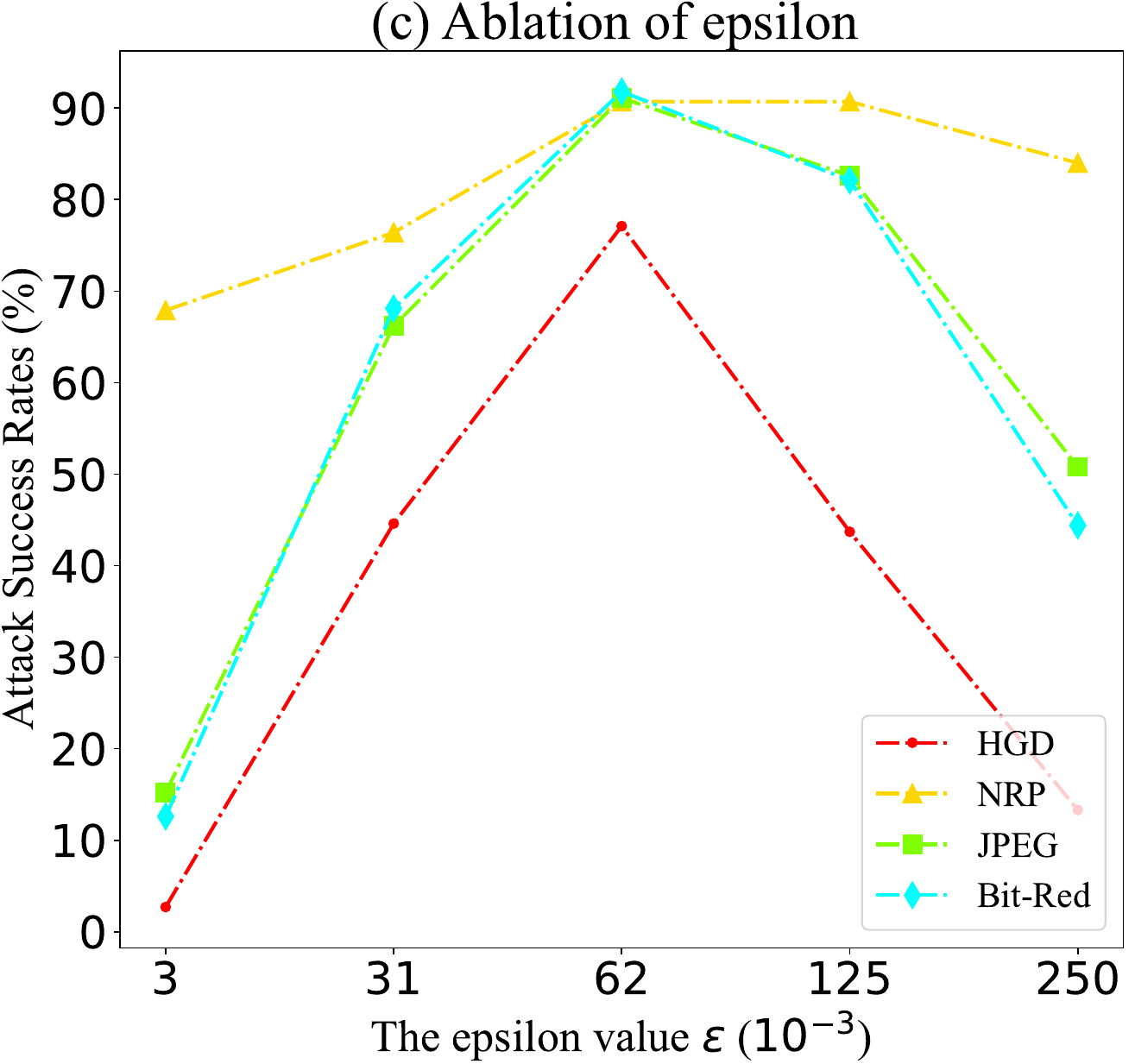}
    \hspace{-0.2mm}
    \includegraphics[width=0.30\linewidth]{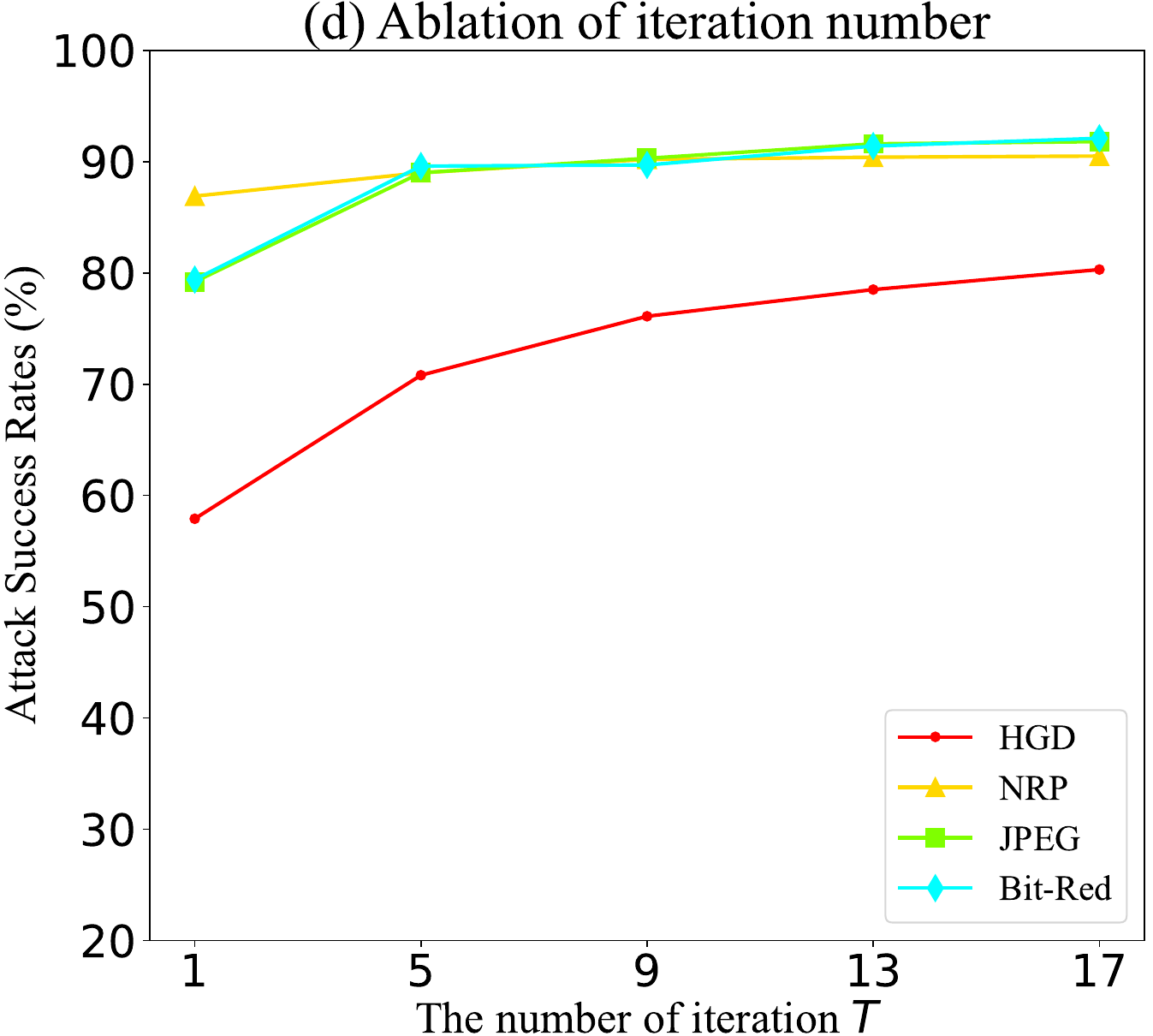}
    
    \caption{Untargeted attack success rates (\%) under four defense models with different hyperparameter ablation settings. The adversarial examples are generated by our AFA on Swin-T.}
    \label{ablation_hyperparam_robustness_applicability}
\end{figure}

\begin{figure}[htbp]
\centering
\includegraphics[width=1\linewidth]{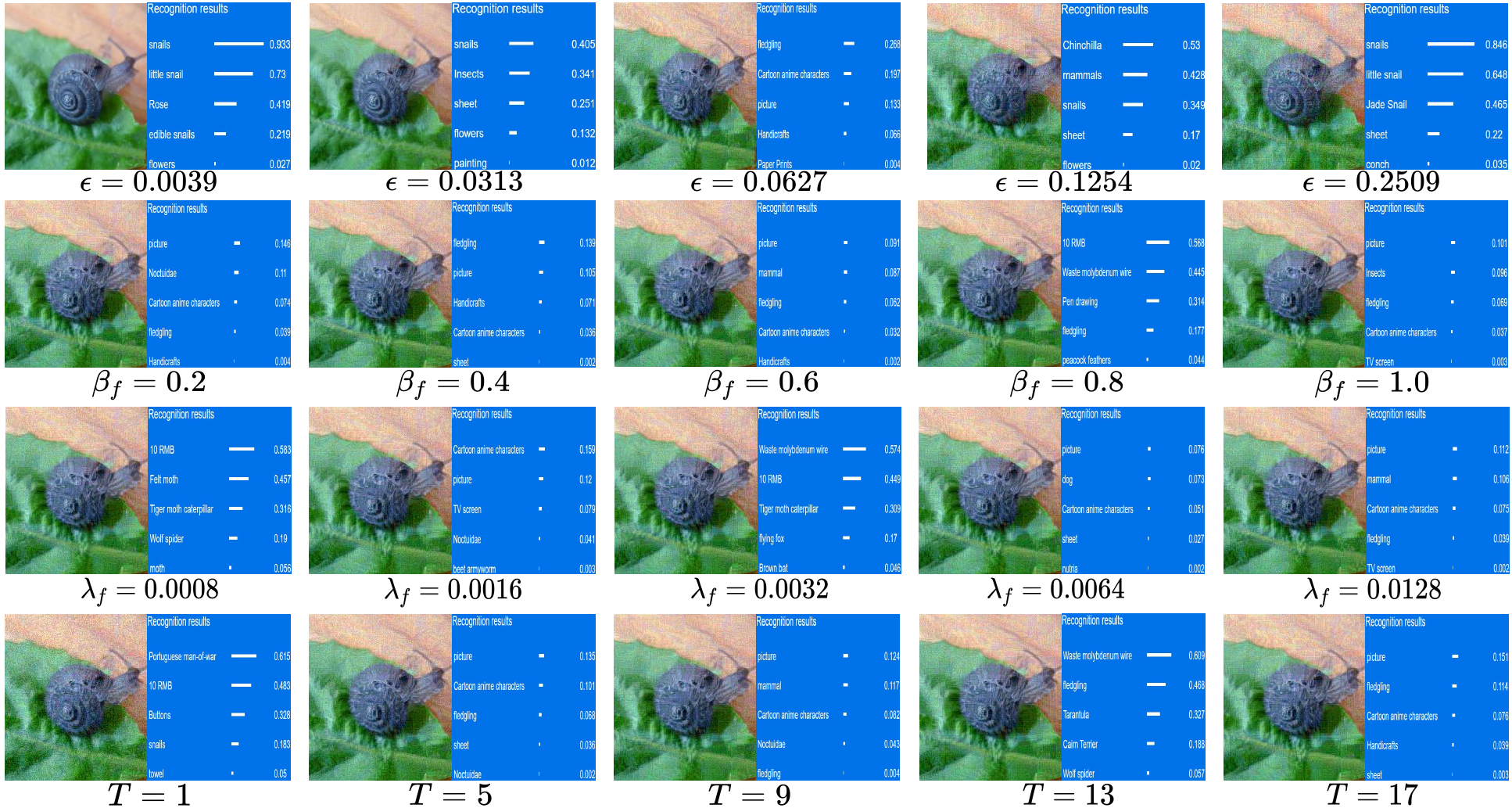}
\caption{Samples of the recognition results of adversarial examples generated by AFA with different hyperparameter ablation settings on the Baidu Cloud API. The white-box source model is Swin-T.}
\label{ablation_bdc_robustness_applicability}
\end{figure}

Given the importance of robustness and applicability in a successful attack method, we conduct further ablation experiments on the transformer-based surrogate (Swin-T) for $\beta_f$, $\lambda_f$, $T$ and $\epsilon$ as follows: (1) attacking defense models such as HGD, NRP, JPEG and Bit-Red. And (2) attack the Baidu Cloud API. In our experiment, robustness is demonstrated through the validation of the defense strategies. On the other hand, the applicability is proved through the substitution of the CNN-based surrogate with the transformer-based surrogate, and the validation on the commercial API. 
Notably, $\epsilon$ defaults to 0.0627, while the default values for the other hyperparameters can be found in \ref{app6}. First, as shown in Fig. \ref{ablation_hyperparam_robustness_applicability} (a) (b) (c), it has been verified that our chosen hyperparameter values for $\beta_f$, $\lambda_f$ and $\epsilon$ are feasible among the selected options for each hyperparameter, as they provide the best robustness to the four defense models. And, in Fig. \ref{ablation_hyperparam_robustness_applicability} (d), it can be observed that the ASRs stabilize gradually when $T$ exceeds 9, although better performance may still be possible. However, to balance computational cost and performance, $T=10$ is also a suitable choice. Besides, as displayed in Fig. \ref{ablation_bdc_robustness_applicability}, when attacking the Baidu Cloud API, there are some meaningful observations: (1) Only when $\epsilon = 0.0627$ does our method successfully deceive the system into recognizing objects as anything other than ``snails''. (2) With $\beta_f$ gets close to 0 or 1, the recognition results appear the classes close to ``snails'', such as ``noctuidae'' in $\beta_f=0.2$ and ``insects'' in $\beta_f=1.0$. It can verify the dual-order solution (AF) again. (3) The ``snails'' class and the similar class vanish when $\lambda_f \ge 0.0032$, corresponding to the results in Fig. \ref{ablation_hyperparam_robustness_applicability} (b). (4) Our method can thoughtfully confuse the system when $T \ge 9$. This section comprehensively explores hyperparameter ablation with respect to robustness and applicability.

\section{Ablation of each component in the defense model setting}
\label{app7}
To sufficiently verify our proposed method, we take another step to carry out more ablation experiments of our AF and MCAS in the defense model setting. The surrogate model is set as Inc-v3, and the defense models maintain consistency with previous experiments. Besides, the three experimental variants are consistent with the previous ablation section. As shown in Table \ref{ablation_defend}, combined with our proposed AF, the inner-sampling MI method significantly shows stronger invasion. And compared with the results of Table \ref{single_defense}, it is observed that our proposed AF can bring higher 3\%, 7\%, 13\%, 20\%, 25\% and 32\% ASRs than PGN, FGSRA, NCS, MuMoDIG, RAP and  GI-MI on average. Therefore, it indicates that our proposed AF consistently maintains the best attack capability even in the defense model setting. Finally, integrated with our MCAS, our performance is further improved, which verifies our designed sampling method.

\begin{table}[!t]
\centering
\caption{Ablation study conducted under the defense model setting. The white-box source model is Inc-v3. The \checkmark and \XSolidBrush symbols indicate our method with and without the corresponding component respectively.}
\resizebox{10cm}{0.8cm}
{
\begin{tabular}{c|c|cccccccccc}
		\hline
		\multicolumn{1}{c|}{Model}  
		 &\multicolumn{1}{c}{AF} & \multicolumn{1}{c|}{MCAS} & \multicolumn{1}{c}{RS} & \multicolumn{1}{c}{HGD} &\multicolumn{1}{c}{FD} & \multicolumn{1}{c}{NRP} & \multicolumn{1}{c}{ComDefend} &\multicolumn{1}{c}{JPEG} & \multicolumn{1}{c}{Bit-Red}& \multicolumn{1}{c}{R\&P}& \multicolumn{1}{c}{NIPS-r3}  \\ \hline

 \multirow{3}{*}{Inc-v3} & \multicolumn{1}{c}{\XSolidBrush} &\multicolumn{1}{c|}{\XSolidBrush} & 40.20 & 39.10 & 75.40 & 70.10 & 73.20 & 76.60 & 79.40 & 79.00 & 50.10\\
 &\multicolumn{1}{c}{\checkmark} &\multicolumn{1}{c|}{\XSolidBrush} & 43.50 & 42.20 & 86.20 & 74.00 & 81.30 & 87.20 & 88.80 &90.50 & 58.80\\

 &\multicolumn{1}{c}{\checkmark} &\multicolumn{1}{c|}{\checkmark} & \textbf{44.20} & \textbf{43.00} & \textbf{87.10} & \textbf{74.90} & \textbf{82.40} & \textbf{87.40} & \textbf{89.60} & \textbf{91.60}& \textbf{59.30}\\ \hline
  \end{tabular}
 \label{ablation_defend}
 }
\end{table}

\begin{table}[!t]
\centering
\caption{Quantity study of the average computational efficiency in our method. The white-box source models are Inc-v3, Res-101, ViT-B/16 and Swin-T. The \checkmark and \XSolidBrush symbols indicate our method with and without the corresponding component respectively.}
\resizebox{8cm}{2.5cm}
{
\begin{tabular}{c|cc|cc}
		\hline
		\multicolumn{1}{c|}{Model}  
		 &\multicolumn{1}{c}{AF} & \multicolumn{1}{c|}{MCAS} & \multicolumn{1}{c}{Wall-Clock Time (s)} & \multicolumn{1}{c}{GPU Memory (MB)}\\ \hline

 \multirow{3}{*}{Inc-v3} & \multicolumn{1}{c}{\XSolidBrush} &\multicolumn{1}{c|}{\XSolidBrush} & 0.62 & 170.0 \\
 &\multicolumn{1}{c}{\checkmark} &\multicolumn{1}{c|}{\XSolidBrush} & 2.34 & 182.8 \\
 &\multicolumn{1}{c}{\checkmark} &\multicolumn{1}{c|}{\checkmark} & 2.43  & 184.0  \\ \hline
 
  \multirow{3}{*}{Res-101} & \multicolumn{1}{c}{\XSolidBrush} &\multicolumn{1}{c|}{\XSolidBrush} & 0.99  & 320.2 \\
 &\multicolumn{1}{c}{\checkmark} &\multicolumn{1}{c|}{\XSolidBrush} & 3.85 &  333.0\\
 &\multicolumn{1}{c}{\checkmark} &\multicolumn{1}{c|}{\checkmark} & 4.05 & 334.2 \\ \hline

   \multirow{3}{*}{ViT-B/16} & \multicolumn{1}{c}{\XSolidBrush} &\multicolumn{1}{c|}{\XSolidBrush} & 1.25 & 212.4 \\
 &\multicolumn{1}{c}{\checkmark} &\multicolumn{1}{c|}{\XSolidBrush} &5.07 & 217.2\\
 &\multicolumn{1}{c}{\checkmark} &\multicolumn{1}{c|}{\checkmark} & 5.18  & 219.2 \\ \hline

    \multirow{3}{*}{Swin-T} & \multicolumn{1}{c}{\XSolidBrush} &\multicolumn{1}{c|}{\XSolidBrush} & 0.65 & 168.2 \\
 &\multicolumn{1}{c}{\checkmark} &\multicolumn{1}{c|}{\XSolidBrush} & 2.51  &  168.8 \\
 &\multicolumn{1}{c}{\checkmark} &\multicolumn{1}{c|}{\checkmark} & 2.67  &  169.4 \\ \hline
 
  \end{tabular}
 \label{computationa_efficiency_MB_s}
 }
\end{table}

\section{Discussion on the computational efficiency of our method}
\label{computational efficiency}
As computational efficiency is important for practical applications, we design some inference validation on a GPU to quantify the wall-clock time and GPU memory. Then, we choose surrogate models with diverse architectures to set different scenarios. Specifically, in each scenario, we all test the computational efficiency for the variants with (1) a clean baseline (referred to in the previous ablation section), (2) only AF and (3) the additional MCAS. Due to the parallel computation in GPUs, we set the batch size as 10. Notably, the results are presented on average. As shown in Table \ref{computationa_efficiency_MB_s}, it is clear that the inference time and memory usage are closely linked to the number of parameters in the surrogate models, especially for AF. Additionally, our proposed MCAS is very lightweight, and the processing time is relatively short.

\end{document}